\documentclass[11pt]{article}

\usepackage{arxiv}

\pdfoutput=1

\usepackage[T1]{fontenc}
\usepackage[utf8]{inputenc}
\usepackage{times}
\usepackage{latexsym}
\usepackage{inconsolata}
\usepackage{microtype}
\usepackage{natbib}
\usepackage{amsmath}
\usepackage{amssymb}
\usepackage{placeins}
\usepackage{booktabs}
\usepackage{tabularx}
\usepackage{array}
\usepackage{multirow}
\usepackage{makecell}

\usepackage{graphicx}
\usepackage{float}
\usepackage{caption}
\usepackage{url}
\usepackage{algorithm}
\usepackage{algorithmic}

\usepackage[most]{tcolorbox}
\usepackage{xcolor}

\newcolumntype{Y}{>{\raggedright\arraybackslash}X}

\usepackage{hyperref}

\hypersetup{
    hidelinks
}

\title{Mechanistic Interpretability of Chain-of-Thought Reasoning via Sequential Activation Patching}

\author{
Murat Dura\thanks{These authors contributed equally to this work.} \\
Department of Computer Engineering \\
İzmir Institute of Technology \\
İzmir, Türkiye
\And
Serkan Öztürk\footnotemark[1] \\
Department of Computer Engineering \\
İzmir Institute of Technology \\
İzmir, Türkiye
\And
Selma Tekir \\
Department of Computer Engineering \\
İzmir Institute of Technology \\
İzmir, Türkiye
}

\begin{document}
\maketitle

\fancyhead{}
\fancyhead[L]{\footnotesize\itshape
Mechanistic Interpretability of CoT via Sequential Activation Patching}
\fancyhead[R]{\footnotesize\itshape A PREPRINT}

\begin{abstract}
Large Language Models (LLMs) demonstrate remarkable problem-solving
capabilities when guided by Chain-of-Thought (CoT) prompting, yet the
internal mechanisms underlying these improvements remain poorly understood.
In this work, we investigate where CoT-related causal effects emerge across
the generated reasoning trajectory and which attention heads carry signals
that contribute to final-answer computation. Because CoT reasoning unfolds
over multiple generated tokens, standard activation patching at a single
static token position is insufficient to characterize these temporally
distributed effects. To address this limitation, we introduce a sequential
activation patching framework that traces CoT-conditioned attention-head
activations across token positions and aggregates their effects using
Part-of-Speech-guided analysis. We further introduce Sequential Multi-Head
Patching to evaluate the joint contribution of distributed head sets, together
with cross-question and random-activation controls. Targeted zero-ablation
experiments show that the identified heads are functionally important for
successful answer generation and affect several overlapping mechanisms,
including reasoning-trajectory maintenance, answer anchoring, exemplar--target
separation, and numerical generation. Overall, our results provide evidence
for distributed reasoning-support sub-circuits associated with
CoT-conditioned computation.
\end{abstract}


\section{Introduction} Large Language Models (LLMs) have demonstrated strong performance on complex reasoning tasks when guided by Chain-of-Thought (CoT) prompting, where the model is encouraged to produce intermediate reasoning steps before committing to a final answer \citep{wei-etal-2022}. This improvement is especially visible in mathematical and multi-step problem solving, where prompt-based CoT often helps the model decompose a problem into smaller computational stages. Throughout this paper, we use “CoT” to refer specifically to prompt-based CoT elicited from standard language models, rather than reasoning traces generated by specialized reasoning models. However, the mechanism behind this improvement remains poorly understood. In particular, it is still unclear whether CoT merely changes the surface form of the generated text, or whether it causally reorganizes the model's internal computation in a way that supports reasoning. This question is closely tied to the broader problem of CoT faithfulness. Although CoT traces are often read as explanations of the model's reasoning process, prior work has shown that generated rationales can be misleading. \citet{turpin-etal-2023} demonstrate that models may rationalize answers influenced by biased or misleading features while failing to acknowledge those influences in the explanation. Similarly, other studies have shown that the causal role of intermediate reasoning steps varies across tasks, models, and evaluation settings \citep{lanham-etal-2023, paul-etal-2024}. These findings suggest that CoT should not be treated as automatically faithful simply because it produces plausible intermediate text. At the same time, recent work also cautions against equating incomplete verbalization with unfaithfulness. \citet{zaman-srivastava2026} argue that a CoT trace may be faithful even when it does not explicitly verbalize every causally relevant feature. In their analysis, many CoT outputs labeled as unfaithful by hint-verbalization-based metrics are judged faithful under alternative metrics, and causal mediation analysis shows that even non-verbalized hints can still influence predictions through the CoT. 

Mechanistic interpretability provides tools for studying this internal role more directly. Activation patching, logit-based analysis, and distribution-level comparisons can test whether specific internal activations causally shift model behavior toward a target computation \citep{zhang2024bestpracticesactivationpatching, dutta2024thinkstepbystepmechanisticunderstanding}. Recent mechanistic work further suggests that CoT changes the structure of internal representations, for example by inducing sparser or more specialized activation patterns during reasoning \citep{chen-etal-2026}. However, standard intervention methods are often applied at isolated token positions, whereas CoT reasoning unfolds across a sequence of generated tokens. This creates a gap between the temporal nature of CoT and the static structure of many existing mechanistic analyses. 

Rather than asking whether CoT prompting alters the model's internal
computation, we focus on where and how these CoT-related effects are
distributed across the generated reasoning trajectory. Specifically, we
investigate which generated token positions and which attention heads carry
causally relevant signals that distinguish CoT from No-CoT computation and
contribute to final-answer generation. This formulation motivates our
sequential intervention framework, which traces CoT-conditioned activations
across token positions and attention heads rather than restricting the
analysis to a single static intervention point.

We investigate these questions in the setting of mathematical word problems, using Qwen2.5-0.5B \citep{qwen2025qwen25technicalreport}, where the reasoning structure is relatively well-defined and the contrast between CoT and No-CoT prompting can be directly examined. We analyze whether CoT-conditioned internal activations can causally move No-CoT executions toward CoT-like behavior. We adopt sequential activation patching, where the outputs of attention heads from the \textbf{clean run} (the CoT prompting condition) are injected into the \textbf{corrupted run} (No-CoT). This setting allows us to ask whether intervention-sensitive internal components carry signals that influence the final answer. We summarize our contributions as follows: \begin{itemize} \item We adopt a many-to-one, sequential activation patching strategy, where the outputs of CoT attention heads are injected into No-CoT. Using distribution-level and logit-level metrics, we show that sequential activation patching moves No-CoT executions toward CoT behavior, suggesting that CoT-conditioned activations causally contribute to final-answer computation. We publicly share our code to support reproducibility.
\footnote{\url{https://github.com/MuratDura/CoT_Mechanistic_Interpretability}}
\item  Targeted CoT activation patching consistently moves No-CoT executions toward CoT-like behavior, while cross-question patching shows weaker but still meaningful transfer, suggesting that the identified activations are not limited to problem-specific tokens. In contrast, random activation patching fails to produce the same effect, indicating that the observed improvements are driven by structured CoT-related signals rather than arbitrary perturbations. \item Targeted zero-ablation further shows that the identified components are functionally important for successful reasoning. Removing them disrupts not only final-answer accuracy, but also auxiliary mechanisms such as answer anchoring, exemplar--target separation, answer-slot binding, and numerical generation. We therefore interpret the discovered structure as a distributed reasoning-support sub-circuit dynamically recruited by CoT. \end{itemize}

\section{Related Work} Our research intersects the behavioral evaluation of CoT prompting in mathematical reasoning and the mechanistic interpretability of transformer-based models. For a comprehensive overview of multi-step reasoning mechanisms in LLMs, we refer readers to the recent survey by \citet{pan2026openingblackboxsurvey}. \subsection{CoT Prompting and Mathematical Reasoning} CoT prompting yields considerable performance gains on mathematical word problems \citep{wei-etal-2022, kojima2023largelanguagemodelszeroshot}, yet its operational faithfulness---whether internal models truly align with behavioral explanations---remains a longstanding challenge \citep{jacovi2020faithfullyinterpretablenlpsystems}. Generated rationales do not always reflect actual internal computation, leading to various failure modes such as post-hoc or encoded reasoning \citep{lanham-etal-2023}. Unfaithfulness metrics have been proposed and subsequently refined to account for factors like positional answer bias \citep{bentham-etal-2024}. Complementary approaches, including unlearning reasoning steps \citep{tutek-etal-2025} and causal mediation analysis \citep{paul-etal-2024}, further reveal that LLMs often bypass their own rationales when producing final answers. Recent work supports this by showing that reasoning traces can become merely performative \citep{li2026reasoningtracesperformativesteplevel}, and when CoT fails, the correct solution may still hide within the model's hidden states \citep{mehrafarin2026chainofthoughtfailssolutionhides}. Conversely, \citet{kudo2026llmsfaithfullyiterativelycompute} demonstrate that language models can indeed faithfully and iteratively compute mathematical answers across intermediate tokens. These mixed behavioral findings highlight a gap between plausible text generation and faithful computation, motivating our mechanistic approach to uncover CoT's true causal role. 

\subsection{Mechanistic Interpretability and Internal Computations} Mechanistic interpretability aims to identify internal components that causally influence model behavior, prominently using activation patching, which has recently been applied to evaluate the mechanistic faithfulness of natural language rationales \citep{yeo2024faithfulnaturallanguageexplanations}. The field is heavily influenced by seminal works that map specific behaviors, such as Indirect Object Identification, to interpretable transformer circuits \citep{wang2022interpretabilitywildcircuitindirect}. As formalized by \citet{zhang2024bestpracticesactivationpatching}, the interpretation of patching heavily depends on methodological choices, particularly the evaluation metrics. Token-level metrics, such as the Normalized Logit Difference Decay (NLDD) proposed by \citet{ye-etal-2026}, effectively capture local preference shifts and determine the decisive ``Reasoning Horizon'' for specific tokens. However, they do not capture broader changes in the model's output behavior. To address this, distribution-level metrics are employed to measure global convergence. For instance, \citet{dutta2024thinkstepbystepmechanisticunderstanding} utilize Kullback--Leibler (KL) divergence to evaluate how closely patched executions recover the global probabilistic patterns associated with CoT reasoning. Beyond evaluating final outputs, mechanistic studies actively map how fundamental operations are distributed. Causal mediation analysis isolates specific circuits dedicated to arithmetic operations \citep{stolfo2023mechanisticinterpretationarithmeticreasoning}, while related methods verify CoT reasoning via its underlying computational graphs \citep{zhao2026verifyingchainofthoughtreasoningcomputational}. Furthermore, research has identified architectural components driving iterative generation, such as Iteration Heads \citep{cabannes2024iterationheadmechanisticstudy}, and foundational sub-circuits like induction heads \citep{olsson2022incontextlearninginductionheads} that govern in-context copying and boundary maintenance. Finally, the sequential nature of CoT is supported by findings that multi-token sequences act as an essential computational buffer for latent mathematical operations \citep{pfau2024letsthinkdotdot}.

\section{Methodology} In mechanistic interpretability studies, the \textbf{clean run} represents the computation where the target behavior is present, while the \textbf{corrupted run} defines a counterfactual setting where this behavior is absent. As highlighted by \citet{zhang2024bestpracticesactivationpatching}, the success of activation patching depends on the careful selection of a corruption strategy and the corresponding evaluation metrics. Afterward, causal influence is evaluated by intervening on internal activations and measuring the resulting change in the model's output using divergence on probability distributions and logit-based metrics. 

\subsection{Dataset and Prompting Strategy} We evaluate our model on the SVAMP dataset \citep{patel-et-al-2021}, a benchmark comprising simple mathematical word problems specifically designed to test multi-step reasoning capabilities. We employ a controlled one-shot CoT prompting setup, adopting the exemplar formatting introduced by \citet{wei-etal-2022}. Our prompt structures are dynamically assembled according to the mathematical operation category. For the CoT prompt, we use a fixed, manually curated one-shot CoT example tailored to the target problem's operation type, following the format ``Q: [Question] A: [CoT Included Answer]. Q: [Target Question] A:''. For the No-CoT prompt, we dynamically sample a one-shot direct-answer example of the same operation type from the dataset, following the format ``Q: [Sampled Dataset Question] A: The answer is [Sampled Answer]. Q: [Target Question] A:''. 

Crucially, we enforce the use of a standardized anchor phrase, ``The answer is ''. In the No-CoT condition, the exemplar directly provides this anchor immediately after the question. In the CoT condition, the exemplar demonstrates a step-by-step reasoning trace before concluding with the same anchor phrase. This structural constraint is vital; it explicitly guides the model to format its output consistently, enabling us to reliably identify and align the exact token position where the model commits to its final answer across both clean and corrupted executions.

\subsection{Corruption Method} CoT prompting presents unique challenges for defining a corruption strategy. Unlike single-token entity retrieval, CoT is a multi-step process that encompasses varying token positions and sequence lengths. It is difficult to validate strategies that intervene in a reasoning trace by substituting mathematical operators or altering numerical values \citep{cheng2025llmsreasonabstractlymath}, as such modifications may interfere with the CoT mechanism itself beyond merely inducing semantic noise. To overcome such issues, in our design, the \textbf{clean run} represents the CoT prompting condition, while the \textbf{corrupted run} corresponds to No-CoT. This setting allows a controlled comparison where the difference between runs is the presence of reasoning traces. The aim is to identify CoT-conditioned internal components whose activations can causally shift No-CoT executions toward the CoT reference.

\subsection{Evaluation Metrics} To quantify the effect of each intervention across these executions, we employ distribution-level and logit-level metrics to comprehensively evaluate the observed shift. \subsubsection{Jensen--Shannon Divergence} We measure the distributional similarity between the patched run and the clean (CoT) run using Jensen--Shannon Divergence (JSD). This choice is further motivated by recent findings indicating that CoT prompting induces significant sparsity in the model's internal representations, effectively redistributing the computation so that only a few specific neurons or components become active \citep{chen-etal-2026}. Since this sparse activation pattern implies that CoT does not merely alter the top-1 prediction but reshapes the model's global probabilistic landscape, we argue that the causal fingerprint of CoT should be evaluated across the entire output space. JSD captures global alignment between output distributions. Lower values indicate that the patched model produces a distribution closer to the CoT behavior. Details are provided in Appendix~\ref{sec:jsd_details}. \subsubsection{Margin Recovery Ratio} To capture decision-level effects, we define a logit-based margin metric. Let $t^*$ denote the correct answer token. The margin is defined as the logit difference between the correct answer token and the highest-scoring alternative token: \begin{equation} M = z(t^*) - \max_{t \neq t^*} z(t). \end{equation} We compute margins for three conditions: $M_{\text{CoT}}$ (clean), $M_{\text{NoCoT}}$ (corrupted), and $M_{\text{patch}}$ (patched). We define the \textbf{Margin Recovery Ratio (MRR)} as: \begin{equation} \mathrm{MRR} = \frac{M_{\text{patch}} - M_{\text{NoCoT}}} {M_{\text{CoT}} - M_{\text{NoCoT}}}. \end{equation} MRR measures how much of the CoT--No-CoT margin difference is recovered by patching. Details are provided in Appendix~\ref{sec:mrr_details}. 
\subsection{Metric Complementarity} These two metrics address complementary causal questions, encompassing both global convergence and local confidence: \textbf{JSD} evaluates shifts in the \textit{patched distribution} toward CoT behavior, while \textbf{MRR} measures shifts in the \textit{decision boundary} favoring the correct answer.

\subsection{Sequential Activation Patching} Standard activation patching methodologies are conventionally designed to intervene at a single, static token position. However, because CoT reasoning inherently unfolds over a sequence of multiple intermediate tokens, applying this standard setup directly is insufficient. Therefore, it is necessary to formulate a sequential patching approach (Appendix~\ref{app:algorithms}, Algorithm~\ref{alg:patching_collection_aggregation}). Building on this, we perform causal interventions using activation patching, where the outputs of attention heads from the clean run are injected into the corrupted run. Treating CoT as a multi-step process and operating under the hypothesis that CoT information is carried throughout the reasoning trajectory, we adopt a many-to-one patching strategy. For each source CoT token $t$, layer $l$, and head $h$, we replace only that head's output at the No-CoT next-token position immediately following the anchor ``The answer is ''. Each $(t,l,h)$ intervention is evaluated in a separate forward pass, while all other head outputs remain unchanged (Appendix~\ref{app:algorithms}, Algorithm~\ref{alg:patching_collection_aggregation}, lines 9--22). This procedure produces a separate patched execution for each intervention. 

\begin{figure*}[t] \centerline{\includegraphics[width=1\textwidth]{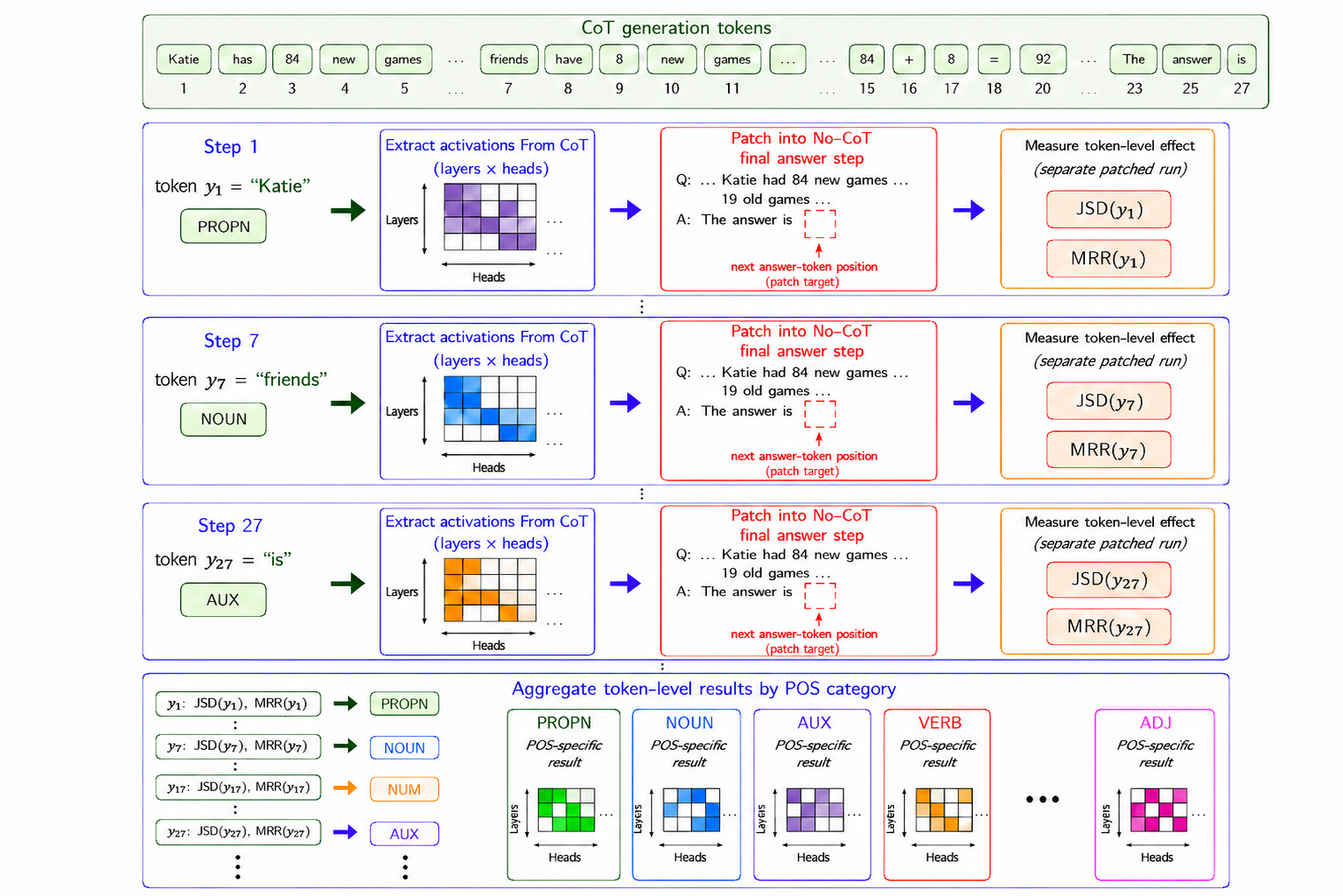}} \caption{\small The figure illustrates our sequential activation patching setup (Appendix~\ref{app:algorithms}, Algorithm~\ref{alg:patching_collection_aggregation}). First, the model generates a Chain-of-Thought (CoT) token sequence, where each token $y_t$ corresponds to a generated reasoning token such as ``Katie,'' ``friends,'' or ``is.'' For each token, we extract the attention-head activations from the clean CoT run and patch them into the corrupted No-CoT run at the final answer generation position. The effect of each patched token is then measured using JSD and MRR. Since individual token-level effects can be noisy, the results are grouped by Part-of-Speech (POS) categories such as [VB], [NNP], [CD], [NNS], and [VBZ]. For each POS category, token-level patching results are aggregated into a layer-by-head heatmap, showing which attention heads consistently carry CoT-related information for that linguistic role.} \label{fig:sequential_activation_patching} \end{figure*}

However, analyzing these individual token-level results directly introduces significant noise, as the causal impact of a patched activation is often highly sensitive to the specific semantic and syntactic properties of tokens. To mitigate this noise and identify more robust computational patterns, we implement a Part-of-Speech (POS)-guided aggregation framework (Figure~\ref{fig:sequential_activation_patching}). We assign POS tags to each token in the reasoning trace and aggregate patching results across these linguistic categories (Appendix~\ref{app:algorithms}, Algorithm~\ref{alg:patching_collection_aggregation}, lines 23--33). Our goal in using POS-based aggregation is not to obscure individual POS-tag behavior, but to mitigate token-level noise and identify stable functional patterns across the CoT trajectory. By aggregating patching results across tokens sharing the same coarse POS category, we obtain more robust estimates of which heads consistently carry CoT-related signals. The selected POS categories and their corresponding head assignments are detailed in Appendix Tables~\ref{tab:normal_patching_jsd} and~\ref{tab:cross_patching_jsd}. 

This aggregation is crucial because CoT reasoning unfolds over multiple heterogeneous tokens, where a single token-level intervention may fail to capture the broader causal role of that segment in the reasoning trace. While semantic role labeling could offer an alternative categorical structure for this aggregation---a direction we identify for future research---POS tagging provides an initial framework for grouping tokens. This aggregation process transforms noisy, token-specific data into a single representative activation patching heatmap and a custom evaluation for each POS category. This, in turn, enables a fine-grained analysis of how causal signals are distributed across the reasoning process and helps identify specific heads that play a central role in processing different types of information within the CoT trajectory. 

On a single NVIDIA RTX 6000 Ada GPU, the patching operation for one specific token position takes approximately 29.58 seconds using the TransformerLens library. Since our methodology involves scanning and patching all generated tokens within a CoT trace, the total analysis time per sample scales linearly with the length of the reasoning chain. Importantly, the overhead introduced by the POS-tagging step is exceptionally low: POS tagging for all 32 samples takes only 0.24 seconds in total, which is negligible compared to the computational cost of the repeated model forward passes required for these sequential evaluations. 

\subsection{Sequential Multi-Head Patching} Our sequential patching setup provides a comprehensive dataset containing evaluation metric results for all layer and attention-head combinations across each POS category (Appendix~\ref{app:head_selection_overlap}, Figure~\ref{fig:multi_head_patching_appendix}). However, during our experiments, we observed that isolated interventions---patching only a single head at a time---failed to produce a meaningful shift in the model's output toward CoT behavior. This indicates that CoT information is not carried by a single component in isolation; rather, it relies on a distributed representation formed by the coordinated interaction of multiple components across different functional roles. 

Motivated by this observation, we implement a selection and merging strategy to reconstruct broader CoT-related sub-circuits (Appendix~\ref{app:algorithms}, Algorithm~\ref{alg:head_selection}). First, we identify the top-$k$ most influential attention heads for each individual POS category (e.g., those yielding the lowest JSD scores). Instead of evaluating these functional categories in isolation, we aggregate these category-specific top heads into a single global intervention set. During this aggregation, we implement a deduplication mechanism: if a specific attention head appears among the top candidates for multiple POS categories, we assign it exclusively to the category in which it achieves its best overall performance. This resolves intervention conflicts by ensuring that each shared head contributes only the activation state corresponding to its most effective functional role. Finally, we perform global multi-head joint patching using this aggregated set. We simultaneously extract the targeted activations for all uniquely assigned heads in the global set from the clean CoT run and jointly inject them into the corrupted No-CoT run during a single forward pass. By computing the final JSD and MRR on the resulting distribution, we measure the synergistic effect of the reconstructed sub-circuit. 

This approach allows us to directly quantify how the collective integration of multiple POS-guided components drives the model's internal representations toward the prompt-based CoT-induced reasoning trajectory. 

To provide further evidence for our hypothesis that patched runs move toward a CoT reasoning trajectory, we introduce two control conditions: Sequential Multi-Head Cross Patching, which performs sequential multi-head patching across different problems, and Random-Activation Sequential Multi-Head Patching. Detailed explanations and corresponding pseudocode are provided in Appendix~\ref{app:algorithms}, Algorithms \ref{alg:cross_patching} and ~\ref{alg:random_patching}. 
\subsection{Zero Ablation of Identified Attention-Head Sets} \label{sec:ablation_methodology} While activation patching highlights which attention heads carry sufficient information to shift the model's behavior toward a CoT-like distribution, it does not strictly establish that these heads are necessary for the reasoning process. To examine whether the heads identified through our patching experiments constitute such critical overlapping sub-circuits, we design targeted zero-ablation experiments (Appendix~\ref{app:algorithms}, Algorithm~\ref{alg:zero_ablation}). 

We systematically zero-ablate the attention heads identified from our previous Sequential Multi-Head Patching, Sequential Multi-Head Cross Patching, and Random-Activation Sequential Multi-Head Patching setups. To evaluate the functional role of these heads, we perform zero-ablation across three scenarios: \textbf{Direct-Equation Prompting}, where we ablate heads during raw arithmetic generation (e.g., \texttt{14 + 312 =}) to test for overlap with fundamental calculation circuits; \textbf{No-CoT Prompting}, where we ablate heads during direct answer generation to test our hypothesis that these heads encompass core abstraction and calculation circuits; and \textbf{CoT Prompting}, where we continuously ablate the heads at every token-generation step to assess their necessity for maintaining reasoning states throughout the entire trajectory. We evaluate the overall functional degradation caused by the intervention across these three scenarios. If the ablation yields the anticipated performance drop, it provides evidence that the heads identified through our patching experiments contribute to the sub-circuits required to solve mathematical problems and that these sub-circuits are effectively recruited under CoT prompting. Conversely, we use the heads from the Random-Activation Sequential Multi-Head Patching setup and completely randomly selected head sets as control conditions to ensure that any observed degradation is specifically tied to the disruption of the targeted circuits rather than being a generalized artifact of network-wide perturbation.

\section{Experimental Results}
\label{sec:experimental_results}

\subsection{Patching Results}

\begin{figure}[!htbp]
    \centering
    \includegraphics[width=0.62\linewidth]{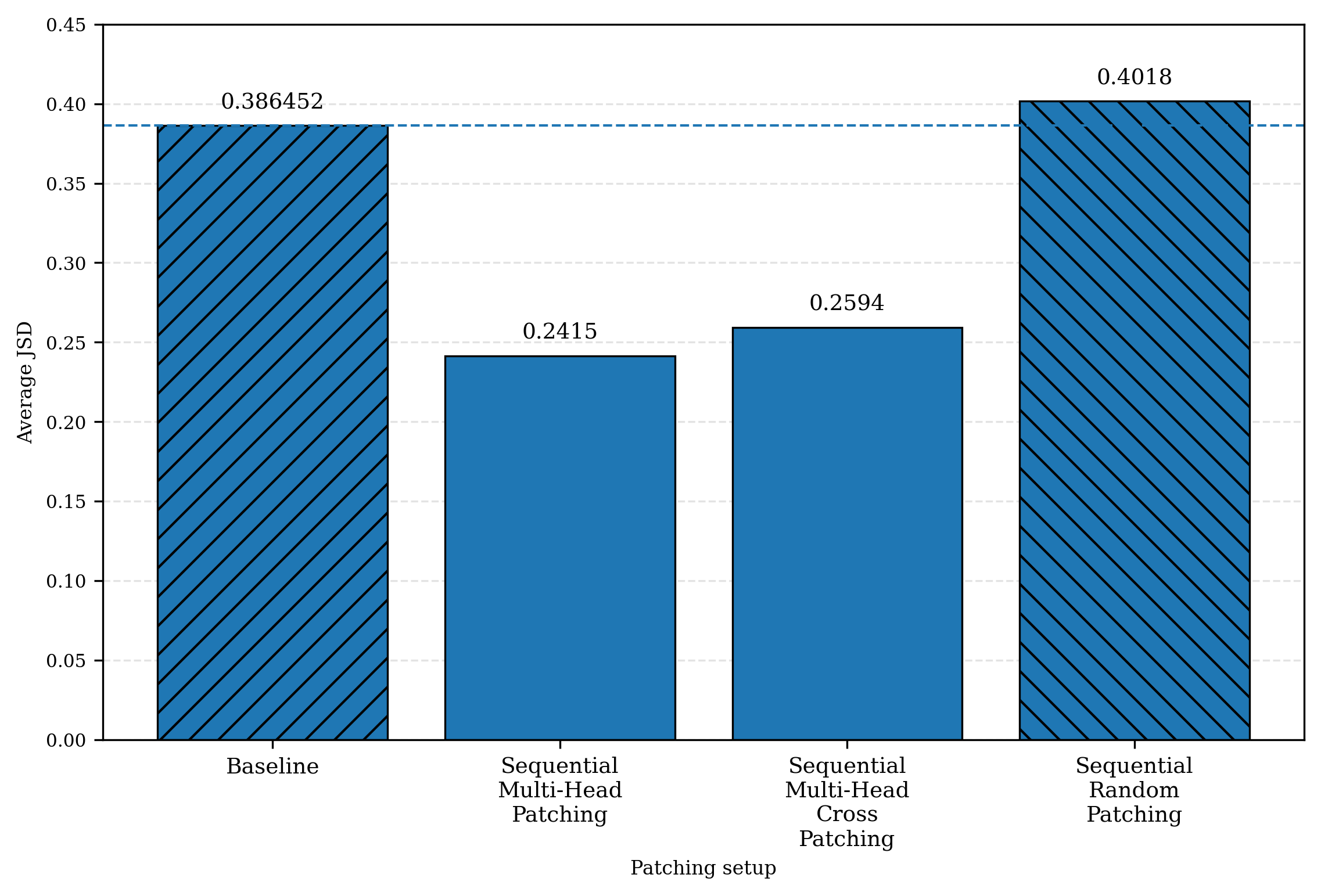}
    \caption{\small
    Comparison of average JSD across different patching setups.
    The figure displays the baseline JSD ($0.3865$) alongside the final
    JSD scores for Sequential Multi-Head Patching
    (Table~\ref{tab:normal_patching_jsd}),
    Sequential Multi-Head Cross Patching
    (Table~\ref{tab:cross_patching_jsd}), and
    Random-Activation Sequential Multi-Head Patching
    (Table~\ref{tab:full_results_random_jsd}).
    Lower JSD indicates greater distributional similarity to the clean CoT run.
    }
    \label{fig:jsd_comparison}
\end{figure}

We evaluate our causal intervention framework on 32 SVAMP mathematical
word problems using three activation patching setups: Sequential Multi-Head
Patching, Sequential Multi-Head Cross Patching, and Random-Activation
Sequential Multi-Head Patching. Intervention effects are evaluated using
JSD and MRR. The aggregate JSD and MRR results are summarized in
Figure~\ref{fig:jsd_comparison} and Figure~\ref{fig:mrr_comparison},
respectively, while detailed instance-level results and explanations are
provided in Appendix~\ref{sec:patching_results}.


\begin{figure}[!htbp]
    \centering
    \includegraphics[width=0.62\linewidth]{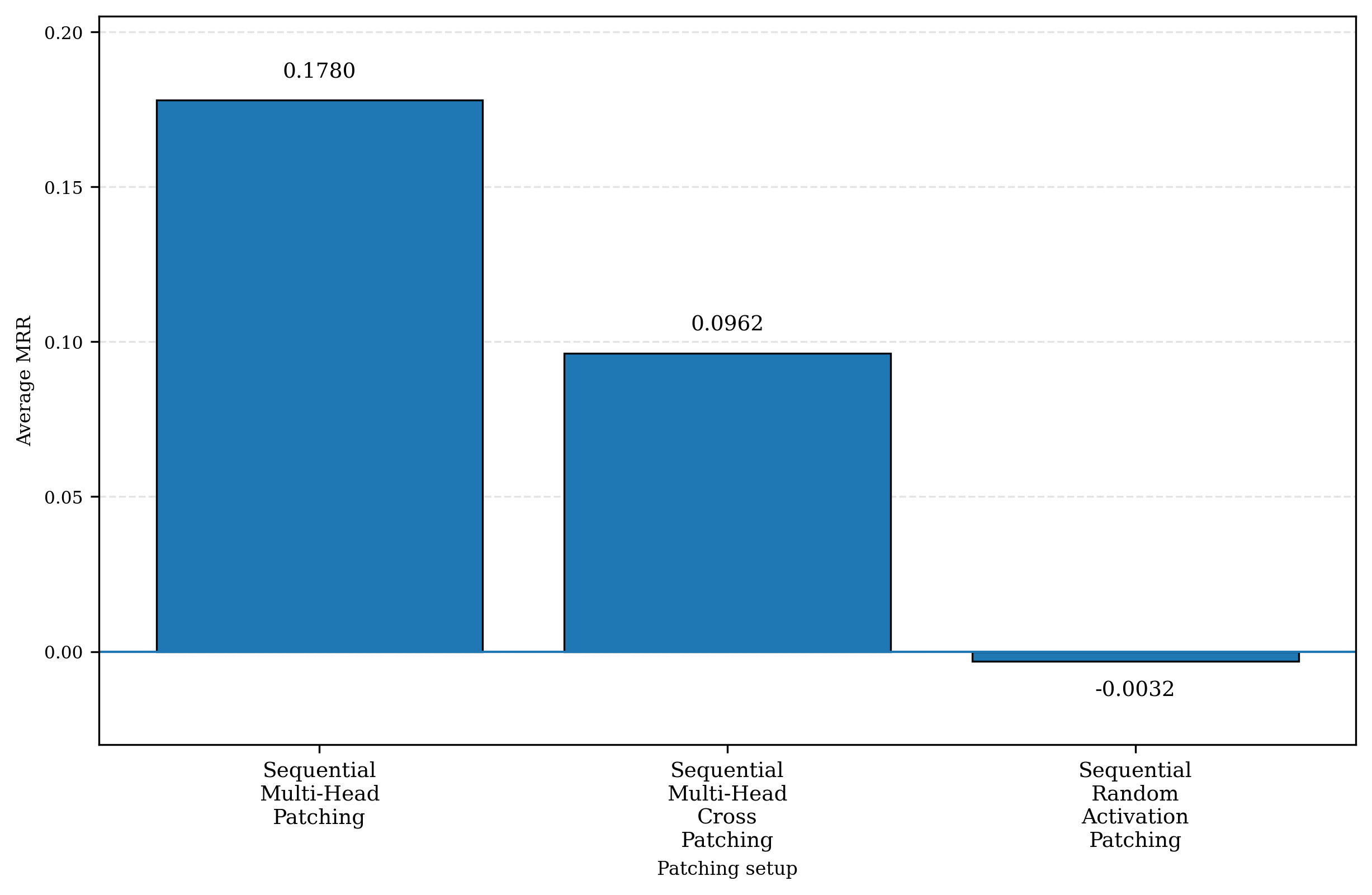}
    \caption{\small
    Comparison of average MRR across Sequential Multi-Head Patching
    (Table~\ref{tab:normal_patching_mrr}),
    Sequential Multi-Head Cross Patching
    (Table~\ref{tab:cross_patching_mrr}), and
    Random-Activation Sequential Multi-Head Patching
    (Table~\ref{tab:full_results_random_mrr}).
    Higher values indicate better recovery of the CoT--No-CoT
    decision-margin difference.
    }
    \label{fig:mrr_comparison}
\end{figure}

\FloatBarrier

\paragraph{Sequential Multi-Head Patching.}
In Sequential Multi-Head Patching, POS-guided activations from the clean
CoT run of a target question are injected into the No-CoT run of the same
question. As shown in Figure~\ref{fig:jsd_comparison} and
Figure~\ref{fig:mrr_comparison}, this intervention produces the strongest
improvement among the evaluated patching configurations. The patched
executions move substantially closer to the clean CoT distribution and
recover a meaningful portion of the CoT decision margin. These results
indicate that jointly injecting structured, POS-guided multi-head activations
can drive both the global output distribution and the local decision boundary
toward CoT-like behavior.

\paragraph{Sequential Multi-Head Cross Patching.}
In the cross-patching setup, source activations are taken from the CoT trace
of a different randomly selected question and injected into the target
question's No-CoT execution. As summarized in the same figures,
cross-patching also shifts the model toward the CoT reference, but the effect
is weaker than in the same-question Sequential Multi-Head Patching setting.
This reduction is expected because the source trace no longer contains the
target question's specific entities, numerical operands, or local semantic
structure. Nevertheless, the fact that cross-patching still produces a
non-trivial improvement suggests that the selected heads encode partially
reusable reasoning-support patterns rather than only problem-specific content.

\paragraph{Random-Activation Sequential Multi-Head Patching.}
Random-Activation Sequential Multi-Head Patching serves as a control
condition in which randomly generated activation tensors are injected instead
of actual CoT activations. As shown in Figure~\ref{fig:jsd_comparison} and
Figure~\ref{fig:mrr_comparison}, this intervention fails to reproduce the
improvements obtained by the targeted patching methods. The random control
therefore supports the interpretation that the observed effects are not caused
solely by arbitrary perturbations or generic sensitivity of attention heads,
but are associated with structured CoT-related signals carried by the selected
components.

\subsection{Zero Ablation Results}
\label{ZeroAblationResults}

\begin{table*}[t]
\centering
\small

\begin{tabular}{llccc}
\toprule

\textbf{Experiment}
& \textbf{Metric}
& \textbf{Equation Prompt}
& \textbf{No-CoT Prompt}
& \textbf{CoT Prompt} \\

&
&
\textbf{Accuracy}
&
\textbf{Accuracy}
&
\textbf{Accuracy} \\

\midrule

\textbf{Normal Run}
& N/A
& 34.38\%
& 31.25\%
& 65.63\% \\

\midrule

\multirow{2}{*}{Sequential Multi-Head Patching}
& JSD
& 9.38\%
& 12.50\%
& 0.00\% \\

& MRR
& 9.38\%
& 9.38\%
& 6.25\% \\

\midrule

\multirow{2}{*}{Sequential Multi-Head Cross Patching}
& JSD
& 6.25\%
& 6.25\%
& 9.38\% \\

& MRR
& 18.75\%
& 12.50\%
& 6.25\% \\

\midrule

\multirow{2}{*}{Random-Activation Sequential Multi-Head Patching}
& JSD
& 18.75\%
& 6.25\%
& 12.50\% \\

& MRR
& 6.25\%
& 12.50\%
& 28.12\% \\

\midrule

\textbf{Random $n$-Head Selection}
& N/A
& 34.38\%
& 15.63\%
& 37.50\% \\

\bottomrule
\end{tabular}

\caption{\small
Zero-ablation accuracy results across prompting settings.
The Normal Run row reports the original accuracy without ablation or
patching. All other rows report accuracy after zero-ablating the
corresponding selected head sets.
}
\label{tab:zero_ablation_results}

\end{table*}

Table~\ref{tab:zero_ablation_results} reports the effect of zero-ablating
the attention heads identified by our activation patching procedures.
The analysis covers three distinct prompting settings: CoT Prompting,
No-CoT Prompting, and Equation Prompting.

\paragraph{CoT Prompting.} The most pronounced degradation occurs in the CoT setting, highlighting the functional importance of the identified components. While the baseline CoT accuracy is $65.63\%$, zero-ablating the heads selected via Sequential Multi-Head Patching using JSD reduces the model's accuracy to $0.00\%$. By comparison, ablating a randomly selected set of heads reduces accuracy to $37.50\%$. It is important to distinguish this baseline from the Random-Activation Sequential Multi-Head Patching control, for which the corresponding ablations reduce accuracy to $12.50\%$ for JSD-based selection and $28.12\%$ for MRR-based selection. This condition is not a purely random baseline: because it ranks and selects heads according to their responses to injected random activations, it can preferentially identify perturbation-sensitive components. Nevertheless, the larger degradation observed for the targeted JSD-based Sequential Multi-Head Patching set provides causal evidence that these selected heads are functionally important for maintaining the CoT reasoning trajectory. \paragraph{No-CoT and Equation Prompting.} In both the one-shot No-CoT and zero-shot Equation settings, targeted ablations generally produce larger accuracy drops than the Random $n$-Head Selection baseline. In the No-CoT setting, accuracies decrease to $6.25\%$--$12.50\%$, compared with $15.60\%$ under random head selection. Similarly, in the Equation-only setting, accuracies fall to $6.25\%$--$18.75\%$, whereas the random baseline remains virtually unchanged at $34.40\%$. These results indicate that the selected heads are not specific to the explicit CoT condition: even outside the CoT setting, they contribute to direct-answer reasoning and arithmetic computation. However, the results obtained from ablations based on Random-Activation Sequential Multi-Head Patching remain relatively close to those obtained from the other sequential patching variants. We attribute this proximity primarily to our current experimental constraints: the evaluation is conducted on only 32 instances, and the head-selection hyperparameter is restricted to $k=3$ per POS category. With such a narrow selection budget, the identified heads may capture only a partial subset of the relevant computational components. In the Equation-only setting, $k=3$ may be insufficient to capture the broader calculation circuit. In the No-CoT setting, this limitation may be even stronger, since direct-answer inference can depend on both abstraction and calculation mechanisms, as suggested by prior work on mathematical reasoning circuits \citep{cheng2025llmsreasonabstractlymath}. Consequently, the effects of the targeted ablations may overlap with those of the random-activation-derived head sets, limiting the separation between these conditions. Despite these constraints in the No-CoT and Equation settings, the particularly strong ablation effect observed in the CoT condition supports the hypothesis that our methodology identifies components that are important for sustaining step-by-step reasoning under CoT prompting. To better characterize this degradation beyond aggregate accuracy, we provide a qualitative breakdown of resulting failure modes---including final-anchor loss, exemplar contamination, and numerical corruption---in Appendix~\ref{app:zero_ablation_qualitative}.

\section{Discussion and Conclusion} Our results support a mechanistic view of CoT prompting in which CoT does not merely add explanatory text, but helps steer the model's internal computation toward a reasoning-compatible trajectory. Sequential activation patching moves No-CoT executions toward CoT behavior, suggesting that CoT-conditioned activations causally contribute to final-answer computation in our setting. However, this does not imply that every generated rationale token is a fully faithful explanation of the model's internal reasoning. Rather, CoT appears to function as a useful internal control signal whose textual faithfulness may remain partial.

The same-question and cross-question patching results refine this interpretation. Stronger recovery under same-question patching indicates that part of the recovered computation is instance-sensitive, while the non-trivial gains under cross-question patching suggest that the discovered heads are not merely copying problem-specific operands or entities. 

The need for joint multi-head interventions and the zero-ablation results further argue against a localized single-head account of CoT. Ablating the selected heads disrupts not only final-answer accuracy, but also final-answer anchoring, exemplar--target separation, answer-slot binding, and numerical generation. We therefore interpret the identified structure as a distributed reasoning-support sub-circuit rather than a pure reasoning-only circuit. Overall, our sequential activation patching framework provides evidence that CoT-related computation in the evaluated model on SVAMP relies on coordinated attention-head activity that supports both reasoning control and auxiliary mechanisms required for successful answer generation. 

\section{Limitations} One limitation of our work is that the analysis depends on how CoT-generated tokens are grouped before aggregation. We use POS tags as a practical labeling scheme, but alternative groupings, such as semantic-role labels or task-specific arithmetic labels, could cluster the reasoning tokens differently and lead to different head rankings. Second, our head-selection procedure relies on a fixed top-$k$ budget. Due to computational constraints, we do not perform a sensitivity analysis over different $k$ values, nor do we assign separate label-specific budgets. Therefore, the stability of the identified head sets under alternative selection policies remains an open question. 

Another limitation is the scope of our empirical setting. Our experiments focus on mathematical word problems and one relatively small base model, so the extent to which the identified reasoning-support sub-circuits generalize to larger models, different architectures, specialized reasoning models, or non-mathematical reasoning tasks remains open. A further limitation is that our intervention metrics are computed at the first token of the final-answer span. This design provides a controlled and aligned decision point across CoT, No-CoT, and patched executions, but it does not fully evaluate the complete generated answer sequence. 

Finally, the discovered heads should not be interpreted as a pure reasoning-only circuit. The ablation results suggest that these heads may also support related functions such as target-question grounding, exemplar separation, answer-slot binding, CoT formatting, and numerical generation. Thus, we interpret the identified structure as a distributed reasoning-support sub-circuit consisting of overlapping computational components that are dynamically recruited via CoT, rather than a complete or isolated reasoning circuit. Crucially, while our results suggest that these components are functionally important for CoT performance and less effectively recruited in No-CoT settings, we currently lack the granularity to isolate the distinct functional roles of individual heads within this overlapping ensemble. Determining which specific subsets of heads are responsible for abstraction, calculation, or other sub-tasks therefore remains an open challenge. 

A promising direction for future work is to use the identified heads as a mechanistic prior for parameter-efficient fine-tuning. Since our interventions suggest that these heads participate in CoT-related reasoning-support computation, future studies could restrict PEFT modules, such as head-specific adapters or LoRA updates, to the corresponding attention-head subspaces instead of fine-tuning the entire model. Such a mechanistically informed PEFT strategy could reduce the number of trainable parameters while making the adaptation process more targeted and interpretable. However, because our ablation results indicate that the selected heads support overlapping functions such as answer anchoring, exemplar--target separation, answer-slot binding, and numerical generation, future work should first disentangle which subsets of heads are responsible for specific computational roles before treating them as isolated reasoning modules. 

\section{Ethics Statement} During the preparation of this manuscript, the authors utilized generative AI tools to assist in improving the linguistic quality, grammar, and overall readability of the text. AI tools were also employed to assist in the design and generation of figures and tables, facilitating the clear presentation of technical data. All AI-assisted outputs were subsequently reviewed, edited, and revised by the authors to ensure the technical accuracy of the arguments, the validity of the claims, and the precision of the visualized data. The authors certify that the final content, including all figures and tables, reflects their original research findings and intellectual contributions. The authors take full responsibility for the content of this manuscript and confirm that no original research data were generated or manipulated by AI tools.

\bibliographystyle{plainnat}
\bibliography{custom}

\clearpage
\appendix
\raggedbottom

\section{Algorithms}
\label{app:algorithms}

\subsection{Sequential Multi-Head Cross Patching}

In our preceding experiments, the source activations and the destination
execution both originated from the same target question (i.e., patching CoT
activations of Problem $A$ into the No-CoT execution of Problem $A$). While
this same-question patching effectively identifies components associated with
the reasoning trajectory, it introduces a potential confounding variable:
it remains ambiguous whether the identified sub-circuits transfer a more
general reasoning-support signal or simply carry problem-specific semantic
information, such as particular numbers, entities, or operands unique to
Problem $A$.

To decouple problem-specific content from the more general Chain-of-Thought
mechanism, we introduce Sequential Multi-Head Cross Patching
(Appendix~\ref{app:algorithms}, Algorithm~\ref{alg:cross_patching}).
In this configuration, for a given target question ($Q_A$), we randomly
sample a distinct source question ($Q_B$) from the dataset. We then extract
the POS-guided, top-performing attention-head activations from the clean CoT
execution of $Q_B$ and jointly inject them into the corrupted No-CoT execution
of $Q_A$. The resulting patched distribution is still evaluated against the
clean reference distribution of the target question $Q_A$.

This setup serves as a control for problem-specific information. If the
patched attention heads predominantly encode local, problem-specific
information, injecting activations from $Q_B$ into $Q_A$ should introduce
semantic mismatch and reduce recovery toward the target CoT distribution.
Conversely, if cross-question patching still yields recovery, this suggests
that at least part of the selected activation structure transfers across
instances rather than depending exclusively on problem-specific content.
Cross-question recovery therefore provides evidence that the identified
distributed components carry partially reusable reasoning-support signals
beyond the source problem's specific tokens.

\begin{algorithm}[H]
\small
\caption{Sequential Activation Patching and POS-Guided Aggregation}
\label{alg:patching_collection_aggregation}

\begin{algorithmic}[1]

\REQUIRE
Target question $q$,
source question $q'$,
CoT prompt $\mathcal{P}_{\mathrm{CoT}}$,
No-CoT prompt $\mathcal{P}_{\mathrm{NoCoT}}$,
model $M$,
POS tagger $\mathcal{T}$,
top-$k$ parameter

\STATE Run $M(q, \mathcal{P}_{\mathrm{CoT}})$
and store $D_{\mathrm{CoT}}$
and $A_{\mathrm{CoT}}^{q}$

\STATE Run $M(q, \mathcal{P}_{\mathrm{NoCoT}})$
and store $D_{\mathrm{NoCoT}}$
and $A_{\mathrm{NoCoT}}^{q}$

\STATE Run $M(q', \mathcal{P}_{\mathrm{CoT}})$
and obtain CoT tokens
$Y = (y_1, \dots, y_T)$

\STATE Store source activations
$A_{\mathrm{CoT}}^{q'}$
for all token positions, layers, and heads

\STATE Assign POS tags:
$p_t \gets \mathcal{T}(y_t)$
for $t = 1,\dots,T$

\STATE Compute baseline divergence once:

\STATE \hspace{1em}
$\mathrm{JSD}_{\mathrm{base}}
\gets
\mathrm{JSD}
(D_{\mathrm{CoT}}, D_{\mathrm{NoCoT}})$

\STATE Initialize
$\mathcal{R} \gets \emptyset$

\FOR{each token position $t$}

    \FOR{each layer $\ell$}

        \FOR{each head $h$}

            \STATE Patch:

            \STATE \hspace{1em}
            $A_{\mathrm{NoCoT}}^{q}[\ell,h]
            \leftarrow
            A_{\mathrm{CoT}}^{q'}[t,\ell,h]$

            \STATE Run forward pass
            $\rightarrow
            D_{\mathrm{patch}}^{(t,\ell,h)}$

            \STATE Compute:

            \STATE \hspace{1em}
            $\mathrm{JSD}_{\mathrm{patch}}^{(t,\ell,h)}$

            \STATE \hspace{1em}
            $\mathrm{MRR}^{(t,\ell,h)}$

            \STATE Store:

            \STATE \hspace{1em}
            $(p_t,
            t,
            \ell,
            h,
            \mathrm{JSD}_{\mathrm{patch}}^{(t,\ell,h)},
            \mathrm{MRR}^{(t,\ell,h)})
            \rightarrow
            \mathcal{R}$

        \ENDFOR

    \ENDFOR

\ENDFOR

\vspace{0.5em}

\STATE \textbf{POS-Guided Aggregation:}

\STATE Initialize
$\mathcal{H} \gets \emptyset$

\FOR{each unique POS tag $p$ in $Y$}

    \STATE
    $\mathcal{R}_p
    \gets$
    subset of records in $\mathcal{R}$
    with POS tag $p$

    \FOR{each head $(\ell,h)$}

        \STATE Compute
        $\mu_{\mathrm{JSD}}
        \gets$
        mean $\mathrm{JSD}_{\mathrm{patch}}$
        over $\mathcal{R}_p$

        \STATE Compute
        $\mu_{\mathrm{MRR}}
        \gets$
        mean $\mathrm{MRR}$
        over $\mathcal{R}_p$

    \ENDFOR

    \STATE
    $\mathcal{H}_p
    \gets$
    select top-$k$ heads for $p$
    according to the chosen metric

    \STATE Store
    $\mathcal{H}_p$
    in $\mathcal{H}$

\ENDFOR

\RETURN
$\mathcal{R}, \mathcal{H}$

\end{algorithmic}
\end{algorithm}

\begin{algorithm}[H]
\small
\caption{Sequential Multi-Head Patching}
\label{alg:head_selection}

\begingroup
\renewcommand{\baselinestretch}{0.85}

\begin{algorithmic}[1]

\REQUIRE
Record set $\mathcal{R}$,
No-CoT run,
CoT reference distribution $D_{\mathrm{CoT}}$,
top-$k$ parameter

\FOR{each POS tag $p$}

    \STATE Collect all records in
    $\mathcal{R}$
    with POS tag $p$

    \STATE Rank candidate heads
    according to the chosen patching metric

    \STATE \hspace{1em}
    (e.g., lower
    $\mathrm{JSD}_{\mathrm{patch}}$
    or higher $\mathrm{MRR}$)

    \STATE Select the top-$k$ heads
    for POS tag $p$:

    \STATE \hspace{1em}
    $\mathcal{H}_p
    \gets
    \{
    (\ell_1,h_1),
    (\ell_2,h_2),
    \dots,
    (\ell_k,h_k)
    \}$

\ENDFOR

\STATE Initialize global selected-head set
$\mathcal{G} \gets \emptyset$

\FOR{each POS tag $p$}

    \FOR{each head
    $(\ell,h) \in \mathcal{H}_p$}

        \IF{$(\ell,h) \notin \mathcal{G}$}

            \STATE Add
            $((\ell,h),p)$
            to $\mathcal{G}$

        \ELSE

            \STATE Compare the existing
            POS assignment
            with the new POS assignment

            \STATE Retain the assignment
            with the better metric value

        \ENDIF

    \ENDFOR

\ENDFOR

\STATE Extract the clean CoT activation
associated with each
$((\ell,h),p) \in \mathcal{G}$

\STATE Jointly patch all uniquely assigned heads
in $\mathcal{G}$
into the No-CoT run

\STATE Run one patched forward pass
and obtain
$D_{\mathrm{patch}}^{\mathrm{multi}}$

\STATE Compute:

\STATE \hspace{1em}
$\mathrm{JSD}^{\mathrm{multi}}
\gets
\mathrm{JSD}
(
D_{\mathrm{CoT}},
D_{\mathrm{patch}}^{\mathrm{multi}}
)$

\STATE \hspace{1em}
$\mathrm{MRR}^{\mathrm{multi}}$

\RETURN
$\mathcal{G}$
and the multi-head metric results

\end{algorithmic}

\endgroup
\end{algorithm}

\subsection{Random-Activation Sequential Multi-Head Patching}

To assess whether the observed effects can be explained by arbitrary internal perturbations, we establish a Random-Activation Sequential Multi-Head Patching. In mechanistic interpretability, intervening on a model's hidden states can sometimes inadvertently disrupt its default computational graph. This disruption might lead to unexpected distributional shifts that could be falsely interpreted as meaningful causal effects, raising the question of whether the model is actually utilizing CoT information or just reacting to random noise.

To rigorously rule out this null hypothesis, we replicate our multi-head patching procedure using randomly generated activation vectors in place of the actual CoT representations. As detailed in Algorithm \ref{alg:random_patching}, for each attention head, we generate a random activation tensor and inject it into the corrupted No-CoT run. We then evaluate these random interventions, rank the heads based on their accidental patching quality, and perform a joint multi-head patch using the top-$n$ random activations. 

\begin{algorithm}[H] \small \caption{Sequential Multi-Head Cross Patching} \label{alg:cross_patching} \begin{algorithmic}[1] \REQUIRE Target question $Q_A$, source question $Q_B$, CoT prompt $\mathcal{P}_{\mathrm{CoT}}$, No-CoT prompt $\mathcal{P}_{\mathrm{NoCoT}}$, model $M$, POS tagger $\mathcal{T}$, top-$k$ parameter \STATE Run the clean target execution $M(Q_A,\mathcal{P}_{\mathrm{CoT}})$ and store $D_{\mathrm{CoT}}^{A}$ \STATE Run the corrupted target execution $M(Q_A,\mathcal{P}_{\mathrm{NoCoT}})$ and store $D_{\mathrm{NoCoT}}^{A}$ and $A_{\mathrm{NoCoT}}^{A}$ \STATE Run $M(Q_B,\mathcal{P}_{\mathrm{CoT}})$ and obtain source CoT tokens $Y^{B} = (y_1,\dots,y_T)$ \STATE Store source activations $A_{\mathrm{CoT}}^{B}$ for all token positions, layers, and heads \STATE Assign POS tags: $p_t \gets \mathcal{T}(y_t)$ for each $y_t \in Y^{B}$ \STATE Initialize record set $\mathcal{R} \gets \emptyset$ \STATE Initialize global assignment map $\mathcal{G} \gets \emptyset$ \FOR{each token position $t$} \FOR{each layer $\ell$} \FOR{each head $h$} \STATE Patch: \STATE \hspace{1em} $A_{\mathrm{NoCoT}}^{A}[\ell,h] \leftarrow A_{\mathrm{CoT}}^{B}[t,\ell,h]$ \STATE Run the patched forward pass and obtain $D_{\mathrm{patch}}^{(t,\ell,h)}$ \STATE Compute: \STATE \hspace{1em} $\mathrm{JSD}_{\mathrm{patch}}^{(t,\ell,h)} \gets \mathrm{JSD} \left( D_{\mathrm{CoT}}^{A}, D_{\mathrm{patch}}^{(t,\ell,h)} \right)$ \STATE \hspace{1em} $\mathrm{MRR}^{(t,\ell,h)}$ \STATE Store: \STATE \hspace{1em} $(p_t, t, \ell, h, \mathrm{JSD}_{\mathrm{patch}}^{(t,\ell,h)}, \mathrm{MRR}^{(t,\ell,h)}) \rightarrow \mathcal{R}$ \ENDFOR \ENDFOR \ENDFOR \FOR{each unique POS tag $p$ in $Y^{B}$} \STATE $\mathcal{R}_p \gets$ subset of records in $\mathcal{R}$ with POS tag $p$ \STATE Compute the mean patching metric for each candidate head $(\ell,h)$ within $\mathcal{R}_p$ \STATE Select the top-$k$ heads $\mathcal{H}_p$ according to the chosen metric \FOR{each head $(\ell,h) \in \mathcal{H}_p$} \IF{$(\ell,h)$ is not assigned in $\mathcal{G}$} \STATE Assign $((\ell,h),p)$ to $\mathcal{G}$ \ELSE \STATE Compare the existing POS assignment with the new POS assignment \STATE Retain the assignment with the better metric value \ENDIF \ENDFOR \ENDFOR \STATE Let $\widehat{\mathcal{H}} \gets \mathrm{keys}(\mathcal{G})$ denote the final deduplicated head set \STATE Extract the source CoT activation associated with each $((\ell,h),p) \in \mathcal{G}$ \STATE Jointly patch all heads in $\widehat{\mathcal{H}}$ into the target No-CoT run \STATE Run one joint patched forward pass and obtain $D_{\mathrm{patch}}^{\mathrm{cross}}$ \STATE Compute: \STATE \hspace{1em} $\mathrm{JSD}^{\mathrm{cross}} \gets \mathrm{JSD} \left( D_{\mathrm{CoT}}^{A}, D_{\mathrm{patch}}^{\mathrm{cross}} \right)$ \STATE \hspace{1em} $\mathrm{MRR}^{\mathrm{cross}}$ \RETURN $\mathcal{G}$, $\widehat{\mathcal{H}}$, $\mathrm{JSD}^{\mathrm{cross}}$, and $\mathrm{MRR}^{\mathrm{cross}}$ \end{algorithmic} \end{algorithm} \begin{algorithm}[H] \small \caption{Random-Activation Sequential Multi-Head Patching} \label{alg:random_patching} \begin{algorithmic}[1] \REQUIRE Target question $q$, CoT prompt $\mathcal{P}_{\mathrm{CoT}}$, No-CoT prompt $\mathcal{P}_{\mathrm{NoCoT}}$, model $M$, integer $n$ \STATE Run $M(q,\mathcal{P}_{\mathrm{CoT}})$ and store output distribution $D_{\mathrm{CoT}}$ \STATE Run $M(q,\mathcal{P}_{\mathrm{NoCoT}})$ and store output distribution $D_{\mathrm{NoCoT}}$ and activations $A_{\mathrm{NoCoT}}^{q}$ \STATE Compute baseline divergence once: \STATE \hspace{1em} $\mathrm{JSD}_{\mathrm{base}} \gets \mathrm{JSD} \left( D_{\mathrm{CoT}}, D_{\mathrm{NoCoT}} \right)$ \STATE Initialize result set $\mathcal{R}_{\mathrm{rand}} \gets \emptyset$ \STATE Initialize random activation table $\mathcal{Z} \gets \emptyset$ \FOR{each layer $\ell$} \FOR{each head $h$} \STATE Generate a random activation $z_{\ell,h}$ for head $(\ell,h)$ \STATE Store $z_{\ell,h}$ in $\mathcal{Z}$ \STATE Patch the No-CoT run with $z_{\ell,h}$ at head $(\ell,h)$ \STATE Run the patched forward pass and obtain $D_{\mathrm{patch}}^{(\ell,h)}$ \STATE Compute: \STATE \hspace{1em} $\mathrm{JSD}_{\mathrm{patch}}^{(\ell,h)} \gets \mathrm{JSD} \left( D_{\mathrm{CoT}}, D_{\mathrm{patch}}^{(\ell,h)} \right)$ \STATE \hspace{1em} $\mathrm{MRR}^{(\ell,h)}$ \STATE Store: \STATE \hspace{1em} $(\ell, h, \mathrm{JSD}_{\mathrm{patch}}^{(\ell,h)}, \mathrm{MRR}^{(\ell,h)}) \rightarrow \mathcal{R}_{\mathrm{rand}}$ \ENDFOR \ENDFOR \STATE Rank all heads in $\mathcal{R}_{\mathrm{rand}}$ according to the chosen patching metric \STATE \hspace{1em} (e.g., lower $\mathrm{JSD}_{\mathrm{patch}}$ or higher $\mathrm{MRR}$) \STATE Select the top-$n$ heads: \STATE \hspace{1em} $\mathcal{H}_{\mathrm{top}\text{-}n} \gets \{ (\ell_1,h_1), \dots, (\ell_n,h_n) \}$ \STATE Jointly patch the No-CoT run using the stored random activations: \STATE \hspace{1em} $\{ z_{\ell,h} : (\ell,h) \in \mathcal{H}_{\mathrm{top}\text{-}n} \}$ \STATE Run one joint patched forward pass and obtain $D_{\mathrm{patch}}^{\mathrm{top}\text{-}n}$ \STATE Compute: \STATE \hspace{1em} $\mathrm{JSD}^{\mathrm{top}\text{-}n} \gets \mathrm{JSD} \left( D_{\mathrm{CoT}}, D_{\mathrm{patch}}^{\mathrm{top}\text{-}n} \right)$ \STATE \hspace{1em} $\mathrm{MRR}^{\mathrm{top}\text{-}n}$ \RETURN $\mathcal{H}_{\mathrm{top}\text{-}n}$, $\mathcal{R}_{\mathrm{rand}}$, $D_{\mathrm{patch}}^{\mathrm{top}\text{-}n}$, $\mathrm{JSD}^{\mathrm{top}\text{-}n}$, and $\mathrm{MRR}^{\mathrm{top}\text{-}n}$ \end{algorithmic} \end{algorithm} 

\begin{algorithm}[H]
\small
\caption{Zero Ablation of Patching-Selected Heads}
\label{alg:zero_ablation}

\begin{algorithmic}[1]

\REQUIRE
Dataset $\mathcal{D}$,
CoT prompt $\mathcal{P}_{\mathrm{CoT}}$,
No-CoT prompt $\mathcal{P}_{\mathrm{NoCoT}}$,
direct-equation prompt $\mathcal{P}_{\mathrm{Eq}}$,
model $M$,
selected-head set $\mathcal{H}_{\mathrm{sel}}$

\STATE Initialize No-CoT results
$\mathcal{R}_{\mathrm{NoCoT}} \gets \emptyset$

\STATE Initialize CoT results
$\mathcal{R}_{\mathrm{CoT}} \gets \emptyset$

\STATE Initialize direct-equation results
$\mathcal{R}_{\mathrm{Eq}} \gets \emptyset$

\FOR{each question $q \in \mathcal{D}$}

    \STATE Obtain patching-selected heads
    $\mathcal{H}_{\mathrm{sel}}^{q}$
    for question $q$

    \STATE Sample matched random heads
    $\mathcal{H}_{\mathrm{rand}}^{q}$
    such that

    \STATE \hspace{1em}
    $
    |\mathcal{H}_{\mathrm{rand}}^{q}|
    =
    |\mathcal{H}_{\mathrm{sel}}^{q}|
    $

    \vspace{0.5em}
    \STATE \textbf{Direct-Equation Ablation:}

    \STATE Construct the equation-only input
    $e_q$
    corresponding to question $q$

    \STATE Run
    $M(e_q,\mathcal{P}_{\mathrm{Eq}})$
    and store the original equation prediction
    $\hat{y}_{\mathrm{Eq}}$

    \STATE Zero-ablate all heads in
    $\mathcal{H}_{\mathrm{sel}}^{q}$
    during direct-equation inference

    \STATE Run the selected-head-ablated
    equation forward pass and obtain
    $\hat{y}_{\mathrm{Eq}}^{\mathrm{sel}}$

    \STATE Zero-ablate all heads in
    $\mathcal{H}_{\mathrm{rand}}^{q}$
    during direct-equation inference

    \STATE Run the random-head-ablated
    equation forward pass and obtain
    $\hat{y}_{\mathrm{Eq}}^{\mathrm{rand}}$

    \STATE Store:
    $
    (q,
    \hat{y}_{\mathrm{Eq}},
    \hat{y}_{\mathrm{Eq}}^{\mathrm{sel}},
    \hat{y}_{\mathrm{Eq}}^{\mathrm{rand}})
    \rightarrow
    \mathcal{R}_{\mathrm{Eq}}
    $

    \vspace{0.5em}
    \STATE \textbf{No-CoT Ablation:}

    \STATE Run
    $M(q,\mathcal{P}_{\mathrm{NoCoT}})$
    and store the original No-CoT prediction
    $\hat{y}_{\mathrm{NoCoT}}$

    \STATE Zero-ablate all heads in
    $\mathcal{H}_{\mathrm{sel}}^{q}$
    during No-CoT inference

    \STATE Run the selected-head-ablated
    No-CoT forward pass and obtain
    $\hat{y}_{\mathrm{NoCoT}}^{\mathrm{sel}}$

    \STATE Zero-ablate all heads in
    $\mathcal{H}_{\mathrm{rand}}^{q}$
    during No-CoT inference

    \STATE Run the random-head-ablated
    No-CoT forward pass and obtain
    $\hat{y}_{\mathrm{NoCoT}}^{\mathrm{rand}}$

    \STATE Store:
    $
    (q,
    \hat{y}_{\mathrm{NoCoT}},
    \hat{y}_{\mathrm{NoCoT}}^{\mathrm{sel}},
    \hat{y}_{\mathrm{NoCoT}}^{\mathrm{rand}})
    \rightarrow
    \mathcal{R}_{\mathrm{NoCoT}}
    $

    \vspace{0.5em}
    \STATE \textbf{CoT Ablation:}

    \STATE Run
    $M(q,\mathcal{P}_{\mathrm{CoT}})$
    and store the original CoT prediction
    $\hat{y}_{\mathrm{CoT}}$

    \STATE Generate the CoT token sequence:
    $
    Y = (y_1,y_2,\dots,y_T)
    $

    \FOR{each generated CoT token $y_t \in Y$}

        \STATE Obtain token-specific selected heads
        $\mathcal{H}_{\mathrm{sel}}^{q,t}$

        \STATE Zero-ablate all heads in
        $\mathcal{H}_{\mathrm{sel}}^{q,t}$
        during generation of token $y_t$

    \ENDFOR

    \STATE Store the selected-head-ablated
    CoT prediction
    $\hat{y}_{\mathrm{CoT}}^{\mathrm{sel}}$

    \FOR{each generated CoT token $y_t \in Y$}

        \STATE Sample matched random heads
        $\mathcal{H}_{\mathrm{rand}}^{q,t}$
        such that

        \STATE
        $
        |\mathcal{H}_{\mathrm{rand}}^{q,t}|
        =
        |\mathcal{H}_{\mathrm{sel}}^{q,t}|
        $

        \STATE Zero-ablate all heads in
        $\mathcal{H}_{\mathrm{rand}}^{q,t}$
        during generation of token $y_t$

    \ENDFOR

    \STATE Store the random-head-ablated
    CoT prediction
    $\hat{y}_{\mathrm{CoT}}^{\mathrm{rand}}$

    \STATE Store:
    $
    (q,
    \hat{y}_{\mathrm{CoT}},
    \hat{y}_{\mathrm{CoT}}^{\mathrm{sel}},
    \hat{y}_{\mathrm{CoT}}^{\mathrm{rand}})
    \rightarrow
    \mathcal{R}_{\mathrm{CoT}}
    $

\ENDFOR

\vspace{0.5em}

\STATE Compute original accuracies for
Direct-Equation, No-CoT, and CoT prompting

\STATE Compute selected-head ablation accuracies
for all three prompting conditions

\STATE Compute random-head ablation accuracies
for all three prompting conditions

\RETURN
$\mathcal{R}_{\mathrm{Eq}}$,
$\mathcal{R}_{\mathrm{NoCoT}}$,
$\mathcal{R}_{\mathrm{CoT}}$,
original accuracies,
selected-head ablation accuracies,
and random-head ablation accuracies

\end{algorithmic}
\end{algorithm}


\begin{figure}[htbp]
    \centering
    \includegraphics[
        width=\textwidth
    ]{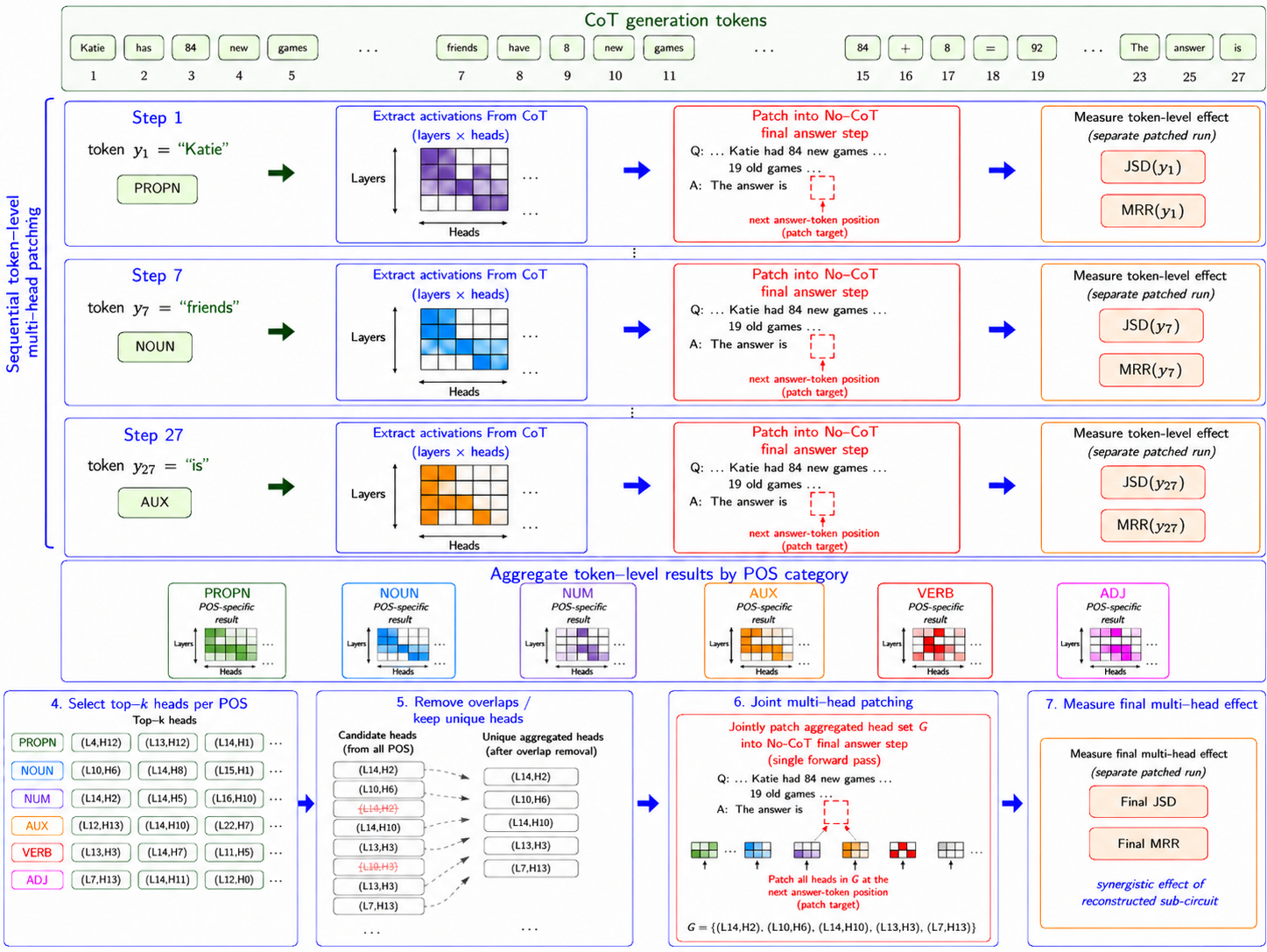}

    \caption{\small
    Overview of the Sequential Multi-Head Patching and deduplication procedure.
    First, the top-$k$ most influential attention heads are selected for each
    individual Part-of-Speech (POS) category based on their isolated patching
    performance. These category-specific heads are then merged into a single
    global multi-head set. To resolve intervention conflicts, a deduplication
    mechanism is applied such that any head appearing in multiple POS categories
    is assigned exclusively to its best-performing functional role. Finally,
    the activations associated with the resulting global set are jointly patched
    into the corrupted No-CoT execution at the final-answer position, and the
    resulting recovery is quantified using JSD and MRR.
    }
    \label{fig:multi_head_patching_appendix}
\end{figure}

\FloatBarrier

\section{Evaluation Metrics Details} \label{sec:metrics_details} In our activation patching experiments, we use two primary metrics to quantify the causal effect of interventions on attention-head activations: Jensen--Shannon Divergence (JSD) and Margin Recovery Ratio (MRR). This section provides the formal definitions and interpretation guidelines for both metrics. \subsection{Jensen--Shannon Divergence (JSD)} \label{sec:jsd_details} We use Jensen--Shannon Divergence (JSD) to measure the divergence between the next-token probability distribution of the patched model, $P_{\text{patch}}$, and that of the clean Chain-of-Thought (CoT) execution, $P_{\text{CoT}}$. JSD is a symmetric and smoothed variant of the Kullback--Leibler (KL) divergence, defined as: \[ \mathrm{JSD} \left( P_{\text{patch}}, P_{\text{CoT}} \right) = \frac{1}{2} \mathrm{KL} \left( P_{\text{patch}} \parallel M \right) + \frac{1}{2} \mathrm{KL} \left( P_{\text{CoT}} \parallel M \right), \] where \[ M = \frac{1}{2} \left( P_{\text{patch}} + P_{\text{CoT}} \right) \] is the pointwise mixture distribution, and $\mathrm{KL}$ denotes the standard Kullback--Leibler divergence. \textbf{Interpretation of the JSD Scale:} Unlike KL divergence, JSD is bounded. Because we compute the KL divergence using the natural logarithm (base $e$), the JSD values in our experiments lie within the range $[0,\ln 2]$, where $\ln 2 \approx 0.693$. \begin{itemize} \item $\mathrm{JSD} = 0$ indicates that the patched and clean CoT next-token distributions are identical at the evaluated decoding position. \item Higher JSD values, approaching the theoretical maximum of $\ln 2 \approx 0.693$, indicate greater divergence between the patched and clean CoT distributions. Accordingly, lower JSD values indicate that the intervention moves the model's next-token distribution closer to the clean CoT reference. \end{itemize} \subsection{Margin Recovery Ratio (MRR)} \label{sec:mrr_details} To capture decision-level effects, we define a logit-based margin metric. Let $t^*$ denote the target answer token at the evaluated decoding position, and let $z(t)$ denote the logit assigned to token $t$. The margin $M$ is defined as: \[ M = z(t^*) - \max_{t \neq t^*} z(t). \] We compute this margin for the clean CoT reference, $M_{\text{CoT}}$, the corrupted No-CoT baseline, $M_{\text{NoCoT}}$, and the patched execution, $M_{\text{patch}}$. The Margin Recovery Ratio (MRR) is then defined as: \[ \mathrm{MRR} = \frac{ M_{\text{patch}} - M_{\text{NoCoT}} }{ M_{\text{CoT}} - M_{\text{NoCoT}} }. \] The Margin Recovery Ratio is defined when $M_{\text{CoT}} \neq M_{\text{NoCoT}}$. This ensures that the denominator represents a non-zero difference between the clean CoT and No-CoT decision margins. When $M_{\text{CoT}} = M_{\text{NoCoT}}$, the ratio is undefined because there is no margin difference between the two reference conditions to recover. \textbf{Interpretation of the MRR Scale:} MRR measures how much of the CoT--No-CoT margin difference is recovered by the patching intervention. Its values are interpreted as follows: \begin{itemize} \item $\mathrm{MRR} < 0$: The patched execution moves the target-token margin in the opposite direction from the clean CoT reference relative to the No-CoT baseline. \item $\mathrm{MRR} = 0$: The patched execution recovers none of the CoT--No-CoT margin difference. \item $0 < \mathrm{MRR} < 1$: The intervention partially recovers the CoT--No-CoT margin difference. \item $\mathrm{MRR} = 1$: The patched execution matches the clean CoT decision margin. \item $\mathrm{MRR} > 1$: The patched execution exceeds the clean CoT decision margin in the direction of the target answer token. \end{itemize}

\section{Sequential Multi-Head Patching Results}
\label{sec:patching_results}
\begin{table*}[t]

\centering

\tiny

\renewcommand{\arraystretch}{0.9}

\setlength{\tabcolsep}{1.5pt}

\begin{tabularx}{\textwidth}{l l c c r @{ $\rightarrow$ } l r @{ $\rightarrow$ } l c}

\toprule

\textbf{ID} & \textbf{Type} & \textbf{Eq} & \textbf{Stat} & \multicolumn{2}{c}{\textbf{Logit (No $\rightarrow$ Pat.)}} & \multicolumn{2}{c}{\textbf{Prob (No $\rightarrow$ Pat.)}} & \textbf{JSD} \\

\midrule

\texttt{chal-961} & Addition & 1 & $\times\times$ & 23.61 & 28.68 & 36.2\% & 97.8\% & 0.0072 \\

\multicolumn{9}{p{0.98\textwidth}}{\quad \scriptsize \textit{Heads:} \texttt{L16H13[NNP], L16H10[CD], L13H12[:], L13H0[NNP], L14H5[VBZ], L14H10[JJ], L12H13[VBZ], L14H13[.], L16H8[.], L14H1[NN], L13H4[NN], L12H0[JJ], L15H1[NNS], L13H7[PRP\$]}} \\

\specialrule{0.05em}{0.1em}{0.1em}

\texttt{chal-569} & Addition & 1 & $\checkmark\times$ & 21.87 & 22.13 & 27.7\% & 21.9\% & 0.4047 \\

\multicolumn{9}{p{0.98\textwidth}}{\quad \scriptsize \textit{Heads:} \texttt{L10H7[:], L13H12[.], L8H7[:], L14H5[IN], L10H6[NN], L14H8[NN], L8H3[VBD], L10H12[JJ], L16H10[TO], L14H4[JJ], L9H2[.], L14H2[CD], L10H3[CD], L13H10[NNS], L9H1[IN], L10H9[IN], L13H13[CC], L13H4[DT], L13H1[DT], L12H4[VBZ], L14H10[VBZ], L11H5[VBZ]}} \\

\specialrule{0.05em}{0.1em}{0.1em}

\texttt{chal-290} & Addition & 1 & $\times\times$ & 22.23 & 22.82 & 28.2\% & 40.6\% & 0.2662 \\

\multicolumn{9}{p{0.98\textwidth}}{\quad \scriptsize \textit{Heads:} \texttt{L13H12[VBD], L20H10[IN], L15H13[:], L14H10[RB], L14H7[CD], L20H12[.], L14H13[.], L14H5[NNS], L13H13[DT], L12H12[TO], L12H6[,], L17H8[VBZ]}} \\

\specialrule{0.05em}{0.1em}{0.1em}

\texttt{chal-821} & Addition & 1 & $\checkmark\times$ & 22.63 & 25.13 & 45.4\% & 85.5\% & 0.0525 \\

\multicolumn{9}{p{0.98\textwidth}}{\quad \scriptsize \textit{Heads:} \texttt{L13H12[IN], L4H12[NNP], L17H7[:], L14H12[NNP], L14H2[CD], L14H5[CD], L14H11[CD], L12H10[.], L14H10[VBZ]}} \\

\specialrule{0.05em}{0.1em}{0.1em}

\texttt{chal-451} & Common-Divis & 1 & $\checkmark\times$ & 22.14 & 24.91 & 30.7\% & 91.5\% & 0.0285 \\

\multicolumn{9}{p{0.98\textwidth}}{\quad \scriptsize \textit{Heads:} \texttt{L11H3[VBZ], L4H12[:], L11H0[:], L10H6[CD], L12H13[NNS], L14H5[PRP], L13H3[IN], L11H5[NN], L14H7[CD], L15H13[CD], L10H12[.], L13H6[NNS], L10H7[CC], L13H10[CC], L12H0[NN], L14H8[DT], L12H8[DT]}} \\

\specialrule{0.05em}{0.1em}{0.1em}

\texttt{chal-970} & Common-Divis & 1 & $\checkmark\times$ & 21.38 & 23.09 & 16.4\% & 49.1\% & 0.2163 \\

\multicolumn{9}{p{0.98\textwidth}}{\quad \scriptsize \textit{Heads:} \texttt{L13H3[NN], L13H12[PRP\$], L16H7[CD], L12H13[CC], L14H10[CC], L15H1[NN], L22H7[VBZ], L14H5[VBZ], L16H11[.], L16H13[.], L14H13[.], L14H12[PRP\$]}} \\

\specialrule{0.05em}{0.1em}{0.1em}

\texttt{chal-215} & Common-Divis & 1 & $\checkmark\times$ & 22.61 & 19.05 & 80.6\% & 98.8\% & 0.0038 \\

\multicolumn{9}{p{0.98\textwidth}}{\quad \scriptsize \textit{Heads:} \texttt{L13H6[:], L13H12[DT], L23H1[NN], L15H6[IN], L14H8[VBZ], L16H13[.], L13H1[CD], L22H0[CD], L13H10[CD], L7H13[JJ], L18H9[VB]}} \\

\specialrule{0.05em}{0.1em}{0.1em}

\texttt{chal-160} & Common-Divis & 1 & $\times\times$ & 21.85 & 24.60 & 14.7\% & 79.0\% & 0.0664 \\

\multicolumn{9}{p{0.98\textwidth}}{\quad \scriptsize \textit{Heads:} \texttt{L10H6[:], L16H8[CC], L16H13[CD], L8H7[EX], L15H12[VBP], L14H10[CD], L22H7[.], L13H3[NN], L15H10[CC], L14H7[DT], L14H12[NN], L12H0[,], L13H10[,], L14H8[JJ], L8H3[VBN]}} \\

\specialrule{0.05em}{0.1em}{0.1em}

\texttt{chal-76} & Multiplicati & 1 & $\times\times$ & 23.77 & 20.01 & 88.3\% & 99.3\% & 0.0021 \\

\multicolumn{9}{p{0.98\textwidth}}{\quad \scriptsize \textit{Heads:} \texttt{L14H10[:], L17H7[PRP], L18H9[NNS], L23H1[NN], L18H3[DT], L14H7[.], L22H7[VBZ], L14H13[.], L14H9[.], L22H0[CD]}} \\

\specialrule{0.05em}{0.1em}{0.1em}

\texttt{chal-854} & Multiplicati & 1 & $\checkmark\times$ & 18.52 & 20.82 & 1.3\% & 4.5\% & 0.5954 \\

\multicolumn{9}{p{0.98\textwidth}}{\quad \scriptsize \textit{Heads:} \texttt{L12H0[CD], L13H8[:], L16H7[:], L14H10[NN], L15H13[NN], L14H13[VBZ], L13H3[VBZ], L10H6[CD], L16H12[DT]}} \\

\specialrule{0.05em}{0.1em}{0.1em}

\texttt{chal-196} & Multiplicati & 1 & $\times\times$ & 20.80 & 23.81 & 39.9\% & 59.1\% & 0.1636 \\

\multicolumn{9}{p{0.98\textwidth}}{\quad \scriptsize \textit{Heads:} \texttt{L14H10[:], L13H12[NN], L8H7[PRP\$], L9H2[NN], L12H0[NN], L13H1[NNS], L13H4[NNS], L16H7[.], L14H7[RB], L12H2[PRP\$], L9H4[,], L11H8[VB], L13H6[VBZ], L6H9[VBZ]}} \\

\specialrule{0.05em}{0.1em}{0.1em}

\texttt{chal-12} & Multiplicati & 1 & $\checkmark\checkmark$ & 23.75 & 23.88 & 75.2\% & 83.1\% & 0.0618 \\

\multicolumn{9}{p{0.98\textwidth}}{\quad \scriptsize \textit{Heads:} \texttt{L16H8[CD], L16H10[.], L13H12[VBZ], L13H1[NN], L20H8[CD], L16H7[CD], L14H1[NNS], L16H13[IN], L6H9[IN], L10H6[NN]}} \\

\specialrule{0.05em}{0.1em}{0.1em}

\texttt{chal-163} & Subtraction & 1 & $\times\times$ & 20.37 & 20.07 & 8.0\% & 2.6\% & 0.6303 \\

\multicolumn{9}{p{0.98\textwidth}}{\quad \scriptsize \textit{Heads:} \texttt{L16H10[IN], L16H6[DT], L15H5[NNS], L10H6[PRP], L14H13[NN], L14H2[CD], L16H8[.], L14H10[CD], L14H9[DT], L14H5[JJ], L14H3[RBR]}} \\

\specialrule{0.05em}{0.1em}{0.1em}

\texttt{chal-514} & Subtraction & 1 & $\times\times$ & 22.31 & 23.18 & 17.7\% & 80.1\% & 0.0697 \\

\multicolumn{9}{p{0.98\textwidth}}{\quad \scriptsize \textit{Heads:} \texttt{L12H0[VBZ], L14H1[RBR], L14H13[CD], L9H10[NN], L10H6[CD], L16H0[.], L16H11[.], L14H5[RBR]}} \\

\specialrule{0.05em}{0.1em}{0.1em}

\texttt{chal-954} & Subtraction & 1 & $\checkmark\times$ & 22.90 & 25.60 & 44.6\% & 96.2\% & 0.0124 \\

\multicolumn{9}{p{0.98\textwidth}}{\quad \scriptsize \textit{Heads:} \texttt{L14H10[.], L21H9[VBZ], L14H1[CD], L14H7[CD], L18H9[.], L11H12[CD], L9H1[NN], L13H9[NN], L2H0[NN], L22H7[VBZ], L16H6[DT], L11H0[VBZ]}} \\

\specialrule{0.05em}{0.1em}{0.1em}

\texttt{chal-83} & Subtraction & 1 & $\times\times$ & 20.75 & 21.81 & 6.6\% & 25.2\% & 0.3642 \\

\multicolumn{9}{p{0.98\textwidth}}{\quad \scriptsize \textit{Heads:} \texttt{L16H8[.], L22H7[.], L16H7[.], L10H6[PRP], L8H3[PRP], L12H0[CD], L16H11[NN], L20H9[NNS], L12H4[DT]}} \\

\specialrule{0.05em}{0.1em}{0.1em}

\texttt{chal-21} & Addition & 2 & $\checkmark\times$ & 21.14 & 22.81 & 25.9\% & 78.1\% & 0.0819 \\

\multicolumn{9}{p{0.98\textwidth}}{\quad \scriptsize \textit{Heads:} \texttt{L14H13[NNP], L14H1[.], L13H3[NNS], L15H3[CD], L14H10[VBZ], L22H7[.], L13H12[DT]}} \\

\specialrule{0.05em}{0.1em}{0.1em}

\texttt{chal-129} & Addition & 2 & $\checkmark\times$ & 21.62 & 22.65 & 17.5\% & 55.1\% & 0.1875 \\

\multicolumn{9}{p{0.98\textwidth}}{\quad \scriptsize \textit{Heads:} \texttt{L13H8[:], L12H0[NNS], L16H8[IN], L15H13[NN], L16H7[NNS], L14H7[.], L14H5[.], L14H13[.], L14H9[DT]}} \\

\specialrule{0.05em}{0.1em}{0.1em}

\texttt{chal-587} & Addition & 2 & $\times\times$ & 21.15 & 25.78 & 21.9\% & 91.3\% & 0.0294 \\

\multicolumn{9}{p{0.98\textwidth}}{\quad \scriptsize \textit{Heads:} \texttt{L12H2[.], L13H3[CD], L12H8[:], L10H7[VBD], L4H12[EX], L8H5[DT], L16H8[.], L14H13[.], L12H6[CD], L7H13[NN], L12H13[NN], L14H7[NNS], L10H3[IN], L10H11[VBZ], L22H7[VBZ]}} \\

\specialrule{0.05em}{0.1em}{0.1em}

\texttt{chal-254} & Addition & 2 & $\checkmark\times$ & 23.73 & 23.41 & 47.7\% & 48.3\% & 0.2230 \\

\multicolumn{9}{p{0.98\textwidth}}{\quad \scriptsize \textit{Heads:} \texttt{L13H3[:], L10H6[:], L14H13[:], L13H4[DT], L14H1[CD], L4H12[VBD], L13H7[VBD], L12H13[CD], L8H7[NNS], L9H1[IN], L17H7[PRP], L5H6[DT], L22H7[VBZ], L15H1[VBZ]}} \\

\specialrule{0.05em}{0.1em}{0.1em}

\texttt{chal-28} & Common-Divis & 2 & $\times\times$ & 18.66 & 20.06 & 1.3\% & 9.8\% & 0.4992 \\

\multicolumn{9}{p{0.98\textwidth}}{\quad \scriptsize \textit{Heads:} \texttt{L16H13[CD], L14H9[VBZ], L16H11[CD], L14H10[CD], L14H13[NN], L13H6[VBD], L14H7[CC], L15H9[.], L15H7[.], L13H1[NNS], L15H10[CC], L22H7[VBZ]}} \\

\specialrule{0.05em}{0.1em}{0.1em}

\texttt{chal-906} & Common-Divis & 2 & $\checkmark\times$ & 20.69 & 22.96 & 7.3\% & 57.5\% & 0.1740 \\

\multicolumn{9}{p{0.98\textwidth}}{\quad \scriptsize \textit{Heads:} \texttt{L8H7[POS], L15H6[IN], L3H8[:], L10H6[NN], L13H7[IN], L14H7[.], L13H3[CD], L15H13[VBZ], L16H10[.], L14H5[RB], L12H13[RB]}} \\

\specialrule{0.05em}{0.1em}{0.1em}

\texttt{chal-829} & Common-Divis & 2 & $\checkmark\times$ & 22.49 & 24.45 & 28.1\% & 91.6\% & 0.0275 \\

\multicolumn{9}{p{0.98\textwidth}}{\quad \scriptsize \textit{Heads:} \texttt{L14H1[:], L14H10[NN], L13H8[:], L15H1[NNP], L11H12[NNS], L16H8[.], L10H6[CD], L14H7[NN], L12H6[IN], L14H8[DT], L22H7[VBZ], L14H5[VBZ]}} \\

\specialrule{0.05em}{0.1em}{0.1em}

\texttt{chal-33} & Common-Divis & 2 & $\checkmark\times$ & 22.35 & 22.70 & 31.7\% & 29.0\% & 0.3436 \\

\multicolumn{9}{p{0.98\textwidth}}{\quad \scriptsize \textit{Heads:} \texttt{L21H9[.], L10H6[:], L4H12[:], L16H11[IN], L5H9[EX], L13H3[CD], L14H7[.], L14H5[.], L14H10[DT], L14H1[NNS], L0H1[NNS], L8H3[DT], L22H7[VBZ], L15H9[VBZ]}} \\

\specialrule{0.05em}{0.1em}{0.1em}

\texttt{chal-596} & Multiplicati & 2 & $\checkmark\times$ & 20.96 & 21.44 & 6.4\% & 16.3\% & 0.4535 \\

\multicolumn{9}{p{0.98\textwidth}}{\quad \scriptsize \textit{Heads:} \texttt{L15H8[:], L15H12[:], L13H3[NN], L15H7[NNS], L12H0[VBZ], L23H1[NN], L16H11[CD], L22H7[.], L16H6[.], L16H7[CD], L22H6[CD], L14H12[IN]}} \\

\specialrule{0.05em}{0.1em}{0.1em}

\texttt{chal-674} & Multiplicati & 2 & $\checkmark\times$ & 21.92 & 19.69 & 42.9\% & 5.1\% & 0.5896 \\

\multicolumn{9}{p{0.98\textwidth}}{\quad \scriptsize \textit{Heads:} \texttt{L16H10[VBZ], L16H7[:], L16H6[:], L14H10[NN], L16H11[.], L16H8[.], L14H0[CD], L10H6[RB], L17H7[,], L20H8[,], L12H4[MD], L14H8[DT], L15H7[DT]}} \\

\specialrule{0.05em}{0.1em}{0.1em}

\texttt{chal-859} & Multiplicati & 2 & $\times\times$ & 20.36 & 27.43 & 19.3\% & 49.0\% & 0.2201 \\

\multicolumn{9}{p{0.98\textwidth}}{\quad \scriptsize \textit{Heads:} \texttt{L10H6[:], L14H10[CD], L16H6[NNS], L13H3[NN], L14H3[NN], L15H5[RB], L12H6[VBZ], L15H12[VBZ], L19H11[CD], L14H1[DT], L13H8[.], L18H3[CD], L14H2[IN], L9H1[IN], L15H0[PRP\$]}} \\

\specialrule{0.05em}{0.1em}{0.1em}

\texttt{chal-847} & Multiplicati & 2 & $\checkmark\times$ & 20.58 & 21.94 & 16.4\% & 19.6\% & 0.4247 \\

\multicolumn{9}{p{0.98\textwidth}}{\quad \scriptsize \textit{Heads:} \texttt{L13H8[:], L16H10[NN], L22H7[IN], L14H10[PRP], L12H0[IN], L5H6[NN], L16H11[.], L14H1[VBZ]}} \\

\specialrule{0.05em}{0.1em}{0.1em}

\texttt{chal-193} & Subtraction & 2 & $\checkmark\times$ & 21.19 & 19.81 & 23.9\% & 2.5\% & 0.6303 \\

\multicolumn{9}{p{0.98\textwidth}}{\quad \scriptsize \textit{Heads:} \texttt{L13H12[.], L14H9[:], L10H3[VBG], L13H13[IN], L13H1[CD], L13H4[VBZ], L14H13[.], L14H7[CD], L12H10[VBZ], L4H12[DT], L14H2[VBG], L14H11[PRP], L12H0[VB], L5H9[VB], L15H1[TO], L23H12[RBR]}} \\

\specialrule{0.05em}{0.1em}{0.1em}

\texttt{chal-889} & Subtraction & 2 & $\checkmark\times$ & 20.51 & 22.94 & 8.1\% & 74.5\% & 0.0926 \\

\multicolumn{9}{p{0.98\textwidth}}{\quad \scriptsize \textit{Heads:} \texttt{L14H1[IN], L16H7[IN], L15H8[NNS], L14H13[.], L22H7[.], L16H11[.], L14H10[IN], L14H5[NNS], L14H7[NN], L4H12[PRP], L16H0[VB], L14H8[RBR], L16H2[DT]}} \\

\specialrule{0.05em}{0.1em}{0.1em}

\texttt{chal-622} & Subtraction & 2 & $\checkmark\times$ & 21.09 & 22.12 & 20.6\% & 36.7\% & 0.2928 \\

\multicolumn{9}{p{0.98\textwidth}}{\quad \scriptsize \textit{Heads:} \texttt{L12H6[:], L14H10[:], L14H9[DT], L15H3[VB], L14H7[VBZ], L16H11[.], L16H7[CD], L16H8[CD], L12H12[NNS], L21H9[VBN], L16H13[TO], L13H8[DT], L15H2[RB], L22H7[VBZ]}} \\

\specialrule{0.05em}{0.1em}{0.1em}

\texttt{chal-462} & Subtraction & 2 & $\checkmark\times$ & 20.45 & 22.31 & 6.5\% & 10.1\% & 0.5123 \\

\multicolumn{9}{p{0.98\textwidth}}{\quad \scriptsize \textit{Heads:} \texttt{L16H8[TO], L14H10[.], L10H6[:], L16H11[PRP], L12H0[CD], L15H13[NNS], L14H8[CD], L14H9[NN], L22H7[VBZ]}} \\

\specialrule{0.05em}{0.1em}{0.1em}

\bottomrule

\end{tabularx}

\caption{\small Detailed results of Sequential Multi-Head Patching using JSD across 32 SVAMP instances.
The Stat column indicates correct (\checkmark) or incorrect ($\times$) answers for CoT (left) and No-CoT (right) prompting. The Eq column indicates how many operations the question includes.
Logit and probability shifts represent values before and after patching attention head outputs into the No-CoT execution.
Lower JSD values denote a closer distributional alignment with the clean CoT run.
The Heads sub-row lists the aggregated sets of the most influential attention heads, selected based on their specific POS tags for each respective problem.
The notation format for these elements indicates the layer number, head number, and the corresponding POS category (e.g., L15H3[VB] refers to Layer 15, Head 3, and POS category VB). The average final JSD across all multi-head patching instances is 0.2415.}

\label{tab:normal_patching_jsd}
\end{table*}

\begin{table*}[t]
\centering
\tiny
\renewcommand{\arraystretch}{0.82}
\setlength{\tabcolsep}{1.2pt}

\begin{tabularx}{\textwidth}{l l c c r @{ $\rightarrow$ } l r @{ $\rightarrow$ } l c}
\toprule
\textbf{T-ID} & \textbf{S-ID} & \textbf{T-Stat} & \textbf{S-Stat} & \multicolumn{2}{c}{\textbf{Logit (No $\rightarrow$ Pat.)}} & \multicolumn{2}{c}{\textbf{Prob (No $\rightarrow$ Pat.)}} & \textbf{JSD} \\
\midrule

\texttt{chal-961} & \texttt{chal-674} & $\times\times$ & $\checkmark$ & 23.61 & 22.07 & 36.2\% & 27.1\% & 0.3626 \\
\multicolumn{9}{p{0.98\textwidth}}{\quad \scriptsize \textit{Cross-Heads:} \texttt{L12H0[CD], L14H5[.], L15H1[:], L14H10[MD], L14H8[DT], L12H12[VBP], L14H13[.], L14H1[NNS], L13H1[,], L8H3[VBZ]}} \\
\specialrule{0.05em}{0.05em}{0.05em}
\texttt{chal-569} & \texttt{chal-514} & $\checkmark\times$ & $\times$ & 21.87 & 23.26 & 27.7\% & 55.2\% & 0.1864 \\
\multicolumn{9}{p{0.98\textwidth}}{\quad \scriptsize \textit{Cross-Heads:} \texttt{L14H5[.], L10H7[:], L13H12[CD], L11H3[NN], L14H8[NN], L14H10[VBZ], L13H4[DT], L13H6[CD], L15H1[NNS], L8H3[IN], L13H1[DT]}} \\
\specialrule{0.05em}{0.05em}{0.05em}
\texttt{chal-290} & \texttt{chal-33} & $\times\times$ & $\checkmark$ & 22.23 & 22.73 & 28.2\% & 52.3\% & 0.1993 \\
\multicolumn{9}{p{0.98\textwidth}}{\quad \scriptsize \textit{Cross-Heads:} \texttt{L13H12[.], L15H8[:], L20H10[NNS], L14H10[.], L14H13[.], L14H1[CD], L12H0[CD], L15H10[NNS], L20H12[IN], L13H1[IN], L14H7[DT], L22H7[VBZ]}} \\
\specialrule{0.05em}{0.05em}{0.05em}
\texttt{chal-821} & \texttt{chal-674} & $\checkmark\times$ & $\checkmark$ & 22.63 & 25.89 & 45.4\% & 88.5\% & 0.0413 \\
\multicolumn{9}{p{0.98\textwidth}}{\quad \scriptsize \textit{Cross-Heads:} \texttt{L14H5[CD], L14H8[MD], L13H12[DT], L14H10[NN], L14H2[CD], L4H12[NN], L12H10[,], L5H9[,]}} \\
\specialrule{0.05em}{0.05em}{0.05em}
\texttt{chal-451} & \texttt{chal-21} & $\checkmark\times$ & $\checkmark$ & 22.14 & 24.05 & 30.7\% & 82.2\% & 0.0636 \\
\multicolumn{9}{p{0.98\textwidth}}{\quad \scriptsize \textit{Cross-Heads:} \texttt{L12H13[CD], L11H3[:], L11H5[JJ], L13H3[NNP], L10H6[CD], L10H7[NN], L14H8[NN], L12H0[VBZ], L15H13[.], L13H6[NNS], L12H5[DT]}} \\
\specialrule{0.05em}{0.05em}{0.05em}
\texttt{chal-970} & \texttt{chal-196} & $\checkmark\times$ & $\times$ & 21.38 & 23.68 & 16.4\% & 72.0\% & 0.1047 \\
\multicolumn{9}{p{0.98\textwidth}}{\quad \scriptsize \textit{Cross-Heads:} \texttt{L20H12[.], L15H10[:], L16H0[:], L14H10[NN], L14H13[PRP\$], L21H9[NN], L15H1[NNS], L22H7[.], L16H7[.], L13H3[CD], L15H7[CD], L10H6[IN], L13H12[VBZ], L12H13[,]}} \\
\specialrule{0.05em}{0.05em}{0.05em}
\texttt{chal-215} & \texttt{chal-28} & $\checkmark\times$ & $\times$ & 22.61 & 20.58 & 80.6\% & 99.5\% & 0.0015 \\
\multicolumn{9}{p{0.98\textwidth}}{\quad \scriptsize \textit{Cross-Heads:} \texttt{L13H3[NNS], L13H12[.], L18H9[.], L23H1[NNS], L13H4[NN], L21H1[VBD], L13H1[.], L14H10[CD], L16H12[CD], L15H3[NNS], L16H9[CC], L22H0[CC], L15H6[TO], L13H10[TO], L10H3[PRP\$], L14H8[VBZ]}} \\
\specialrule{0.05em}{0.05em}{0.05em}
\texttt{chal-160} & \texttt{chal-33} & $\times\times$ & $\checkmark$ & 21.85 & 23.40 & 14.7\% & 52.4\% & 0.1828 \\
\multicolumn{9}{p{0.98\textwidth}}{\quad \scriptsize \textit{Cross-Heads:} \texttt{L16H8[.], L10H6[:], L16H13[NNS], L15H3[VBP], L14H10[DT], L12H0[.], L10H10[CD], L13H3[NN], L14H7[IN], L14H12[VBZ]}} \\
\specialrule{0.05em}{0.05em}{0.05em}
\texttt{chal-76} & \texttt{chal-859} & $\times\times$ & $\times$ & 23.77 & 22.68 & 88.3\% & 99.8\% & 0.0004 \\
\multicolumn{9}{p{0.98\textwidth}}{\quad \scriptsize \textit{Cross-Heads:} \texttt{L14H10[.], L15H10[PRP], L22H7[PRP\$], L14H7[CD], L16H6[.], L16H10[CD], L14H9[IN], L14H8[DT], L23H1[JJ], L13H2[JJ], L18H9[CC]}} \\
\specialrule{0.05em}{0.05em}{0.05em}
\texttt{chal-854} & \texttt{chal-847} & $\checkmark\times$ & $\checkmark$ & 18.52 & 21.55 & 1.3\% & 4.9\% & 0.5883 \\
\multicolumn{9}{p{0.98\textwidth}}{\quad \scriptsize \textit{Cross-Heads:} \texttt{L12H0[CD], L14H13[IN], L13H8[:], L14H10[IN], L10H6[PRP], L14H1[NN], L16H7[.], L16H11[.], L22H0[CD], L13H3[RB], L8H3[RB], L15H13[DT], L14H12[DT]}} \\
\specialrule{0.05em}{0.05em}{0.05em}
\texttt{chal-196} & \texttt{chal-889} & $\times\times$ & $\checkmark$ & 20.80 & 21.32 & 39.9\% & 11.2\% & 0.5066 \\
\multicolumn{9}{p{0.98\textwidth}}{\quad \scriptsize \textit{Cross-Heads:} \texttt{L14H10[IN], L14H1[:], L14H5[CD], L14H7[NNS], L12H10[.], L8H7[CD], L12H0[NNS], L9H2[NN], L13H3[VB], L13H12[DT], L4H12[PRP], L11H0[VB], L12H2[VBZ], L14H8[IN], L13H6[DT]}} \\
\specialrule{0.05em}{0.05em}{0.05em}
\texttt{chal-12} & \texttt{chal-569} & $\checkmark\checkmark$ & $\checkmark$ & 23.75 & 23.97 & 75.2\% & 97.2\% & 0.0092 \\
\multicolumn{9}{p{0.98\textwidth}}{\quad \scriptsize \textit{Cross-Heads:} \texttt{L16H8[.], L16H10[NN], L13H12[JJ], L13H1[NN], L16H7[CD], L14H10[CD], L14H8[DT], L10H6[VBZ]}} \\
\specialrule{0.05em}{0.05em}{0.05em}
\texttt{chal-163} & \texttt{chal-76} & $\times\times$ & $\times$ & 20.37 & 21.81 & 8.0\% & 31.5\% & 0.3291 \\
\multicolumn{9}{p{0.98\textwidth}}{\quad \scriptsize \textit{Cross-Heads:} \texttt{L16H10[CD], L16H6[.], L15H10[NN], L10H6[IN], L14H9[DT], L16H11[.], L14H3[NNS], L14H10[VB]}} \\
\specialrule{0.05em}{0.05em}{0.05em}
\texttt{chal-514} & \texttt{chal-859} & $\times\times$ & $\times$ & 22.31 & 21.81 & 17.7\% & 23.2\% & 0.3889 \\
\multicolumn{9}{p{0.98\textwidth}}{\quad \scriptsize \textit{Cross-Heads:} \texttt{L14H1[VBZ], L16H0[PRP], L14H13[,], L10H6[CD], L15H1[VBZ], L16H7[.], L16H11[.], L14H10[CD], L12H0[IN], L14H12[DT], L8H3[RB]}} \\
\specialrule{0.05em}{0.05em}{0.05em}
\texttt{chal-954} & \texttt{chal-196} & $\checkmark\times$ & $\times$ & 22.90 & 25.28 & 44.6\% & 97.7\% & 0.0071 \\
\multicolumn{9}{p{0.98\textwidth}}{\quad \scriptsize \textit{Cross-Heads:} \texttt{L14H10[IN], L14H0[:], L12H10[:], L21H9[NN], L9H1[DT], L14H7[.], L11H12[NNS], L13H3[NNS], L22H0[CD], L15H1[CD], L10H7[RB], L14H1[,], L15H10[PRP], L13H7[PRP], L11H0[VBZ], L12H13[VBZ]}} \\
\specialrule{0.05em}{0.05em}{0.05em}
\texttt{chal-83} & \texttt{chal-21} & $\times\times$ & $\checkmark$ & 20.75 & 22.53 & 6.6\% & 18.0\% & 0.4264 \\
\multicolumn{9}{p{0.98\textwidth}}{\quad \scriptsize \textit{Cross-Heads:} \texttt{L16H8[.], L10H6[NN], L16H7[.], L14H8[NNP], L16H11[.], L14H10[CD], L12H0[CD], L12H13[JJ], L20H9[DT]}} \\
\specialrule{0.05em}{0.05em}{0.05em}
\texttt{chal-21} & \texttt{chal-889} & $\checkmark\times$ & $\checkmark$ & 21.14 & 22.77 & 25.9\% & 79.4\% & 0.0768 \\
\multicolumn{9}{p{0.98\textwidth}}{\quad \scriptsize \textit{Cross-Heads:} \texttt{L14H1[.], L15H8[:], L13H3[:], L14H13[.], L15H3[NNS], L16H7[IN], L13H12[DT], L14H10[VBZ]}} \\
\specialrule{0.05em}{0.05em}{0.05em}
\texttt{chal-129} & \texttt{chal-12} & $\checkmark\times$ & $\checkmark$ & 21.62 & 22.25 & 17.5\% & 24.4\% & 0.3847 \\
\multicolumn{9}{p{0.98\textwidth}}{\quad \scriptsize \textit{Cross-Heads:} \texttt{L14H9[NN], L14H7[.], L16H8[:], L12H0[NNS], L13H3[NN], L15H3[CD]}} \\
\specialrule{0.05em}{0.05em}{0.05em}
\texttt{chal-587} & \texttt{chal-129} & $\times\times$ & $\checkmark$ & 21.15 & 25.46 & 21.9\% & 99.2\% & 0.0019 \\
\multicolumn{9}{p{0.98\textwidth}}{\quad \scriptsize \textit{Cross-Heads:} \texttt{L13H3[CD], L15H7[IN], L14H10[:], L13H6[VBN], L14H0[NN], L22H0[CD], L12H2[NNS], L13H1[DT], L8H7[DT], L13H10[RB]}} \\
\specialrule{0.05em}{0.05em}{0.05em}
\texttt{chal-254} & \texttt{chal-76} & $\checkmark\times$ & $\times$ & 23.73 & 22.47 & 47.7\% & 22.7\% & 0.3970 \\
\multicolumn{9}{p{0.98\textwidth}}{\quad \scriptsize \textit{Cross-Heads:} \texttt{L13H3[,], L10H6[:], L15H1[CD], L13H7[VBZ], L9H1[VB], L8H3[NN], L4H12[MD], L20H8[CD], L5H6[.], L16H12[.], L22H6[CD], L10H4[,], L14H10[PRP], L13H0[VB]}} \\
\specialrule{0.05em}{0.05em}{0.05em}
\texttt{chal-28} & \texttt{chal-674} & $\times\times$ & $\checkmark$ & 18.66 & 19.92 & 1.3\% & 5.0\% & 0.5627 \\
\multicolumn{9}{p{0.98\textwidth}}{\quad \scriptsize \textit{Cross-Heads:} \texttt{L14H7[CD], L14H9[CD], L14H13[IN], L16H11[.], L14H10[NN]}} \\
\specialrule{0.05em}{0.05em}{0.05em}
\texttt{chal-906} & \texttt{chal-33} & $\checkmark\times$ & $\checkmark$ & 20.69 & 21.07 & 7.3\% & 3.8\% & 0.6083 \\
\multicolumn{9}{p{0.98\textwidth}}{\quad \scriptsize \textit{Cross-Heads:} \texttt{L15H6[DT], L8H7[:], L15H9[EX], L10H6[CD], L14H7[.], L14H13[.], L14H8[CD], L12H0[NN], L14H0[NN], L14H9[NNS], L13H3[IN]}} \\
\specialrule{0.05em}{0.05em}{0.05em}
\texttt{chal-829} & \texttt{chal-462} & $\checkmark\times$ & $\checkmark$ & 22.49 & 24.08 & 28.1\% & 75.4\% & 0.0906 \\
\multicolumn{9}{p{0.98\textwidth}}{\quad \scriptsize \textit{Cross-Heads:} \texttt{L14H1[:], L14H5[.], L14H10[NN], L15H1[PRP], L12H0[NN], L13H8[DT], L12H8[DT], L14H6[VBZ]}} \\
\specialrule{0.05em}{0.05em}{0.05em}
\texttt{chal-33} & \texttt{chal-906} & $\checkmark\times$ & $\checkmark$ & 22.35 & 22.58 & 31.7\% & 38.8\% & 0.2762 \\
\multicolumn{9}{p{0.98\textwidth}}{\quad \scriptsize \textit{Cross-Heads:} \texttt{L21H9[NN], L10H6[:], L4H12[:], L14H10[VBZ], L14H1[NNS], L2H9[IN], L14H11[POS], L9H0[POS], L9H13[POS], L15H7[,], L20H8[CD], L15H9[CD], L13H3[CD], L18H7[VBN], L16H11[VBN]}} \\
\specialrule{0.05em}{0.05em}{0.05em}
\texttt{chal-596} & \texttt{chal-163} & $\checkmark\times$ & $\times$ & 20.96 & 20.85 & 6.4\% & 18.5\% & 0.4330 \\
\multicolumn{9}{p{0.98\textwidth}}{\quad \scriptsize \textit{Cross-Heads:} \texttt{L16H11[.], L16H7[.], L14H13[:], L12H0[VBZ], L13H3[.], L23H1[NN], L22H0[CD], L16H6[NNS], L15H5[NNS], L10H4[JJ], L8H7[RBR]}} \\
\specialrule{0.05em}{0.05em}{0.05em}
\texttt{chal-674} & \texttt{chal-906} & $\checkmark\times$ & $\checkmark$ & 21.92 & 22.75 & 42.9\% & 56.7\% & 0.1784 \\
\multicolumn{9}{p{0.98\textwidth}}{\quad \scriptsize \textit{Cross-Heads:} \texttt{L16H10[NN], L16H7[NN], L0H11[:], L10H6[DT], L14H0[NNS], L16H6[NN], L15H7[POS], L19H11[.], L13H1[CD], L14H10[CD], L13H10[NNS], L15H5[VBN]}} \\
\specialrule{0.05em}{0.05em}{0.05em}
\texttt{chal-859} & \texttt{chal-160} & $\times\times$ & $\times$ & 20.36 & 23.69 & 19.3\% & 37.3\% & 0.2901 \\
\multicolumn{9}{p{0.98\textwidth}}{\quad \scriptsize \textit{Cross-Heads:} \texttt{L10H6[:], L16H6[.], L13H8[:], L15H12[EX], L14H1[NN], L13H3[NNS], L9H1[,], L15H0[DT], L16H11[.], L22H7[.], L14H10[CD], L18H3[CD], L19H11[CD], L14H3[CC], L15H5[NN], L15H7[RB], L11H2[,], L12H6[VBN], L15H6[VBZ]}} \\
\specialrule{0.05em}{0.05em}{0.05em}
\texttt{chal-847} & \texttt{chal-254} & $\checkmark\times$ & $\checkmark$ & 20.58 & 22.71 & 16.4\% & 20.4\% & 0.4176 \\
\multicolumn{9}{p{0.98\textwidth}}{\quad \scriptsize \textit{Cross-Heads:} \texttt{L13H8[:], L16H10[IN], L12H0[IN], L15H13[.], L13H3[CD], L4H12[VBD], L15H8[.], L14H10[CD], L14H1[NNS], L12H13[IN], L15H1[DT], L13H10[VBZ]}} \\
\specialrule{0.05em}{0.05em}{0.05em}
\texttt{chal-193} & \texttt{chal-215} & $\checkmark\times$ & $\checkmark$ & 21.19 & 20.81 & 23.9\% & 22.0\% & 0.4025 \\
\multicolumn{9}{p{0.98\textwidth}}{\quad \scriptsize \textit{Cross-Heads:} \texttt{L13H12[:], L13H4[NNS], L10H3[:], L15H6[IN], L4H12[JJ], L13H1[NNS], L12H10[.], L20H7[CD], L12H2[,]}} \\
\specialrule{0.05em}{0.05em}{0.05em}
\texttt{chal-889} & \texttt{chal-854} & $\checkmark\times$ & $\checkmark$ & 20.51 & 22.39 & 8.1\% & 65.9\% & 0.1307 \\
\multicolumn{9}{p{0.98\textwidth}}{\quad \scriptsize \textit{Cross-Heads:} \texttt{L14H13[.], L15H1[:], L17H7[:], L14H10[CD], L14H1[NNS], L16H0[VBZ], L16H8[.], L7H13[NNS], L0H11[NNS], L13H3[DT]}} \\
\specialrule{0.05em}{0.05em}{0.05em}
\texttt{chal-622} & \texttt{chal-462} & $\checkmark\times$ & $\checkmark$ & 21.09 & 24.13 & 20.6\% & 54.5\% & 0.1895 \\
\multicolumn{9}{p{0.98\textwidth}}{\quad \scriptsize \textit{Cross-Heads:} \texttt{L14H9[VBZ], L20H12[.], L14H7[VBZ], L15H3[IN], L14H3[IN], L13H3[VB], L16H11[CD], L16H7[.], L16H8[CD], L13H12[CD], L12H12[RBR], L10H6[RBR], L13H8[DT], L16H13[VBZ]}} \\
\specialrule{0.05em}{0.05em}{0.05em}
\texttt{chal-462} & \texttt{chal-129} & $\checkmark\times$ & $\checkmark$ & 20.45 & 22.07 & 6.5\% & 14.6\% & 0.4628 \\
\multicolumn{9}{p{0.98\textwidth}}{\quad \scriptsize \textit{Cross-Heads:} \texttt{L16H8[.], L16H7[CD], L16H11[CD], L13H2[VBN], L14H10[.], L14H9[NN], L14H8[RB], L13H13[IN], L10H6[DT], L14H7[VBZ]}} \\
\specialrule{0.05em}{0.05em}{0.05em}

\bottomrule
\end{tabularx}
\caption{\small Detailed results of Sequential Multi-Head Cross Patching using JSD across 32 SVAMP instances.
T-ID and S-ID represent the Target and Source problem IDs, respectively.
The T-Stat column indicates correct (\checkmark) or incorrect ($\times$) answers for the target problem under CoT (left) and No-CoT (right) prompting, while the S-Stat column shows the correctness of the source CoT generation.
Logit and probability shifts represent values before and after patching the source attention-head outputs into the target No-CoT execution.
Lower JSD values denote a closer distributional alignment with the clean target CoT run.
The Cross-Heads sub-row lists the aggregated sets of the most influential attention heads extracted from the source problem, selected based on their specific POS tags.
The notation indicates the layer number, head number, and corresponding POS category (e.g., L15H3[VB] denotes Layer 15, Head 3, with POS category VB). The average final JSD across all multi-head cross-patching instances is 0.2594.}
\label{tab:cross_patching_jsd}
\end{table*}

\begin{table*}[t]

\centering

\tiny

\renewcommand{\arraystretch}{0.82}

\setlength{\tabcolsep}{1.2pt}

\begin{tabularx}{\textwidth}{l l c c r @{ $\rightarrow$ } l r @{ $\rightarrow$ } l c}

\toprule

\textbf{ID} & \textbf{Type} & \textbf{Eq} & \textbf{Stat} & \multicolumn{2}{c}{\textbf{Logit (No $\rightarrow$ Pat.)}} & \multicolumn{2}{c}{\textbf{Prob (No $\rightarrow$ Pat.)}} & \textbf{MRR} \\

\midrule

\texttt{chal-961} & Addition & 1 & $\times\times$ & 23.61 & 19.38 & 36.2\% & 0.3\% & -0.6122 \\

\multicolumn{9}{p{0.98\textwidth}}{\quad \scriptsize \textit{Heads:} \texttt{L16H13[NNP], L16H10[CD], L13H12[:], L13H1[NNP], L13H0[NNP], L14H5[VBZ], L14H10[CD], L12H13[VBZ], L16H8[.], L14H13[.], L13H4[NN], L14H1[NN], L12H0[JJ], L15H1[NNS], L13H7[PRP\$]}} \\

\specialrule{0.05em}{0.05em}{0.05em}

\texttt{chal-569} & Addition & 1 & $\checkmark\times$ & 21.87 & 21.07 & 27.7\% & 6.7\% & -0.3062 \\

\multicolumn{9}{p{0.98\textwidth}}{\quad \scriptsize \textit{Heads:} \texttt{L10H7[:], L13H12[NN], L16H10[NNS], L14H5[CC], L14H8[NN], L8H3[VBD], L10H12[JJ], L8H5[JJ], L9H2[.], L10H3[CD], L9H1[CD], L10H9[IN], L13H13[CC], L13H4[DT], L13H10[DT], L14H10[VBZ]}} \\

\specialrule{0.05em}{0.05em}{0.05em}

\texttt{chal-290} & Addition & 1 & $\times\times$ & 22.23 & 24.18 & 28.2\% & 52.3\% & -0.0281 \\

\multicolumn{9}{p{0.98\textwidth}}{\quad \scriptsize \textit{Heads:} \texttt{L14H10[NN], L13H8[:], L22H7[:], L13H12[VBD], L8H3[EX], L12H12[TO], L20H12[.], L14H7[CD], L15H10[NNS], L15H3[NNP], L13H4[RB], L12H6[,], L14H12[VBZ], L14H13[VBZ], L20H10[VBZ]}} \\

\specialrule{0.05em}{0.05em}{0.05em}

\texttt{chal-821} & Addition & 1 & $\checkmark\times$ & 22.63 & 24.00 & 45.4\% & 98.3\% & 0.4600 \\

\multicolumn{9}{p{0.98\textwidth}}{\quad \scriptsize \textit{Heads:} \texttt{L13H3[NNP], L14H5[CD], L19H8[:], L4H12[NNP], L14H10[VBZ], L14H2[CD], L9H6[VBD], L12H10[.], L15H9[CD], L14H1[NN], L14H11[NNS], L11H12[IN], L22H7[VBZ]}} \\

\specialrule{0.05em}{0.05em}{0.05em}

\texttt{chal-451} & Common-Divis & 1 & $\checkmark\times$ & 22.14 & 24.33 & 30.7\% & 84.9\% & 0.3180 \\

\multicolumn{9}{p{0.98\textwidth}}{\quad \scriptsize \textit{Heads:} \texttt{L11H3[VBZ], L0H11[:], L4H12[:], L10H6[CD], L12H13[NNS], L1H4[PRP], L13H3[IN], L13H10[VBN], L14H7[CD], L15H13[CD], L10H12[.], L13H6[NNS], L11H5[NN], L10H7[CC], L12H0[NN], L14H8[DT], L22H7[DT]}} \\

\specialrule{0.05em}{0.05em}{0.05em}

\texttt{chal-970} & Common-Divis & 1 & $\checkmark\times$ & 21.38 & 22.44 & 16.4\% & 23.8\% & 0.0895 \\

\multicolumn{9}{p{0.98\textwidth}}{\quad \scriptsize \textit{Heads:} \texttt{L13H12[PRP\$], L14H12[PRP\$], L13H8[NN], L12H13[CC], L14H10[DT], L10H6[VBZ], L22H7[VBZ], L20H12[.], L16H13[.], L16H11[.], L8H7[CD], L13H3[IN]}} \\

\specialrule{0.05em}{0.05em}{0.05em}

\texttt{chal-215} & Common-Divis & 1 & $\checkmark\times$ & 22.61 & 18.88 & 80.6\% & 98.2\% & 0.5260 \\

\multicolumn{9}{p{0.98\textwidth}}{\quad \scriptsize \textit{Heads:} \texttt{L13H6[:], L23H1[NN], L13H4[:], L13H12[DT], L15H6[IN], L12H4[DT], L16H12[NN], L15H9[VBZ], L16H13[.], L13H1[CD], L15H3[.], L13H10[CD], L22H0[CD], L14H11[JJ], L11H12[JJ], L18H9[VB], L8H3[RB]}} \\

\specialrule{0.05em}{0.05em}{0.05em}

\texttt{chal-160} & Common-Divis & 1 & $\times\times$ & 21.85 & 23.95 & 14.7\% & 48.4\% & 0.2240 \\

\multicolumn{9}{p{0.98\textwidth}}{\quad \scriptsize \textit{Heads:} \texttt{L16H8[CC], L10H6[RB], L16H13[CD], L8H7[EX], L15H13[.], L15H12[RB], L14H10[CD], L13H3[NN], L15H10[NNS], L12H0[CC], L14H12[NN], L14H8[JJ], L8H3[VBN]}} \\

\specialrule{0.05em}{0.05em}{0.05em}

\texttt{chal-76} & Multiplicati & 1 & $\times\times$ & 23.77 & 20.00 & 88.3\% & 99.3\% & 0.9293 \\

\multicolumn{9}{p{0.98\textwidth}}{\quad \scriptsize \textit{Heads:} \texttt{L14H10[:], L17H7[PRP], L18H9[NNS], L23H1[NN], L18H3[DT], L14H7[CD], L15H10[VBZ], L14H13[.], L14H9[.], L22H0[CD], L22H6[CD], L17H8[PRP]}} \\

\specialrule{0.05em}{0.05em}{0.05em}

\texttt{chal-854} & Multiplicati & 1 & $\checkmark\times$ & 18.52 & 20.75 & 1.3\% & 4.3\% & 0.1885 \\

\multicolumn{9}{p{0.98\textwidth}}{\quad \scriptsize \textit{Heads:} \texttt{L12H0[CD], L13H8[:], L16H7[:], L14H10[NN], L15H13[NN], L14H13[DT], L13H3[VBZ], L10H6[CD], L14H12[DT]}} \\

\specialrule{0.05em}{0.05em}{0.05em}

\texttt{chal-196} & Multiplicati & 1 & $\times\times$ & 20.80 & 24.14 & 39.9\% & 95.3\% & 0.4262 \\

\multicolumn{9}{p{0.98\textwidth}}{\quad \scriptsize \textit{Heads:} \texttt{L14H10[:], L11H0[:], L14H8[:], L12H0[NN], L9H2[NN], L14H7[.], L13H1[NNS], L13H4[NNS], L16H7[.], L13H3[.], L8H7[CD], L12H2[,], L9H4[,], L14H1[RB], L14H5[JJ], L19H6[VB], L11H8[VB], L13H6[VBZ], L6H9[VBZ], L14H3[VBZ]}} \\

\specialrule{0.05em}{0.05em}{0.05em}

\texttt{chal-12} & Multiplicati & 1 & $\checkmark\checkmark$ & 23.75 & 25.35 & 75.2\% & 99.4\% & 0.5120 \\

\multicolumn{9}{p{0.98\textwidth}}{\quad \scriptsize \textit{Heads:} \texttt{L16H8[CD], L16H10[NN], L13H12[DT], L13H1[NN], L20H8[CD], L14H2[NNS], L16H13[IN]}} \\

\specialrule{0.05em}{0.05em}{0.05em}

\texttt{chal-163} & Subtraction & 1 & $\times\times$ & 20.37 & 22.23 & 8.0\% & 8.9\% & 0.0518 \\

\multicolumn{9}{p{0.98\textwidth}}{\quad \scriptsize \textit{Heads:} \texttt{L16H10[:], L16H6[DT], L14H10[CD], L10H6[PRP], L14H13[CD], L14H3[.], L14H2[NNS], L16H8[.], L12H0[CD], L14H5[JJ], L16H7[CC], L14H9[DT], L12H12[RBR]}} \\

\specialrule{0.05em}{0.05em}{0.05em}

\texttt{chal-514} & Subtraction & 1 & $\times\times$ & 22.31 & 23.52 & 17.7\% & 77.2\% & 0.3264 \\

\multicolumn{9}{p{0.98\textwidth}}{\quad \scriptsize \textit{Heads:} \texttt{L12H0[:], L14H1[RBR], L14H13[RBR], L16H0[.], L10H6[CD], L14H5[RBR], L16H11[.], L16H7[CD]}} \\

\specialrule{0.05em}{0.05em}{0.05em}

\texttt{chal-954} & Subtraction & 1 & $\checkmark\times$ & 22.90 & 23.79 & 44.6\% & 69.2\% & 0.1870 \\

\multicolumn{9}{p{0.98\textwidth}}{\quad \scriptsize \textit{Heads:} \texttt{L14H10[:], L16H6[DT], L21H9[VBZ], L10H6[.], L11H12[CD], L14H1[CD], L15H13[NN], L11H0[VBZ]}} \\

\specialrule{0.05em}{0.05em}{0.05em}

\texttt{chal-83} & Subtraction & 1 & $\times\times$ & 20.75 & 21.02 & 6.6\% & 11.8\% & 0.1335 \\

\multicolumn{9}{p{0.98\textwidth}}{\quad \scriptsize \textit{Heads:} \texttt{L16H8[.], L16H7[.], L12H0[CD], L10H6[PRP], L16H11[.], L14H13[CD], L15H13[DT], L14H9[VBZ]}} \\

\specialrule{0.05em}{0.05em}{0.05em}

\texttt{chal-21} & Addition & 2 & $\checkmark\times$ & 21.14 & 22.11 & 25.9\% & 76.1\% & 0.2541 \\

\multicolumn{9}{p{0.98\textwidth}}{\quad \scriptsize \textit{Heads:} \texttt{L14H13[CD], L14H1[.], L16H11[:], L15H3[NNP], L14H10[VBZ], L22H7[.], L16H8[CD], L15H5[JJ], L13H12[DT]}} \\

\specialrule{0.05em}{0.05em}{0.05em}

\texttt{chal-129} & Addition & 2 & $\checkmark\times$ & 21.62 & 21.89 & 17.5\% & 28.2\% & 0.1605 \\

\multicolumn{9}{p{0.98\textwidth}}{\quad \scriptsize \textit{Heads:} \texttt{L16H8[VBZ], L16H6[:], L16H7[.], L12H0[NN], L15H13[NN], L14H7[.], L16H11[CD], L14H9[DT], L14H10[RB]}} \\

\specialrule{0.05em}{0.05em}{0.05em}

\texttt{chal-587} & Addition & 2 & $\times\times$ & 21.15 & 26.01 & 21.9\% & 88.4\% & 0.2771 \\

\multicolumn{9}{p{0.98\textwidth}}{\quad \scriptsize \textit{Heads:} \texttt{L12H2[.], L12H8[:], L13H3[CD], L4H12[EX], L8H5[DT], L10H7[VBD], L16H8[.], L14H7[NNS], L12H6[CD], L7H13[NN], L12H13[NN], L10H11[VBZ]}} \\

\specialrule{0.05em}{0.05em}{0.05em}

\texttt{chal-254} & Addition & 2 & $\checkmark\times$ & 23.73 & 22.09 & 47.7\% & 42.4\% & 0.0211 \\

\multicolumn{9}{p{0.98\textwidth}}{\quad \scriptsize \textit{Heads:} \texttt{L13H3[JJ], L14H10[:], L14H13[JJ], L14H1[CD], L4H12[VBD], L16H8[CD], L13H4[DT], L16H10[.], L12H0[.], L9H1[IN], L14H8[PRP]}} \\

\specialrule{0.05em}{0.05em}{0.05em}

\texttt{chal-28} & Common-Divis & 2 & $\times\times$ & 18.66 & 20.32 & 1.3\% & 5.1\% & 0.1803 \\

\multicolumn{9}{p{0.98\textwidth}}{\quad \scriptsize \textit{Heads:} \texttt{L16H13[CC], L14H9[VBZ], L16H11[CD], L12H0[NN], L14H10[.], L15H1[NN], L13H6[VBD], L13H7[VBD], L15H9[CD], L16H7[CD], L15H10[CC], L13H1[NNS], L14H7[PRP\$], L14H13[DT], L8H3[VBZ]}} \\

\specialrule{0.05em}{0.05em}{0.05em}

\texttt{chal-906} & Common-Divis & 2 & $\checkmark\times$ & 20.69 & 20.87 & 7.3\% & 11.4\% & 0.0537 \\

\multicolumn{9}{p{0.98\textwidth}}{\quad \scriptsize \textit{Heads:} \texttt{L15H6[IN], L16H8[:], L8H7[POS], L10H6[NN], L12H10[DT], L14H7[.], L14H8[NN], L13H7[IN], L13H3[CD], L16H7[.], L14H5[RB], L12H13[,], L14H6[VBN]}} \\

\specialrule{0.05em}{0.05em}{0.05em}

\texttt{chal-829} & Common-Divis & 2 & $\checkmark\times$ & 22.49 & 23.47 & 28.1\% & 84.9\% & 0.3686 \\

\multicolumn{9}{p{0.98\textwidth}}{\quad \scriptsize \textit{Heads:} \texttt{L14H1[:], L14H10[NN], L10H2[:], L13H12[DT], L15H1[NNP], L11H12[NNS], L14H7[.], L10H6[CD], L12H6[IN], L12H2[IN], L11H0[PRP], L22H7[VBZ]}} \\

\specialrule{0.05em}{0.05em}{0.05em}

\texttt{chal-33} & Common-Divis & 2 & $\checkmark\times$ & 22.35 & 20.36 & 31.7\% & 16.7\% & -0.1202 \\

\multicolumn{9}{p{0.98\textwidth}}{\quad \scriptsize \textit{Heads:} \texttt{L16H11[CD], L21H9[NN], L10H6[:], L14H10[DT], L13H11[VBP], L14H5[.], L14H7[.], L14H1[CD], L23H1[NNS], L0H1[NNS], L14H13[IN], L10H12[DT], L22H7[VBZ]}} \\

\specialrule{0.05em}{0.05em}{0.05em}

\texttt{chal-596} & Multiplicati & 2 & $\checkmark\times$ & 20.96 & 21.09 & 6.4\% & 9.8\% & 0.0706 \\

\multicolumn{9}{p{0.98\textwidth}}{\quad \scriptsize \textit{Heads:} \texttt{L13H3[PRP], L14H13[PRP], L15H1[:], L12H13[NN], L23H1[VBZ], L16H11[CD], L16H6[.], L22H7[.], L22H4[CD], L16H7[CD], L12H0[VBZ]}} \\

\specialrule{0.05em}{0.05em}{0.05em}

\texttt{chal-674} & Multiplicati & 2 & $\checkmark\times$ & 21.92 & 20.24 & 42.9\% & 10.2\% & -0.1944 \\

\multicolumn{9}{p{0.98\textwidth}}{\quad \scriptsize \textit{Heads:} \texttt{L16H10[VBZ], L16H7[CD], L13H3[:], L14H13[IN], L16H11[.], L16H8[.], L14H10[NN], L10H6[DT]}} \\

\specialrule{0.05em}{0.05em}{0.05em}

\texttt{chal-859} & Multiplicati & 2 & $\times\times$ & 20.36 & 26.92 & 19.3\% & 97.2\% & 0.5255 \\

\multicolumn{9}{p{0.98\textwidth}}{\quad \scriptsize \textit{Heads:} \texttt{L21H9[NN], L14H1[RB], L10H6[RB], L14H10[CD], L13H3[DT], L19H11[CD], L15H3[NNS], L15H7[VBZ], L16H7[.], L18H3[CD], L22H7[PRP\$], L15H5[PRP\$], L20H8[PRP\$], L4H12[JJ], L9H1[JJ], L20H9[CC]}} \\

\specialrule{0.05em}{0.05em}{0.05em}

\texttt{chal-847} & Multiplicati & 2 & $\checkmark\times$ & 20.58 & 23.02 & 16.4\% & 38.0\% & 0.1337 \\

\multicolumn{9}{p{0.98\textwidth}}{\quad \scriptsize \textit{Heads:} \texttt{L13H8[:], L14H13[IN], L13H3[NNS], L12H0[VB], L14H1[NNS], L14H10[CD], L16H11[CD], L22H7[IN], L16H7[CD], L5H6[RB], L4H12[,], L9H1[VB], L14H5[DT]}} \\

\specialrule{0.05em}{0.05em}{0.05em}

\texttt{chal-193} & Subtraction & 2 & $\checkmark\times$ & 21.19 & 21.78 & 23.9\% & 56.4\% & 0.1533 \\

\multicolumn{9}{p{0.98\textwidth}}{\quad \scriptsize \textit{Heads:} \texttt{L13H12[PRP], L14H10[.], L14H9[:], L5H9[VB], L4H12[VBZ], L13H1[VBP], L14H2[CD], L14H13[.], L22H4[CD], L15H1[NNS], L10H3[NNS], L14H11[NN], L10H12[CC], L16H7[DT], L13H4[VBZ], L15H12[RB], L15H0[PRP], L14H8[TO], L23H12[RBR]}} \\

\specialrule{0.05em}{0.05em}{0.05em}

\texttt{chal-889} & Subtraction & 2 & $\checkmark\times$ & 20.51 & 22.83 & 8.1\% & 72.7\% & 0.4177 \\

\multicolumn{9}{p{0.98\textwidth}}{\quad \scriptsize \textit{Heads:} \texttt{L16H8[CD], L16H7[IN], L14H1[IN], L14H13[.], L16H11[.], L22H7[.], L14H10[IN], L13H3[PRP]}} \\

\specialrule{0.05em}{0.05em}{0.05em}

\texttt{chal-622} & Subtraction & 2 & $\checkmark\times$ & 21.09 & 23.48 & 20.6\% & 56.7\% & 0.0842 \\

\multicolumn{9}{p{0.98\textwidth}}{\quad \scriptsize \textit{Heads:} \texttt{L12H6[:], L15H3[VB], L10H5[:], L10H6[VB], L12H12[PRP], L16H11[.], L16H7[CD], L16H8[CD], L14H9[NN], L11H1[NNS], L15H11[NNS], L21H9[VBN], L13H8[VBN], L15H2[TO], L1H9[,], L22H7[VBZ], L15H7[VBZ]}} \\

\specialrule{0.05em}{0.05em}{0.05em}

\texttt{chal-462} & Subtraction & 2 & $\checkmark\times$ & 20.45 & 21.31 & 6.5\% & 5.3\% & -0.1158 \\

\multicolumn{9}{p{0.98\textwidth}}{\quad \scriptsize \textit{Heads:} \texttt{L16H8[IN], L14H10[:], L10H6[CD], L16H11[PRP], L15H13[NNS], L12H0[NNS], L14H8[CD], L16H7[.], L14H9[NN]}} \\

\specialrule{0.05em}{0.05em}{0.05em}

\bottomrule

\end{tabularx}

\caption{\small Detailed results of Sequential Multi-Head Patching using MRR (Margin Recovery Ratio) across 32 SVAMP instances.
The Stat column indicates correct (\checkmark) or incorrect ($\times$) answers for CoT (left) and No-CoT (right) prompting.
Logit and probability shifts represent values before and after patching attention head outputs into the No-CoT execution.
Higher MRR values indicate greater recovery of the CoT--No-CoT margin difference.
The Heads sub-row lists the aggregated sets of the most influential attention heads, selected based on their specific POS tags for each respective problem.
The notation indicates the layer number, head number, and corresponding POS category (e.g., L15H3[VB] denotes Layer 15, Head 3, with POS category VB). The average MRR across all multi-head patching instances is 0.1780.}

\label{tab:normal_patching_mrr}

\end{table*}

\begin{table*}[t]

\centering

\tiny

\renewcommand{\arraystretch}{0.82}

\setlength{\tabcolsep}{1.2pt}

\begin{tabularx}{\textwidth}{l l c c r @{ $\rightarrow$ } l r @{ $\rightarrow$ } l c}

\toprule

\textbf{T-ID} & \textbf{S-ID} & \textbf{T-Stat} & \textbf{S-Stat} & \multicolumn{2}{c}{\textbf{Logit (No $\rightarrow$ Pat.)}} & \multicolumn{2}{c}{\textbf{Prob (No $\rightarrow$ Pat.)}} & \textbf{MRR} \\

\midrule

\texttt{chal-961} & \texttt{chal-674} & $\times\times$ & $\checkmark$ & 23.61 & 21.54 & 36.2\% & 14.6\% & -0.0750 \\

\multicolumn{9}{p{0.98\textwidth}}{\quad \scriptsize \textit{Cross-Heads:} \texttt{L12H0[CD], L14H5[.], L12H12[VBP], L14H10[MD], L14H8[IN], L14H13[.], L14H1[NNS], L13H1[,], L8H3[VBZ]}} \\

\specialrule{0.05em}{0.05em}{0.05em}

\texttt{chal-569} & \texttt{chal-290} & $\checkmark\times$ & $\times$ & 21.87 & 22.71 & 27.7\% & 4.6\% & -0.3640 \\

\multicolumn{9}{p{0.98\textwidth}}{\quad \scriptsize \textit{Cross-Heads:} \texttt{L10H7[:], L13H12[IN], L14H8[TO], L13H10[NN], L15H1[NNP], L9H1[NN], L13H4[.], L16H8[CD], L14H10[CD], L13H7[NNP], L14H5[NNS], L10H12[JJ], L10H9[TO], L10H13[VBZ]}} \\

\specialrule{0.05em}{0.05em}{0.05em}

\texttt{chal-290} & \texttt{chal-622} & $\times\times$ & $\checkmark$ & 22.23 & 23.30 & 28.2\% & 35.5\% & -0.0634 \\

\multicolumn{9}{p{0.98\textwidth}}{\quad \scriptsize \textit{Cross-Heads:} \texttt{L14H10[:], L13H12[.], L15H10[CD], L15H3[CD], L20H12[VBD], L20H10[.], L13H11[VBD], L15H1[NN], L14H7[NNS], L12H0[VBN], L15H7[,], L12H6[PRP], L14H12[VBZ]}} \\

\specialrule{0.05em}{0.05em}{0.05em}

\texttt{chal-821} & \texttt{chal-587} & $\checkmark\times$ & $\times$ & 22.63 & 23.71 & 45.4\% & 89.3\% & 0.2118 \\

\multicolumn{9}{p{0.98\textwidth}}{\quad \scriptsize \textit{Cross-Heads:} \texttt{L14H5[CD], L15H9[:], L13H3[NN], L14H2[VBD], L15H13[.], L17H8[.], L13H2[CD], L14H3[CD], L14H1[NN], L15H1[NN], L13H6[NNS], L12H10[NNS], L11H6[NNS], L11H12[IN], L10H3[DT], L14H10[VBZ], L9H6[VBZ]}} \\

\specialrule{0.05em}{0.05em}{0.05em}

\texttt{chal-451} & \texttt{chal-76} & $\checkmark\times$ & $\times$ & 22.14 & 25.76 & 30.7\% & 91.3\% & 0.3618 \\

\multicolumn{9}{p{0.98\textwidth}}{\quad \scriptsize \textit{Cross-Heads:} \texttt{L13H6[NN], L11H3[:], L5H9[:], L10H6[IN], L12H13[NNS], L14H8[NN], L12H5[DT], L12H8[NN], L12H0[CD], L8H7[.], L11H5[CD], L10H7[NNS], L13H3[CD], L13H10[,]}} \\

\specialrule{0.05em}{0.05em}{0.05em}

\texttt{chal-970} & \texttt{chal-674} & $\checkmark\times$ & $\checkmark$ & 21.38 & 21.99 & 16.4\% & 18.9\% & 0.0602 \\

\multicolumn{9}{p{0.98\textwidth}}{\quad \scriptsize \textit{Cross-Heads:} \texttt{L13H12[IN], L16H7[.], L13H8[:], L14H10[RB], L10H6[NNS], L14H13[.], L13H3[NN], L12H10[,], L15H9[DT], L8H7[VBZ]}} \\

\specialrule{0.05em}{0.05em}{0.05em}

\texttt{chal-215} & \texttt{chal-160} & $\checkmark\times$ & $\times$ & 22.61 & 17.50 & 80.6\% & 78.2\% & -0.0044 \\

\multicolumn{9}{p{0.98\textwidth}}{\quad \scriptsize \textit{Cross-Heads:} \texttt{L13H3[NNS], L4H12[:], L18H9[:], L8H3[EX], L23H1[NN], L15H9[VBP], L13H10[,], L16H11[.], L16H10[.], L13H1[CD], L22H0[CD], L15H1[CD], L15H3[NNS], L15H10[NN], L9H1[CC], L20H10[DT], L16H6[NN], L13H12[JJ], L15H5[VBN], L14H8[VBZ]}} \\

\specialrule{0.05em}{0.05em}{0.05em}

\texttt{chal-160} & \texttt{chal-854} & $\times\times$ & $\checkmark$ & 21.85 & 24.82 & 14.7\% & 64.9\% & 0.4630 \\

\multicolumn{9}{p{0.98\textwidth}}{\quad \scriptsize \textit{Cross-Heads:} \texttt{L16H8[VBZ], L10H6[.], L16H13[VBZ], L14H12[VBZ], L12H0[CD], L16H10[.], L13H3[CD], L15H3[NNS]}} \\

\specialrule{0.05em}{0.05em}{0.05em}

\texttt{chal-76} & \texttt{chal-821} & $\times\times$ & $\checkmark$ & 23.77 & 23.60 & 88.3\% & 99.3\% & 0.4388 \\

\multicolumn{9}{p{0.98\textwidth}}{\quad \scriptsize \textit{Cross-Heads:} \texttt{L14H8[DT], L14H10[NN], L15H10[NNP], L23H1[NNP], L16H6[.], L16H10[CD], L18H3[DT], L18H9[NNS], L12H4[VBZ]}} \\

\specialrule{0.05em}{0.05em}{0.05em}

\texttt{chal-854} & \texttt{chal-254} & $\checkmark\times$ & $\checkmark$ & 18.52 & 20.78 & 1.3\% & 4.6\% & 0.1923 \\

\multicolumn{9}{p{0.98\textwidth}}{\quad \scriptsize \textit{Cross-Heads:} \texttt{L14H13[.], L13H8[:], L12H0[CD], L10H6[NN], L16H7[CD], L14H1[JJ], L16H8[.], L15H13[NNS], L14H12[NNS], L15H8[PRP], L13H3[VBZ]}} \\

\specialrule{0.05em}{0.05em}{0.05em}

\texttt{chal-196} & \texttt{chal-889} & $\times\times$ & $\checkmark$ & 20.80 & 21.51 & 39.9\% & 0.5\% & -0.9373 \\

\multicolumn{9}{p{0.98\textwidth}}{\quad \scriptsize \textit{Cross-Heads:} \texttt{L14H10[IN], L14H5[:], L8H7[CD], L14H7[NNS], L12H10[.], L13H1[CD], L12H0[NNS], L9H2[NN], L8H3[NN], L13H3[VB], L11H0[VB], L13H6[DT], L12H2[VBZ], L14H8[IN], L14H3[DT], L13H12[DT], L21H9[VBZ]}} \\

\specialrule{0.05em}{0.05em}{0.05em}

\texttt{chal-12} & \texttt{chal-163} & $\checkmark\checkmark$ & $\times$ & 23.75 & 26.83 & 75.2\% & 99.8\% & 0.6219 \\

\multicolumn{9}{p{0.98\textwidth}}{\quad \scriptsize \textit{Cross-Heads:} \texttt{L16H8[.], L16H10[CD], L14H10[CD], L14H8[VBZ], L13H1[NN], L13H3[CD], L13H12[CC]}} \\

\specialrule{0.05em}{0.05em}{0.05em}

\texttt{chal-163} & \texttt{chal-889} & $\times\times$ & $\checkmark$ & 20.37 & 20.78 & 8.0\% & 11.3\% & 0.0810 \\

\multicolumn{9}{p{0.98\textwidth}}{\quad \scriptsize \textit{Cross-Heads:} \texttt{L15H10[CD], L14H3[IN], L14H13[.], L14H10[CD], L12H0[CD], L10H6[NNS], L16H7[NN], L16H10[RBR], L16H6[DT], L14H0[IN], L14H9[DT]}} \\

\specialrule{0.05em}{0.05em}{0.05em}

\texttt{chal-514} & \texttt{chal-462} & $\times\times$ & $\checkmark$ & 22.31 & 22.02 & 17.7\% & 37.6\% & 0.2251 \\

\multicolumn{9}{p{0.98\textwidth}}{\quad \scriptsize \textit{Cross-Heads:} \texttt{L10H6[NN], L16H0[.], L12H0[:], L14H13[VBZ], L14H10[TO], L14H1[IN], L14H5[.], L14H12[VBZ]}} \\

\specialrule{0.05em}{0.05em}{0.05em}

\texttt{chal-954} & \texttt{chal-854} & $\checkmark\times$ & $\checkmark$ & 22.90 & 24.86 & 44.6\% & 61.8\% & 0.1293 \\

\multicolumn{9}{p{0.98\textwidth}}{\quad \scriptsize \textit{Cross-Heads:} \texttt{L14H10[:], L16H6[.], L15H2[NN], L11H0[NN], L21H9[.], L16H11[.], L16H7[CD], L12H11[NNS], L13H6[DT]}} \\

\specialrule{0.05em}{0.05em}{0.05em}

\texttt{chal-83} & \texttt{chal-821} & $\times\times$ & $\checkmark$ & 20.75 & 21.78 & 6.6\% & 9.0\% & 0.1121 \\

\multicolumn{9}{p{0.98\textwidth}}{\quad \scriptsize \textit{Cross-Heads:} \texttt{L16H8[CD], L16H7[CD], L13H8[:], L14H8[DT], L12H0[.], L14H10[CD], L14H1[NN], L14H9[VBZ], L15H13[IN]}} \\

\specialrule{0.05em}{0.05em}{0.05em}

\texttt{chal-21} & \texttt{chal-954} & $\checkmark\times$ & $\checkmark$ & 21.14 & 21.60 & 25.9\% & 30.5\% & 0.0014 \\

\multicolumn{9}{p{0.98\textwidth}}{\quad \scriptsize \textit{Cross-Heads:} \texttt{L14H13[CD], L16H11[.], L14H1[CD], L16H7[.], L14H10[CD], L13H12[DT]}} \\

\specialrule{0.05em}{0.05em}{0.05em}

\texttt{chal-129} & \texttt{chal-196} & $\checkmark\times$ & $\times$ & 21.62 & 21.81 & 17.5\% & 28.6\% & 0.1431 \\

\multicolumn{9}{p{0.98\textwidth}}{\quad \scriptsize \textit{Cross-Heads:} \texttt{L15H13[CD], L14H10[RB], L16H8[.], L14H7[IN], L13H3[NN], L12H0[CD], L16H7[.], L14H5[IN], L14H3[PRP\$], L14H13[,], L14H9[VBZ]}} \\

\specialrule{0.05em}{0.05em}{0.05em}

\texttt{chal-587} & \texttt{chal-954} & $\times\times$ & $\checkmark$ & 21.15 & 23.45 & 21.9\% & 92.0\% & 0.4129 \\

\multicolumn{9}{p{0.98\textwidth}}{\quad \scriptsize \textit{Cross-Heads:} \texttt{L14H0[:], L14H7[:], L14H10[:], L14H13[.], L14H12[.], L13H3[CD], L12H13[NN], L12H2[NN], L14H1[DT], L10H11[VBZ], L15H0[VBZ]}} \\

\specialrule{0.05em}{0.05em}{0.05em}

\texttt{chal-254} & \texttt{chal-83} & $\checkmark\times$ & $\times$ & 23.73 & 22.50 & 47.7\% & 24.1\% & -0.0267 \\

\multicolumn{9}{p{0.98\textwidth}}{\quad \scriptsize \textit{Cross-Heads:} \texttt{L13H3[:], L10H6[:], L16H8[NNS], L14H1[NN], L4H12[NN], L15H1[.], L14H8[CD], L14H13[NNS], L5H6[NNS], L14H10[DT], L12H0[VBZ]}} \\

\specialrule{0.05em}{0.05em}{0.05em}

\texttt{chal-28} & \texttt{chal-954} & $\times\times$ & $\checkmark$ & 18.66 & 19.33 & 1.3\% & 4.6\% & 0.1507 \\

\multicolumn{9}{p{0.98\textwidth}}{\quad \scriptsize \textit{Cross-Heads:} \texttt{L16H11[CD], L14H13[:], L14H9[VBZ], L14H10[.], L16H7[CD], L16H9[CD], L12H0[VBZ], L16H13[DT], L13H3[VBZ]}} \\

\specialrule{0.05em}{0.05em}{0.05em}

\texttt{chal-906} & \texttt{chal-829} & $\checkmark\times$ & $\checkmark$ & 20.69 & 21.00 & 7.3\% & 12.8\% & 0.0384 \\

\multicolumn{9}{p{0.98\textwidth}}{\quad \scriptsize \textit{Cross-Heads:} \texttt{L10H6[CD], L15H6[.], L16H7[.], L16H8[NNP], L14H4[NN], L14H9[NNS], L14H7[.], L12H13[PRP], L14H10[VBZ]}} \\

\specialrule{0.05em}{0.05em}{0.05em}

\texttt{chal-829} & \texttt{chal-622} & $\checkmark\times$ & $\checkmark$ & 22.49 & 23.02 & 28.1\% & 73.3\% & 0.2167 \\

\multicolumn{9}{p{0.98\textwidth}}{\quad \scriptsize \textit{Cross-Heads:} \texttt{L14H1[:], L14H10[NN], L13H3[DT], L10H6[VB], L11H12[VB], L16H8[.], L22H7[.], L10H9[VBN], L13H13[TO], L12H2[,], L12H8[RB], L11H0[PRP], L13H12[VBZ], L11H5[VBZ]}} \\

\specialrule{0.05em}{0.05em}{0.05em}

\texttt{chal-33} & \texttt{chal-193} & $\checkmark\times$ & $\checkmark$ & 22.35 & 21.06 & 31.7\% & 43.7\% & 0.1382 \\

\multicolumn{9}{p{0.98\textwidth}}{\quad \scriptsize \textit{Cross-Heads:} \texttt{L16H11[CD], L9H1[:], L21H9[:], L14H13[NN], L22H7[VBP], L20H12[.], L13H3[.], L14H10[PRP], L14H7[CD], L16H7[IN], L14H11[IN], L17H8[CC], L18H11[DT], L15H9[NN], L10H6[VBG], L0H1[VBG], L15H13[MD], L20H7[RBR]}} \\

\specialrule{0.05em}{0.05em}{0.05em}

\texttt{chal-596} & \texttt{chal-970} & $\checkmark\times$ & $\checkmark$ & 20.96 & 21.18 & 6.4\% & 7.9\% & 0.0573 \\

\multicolumn{9}{p{0.98\textwidth}}{\quad \scriptsize \textit{Cross-Heads:} \texttt{L14H13[CD], L23H1[NN], L14H0[:], L16H11[.], L12H0[NN], L16H13[.], L13H3[CD], L14H12[NNS], L12H13[CC], L15H5[PRP\$]}} \\

\specialrule{0.05em}{0.05em}{0.05em}

\texttt{chal-674} & \texttt{chal-163} & $\checkmark\times$ & $\times$ & 21.92 & 21.03 & 42.9\% & 14.5\% & -0.0933 \\

\multicolumn{9}{p{0.98\textwidth}}{\quad \scriptsize \textit{Cross-Heads:} \texttt{L16H10[VBZ], L16H7[PRP], L14H13[:], L13H3[NN], L16H8[.], L15H5[NNS], L14H9[IN], L14H1[RBR], L10H12[VBZ]}} \\

\specialrule{0.05em}{0.05em}{0.05em}

\texttt{chal-859} & \texttt{chal-12} & $\times\times$ & $\checkmark$ & 20.36 & 24.07 & 19.3\% & 73.2\% & 0.2699 \\

\multicolumn{9}{p{0.98\textwidth}}{\quad \scriptsize \textit{Cross-Heads:} \texttt{L21H9[NN], L14H13[NN], L10H6[:], L14H1[RB], L14H9[PRP], L13H3[VBZ], L20H9[VB], L16H11[CD], L19H11[.], L14H10[CD], L18H3[CD], L4H12[IN], L9H1[IN]}} \\

\specialrule{0.05em}{0.05em}{0.05em}

\texttt{chal-847} & \texttt{chal-829} & $\checkmark\times$ & $\checkmark$ & 20.58 & 23.60 & 16.4\% & 18.9\% & 0.0259 \\

\multicolumn{9}{p{0.98\textwidth}}{\quad \scriptsize \textit{Cross-Heads:} \texttt{L13H3[NN], L13H8[:], L14H1[CD], L13H10[NNP], L12H0[IN], L16H8[.], L14H13[.], L14H10[.], L14H5[CD]}} \\

\specialrule{0.05em}{0.05em}{0.05em}

\texttt{chal-193} & \texttt{chal-622} & $\checkmark\times$ & $\checkmark$ & 21.19 & 22.03 & 23.9\% & 15.7\% & -0.0439 \\

\multicolumn{9}{p{0.98\textwidth}}{\quad \scriptsize \textit{Cross-Heads:} \texttt{L14H10[NN], L13H12[CD], L16H7[CD], L14H3[VB], L4H12[VB], L14H1[NN], L15H1[NNS], L15H12[VBN], L14H8[VBN], L10H3[TO], L14H11[TO], L12H12[DT], L9H6[PRP], L22H7[VBZ], L15H7[VBZ]}} \\

\specialrule{0.05em}{0.05em}{0.05em}

\texttt{chal-889} & \texttt{chal-33} & $\checkmark\times$ & $\checkmark$ & 20.51 & 20.35 & 8.1\% & 17.5\% & 0.1234 \\

\multicolumn{9}{p{0.98\textwidth}}{\quad \scriptsize \textit{Cross-Heads:} \texttt{L16H7[CD], L16H8[NN], L14H13[VBZ], L13H3[NNS], L14H8[VBP], L14H1[NN], L16H11[.], L22H7[.], L14H10[CD], L12H0[NNS], L15H1[IN], L14H9[VBZ]}} \\

\specialrule{0.05em}{0.05em}{0.05em}

\texttt{chal-622} & \texttt{chal-290} & $\checkmark\times$ & $\times$ & 21.09 & 23.83 & 20.6\% & 60.3\% & 0.0718 \\

\multicolumn{9}{p{0.98\textwidth}}{\quad \scriptsize \textit{Cross-Heads:} \texttt{L14H9[VBZ], L15H3[NNP], L14H5[CD], L20H12[.], L15H12[.], L16H13[CD], L13H3[NNP], L14H2[NNS], L12H6[IN], L12H12[PRP], L5H8[NNP], L14H7[RB], L15H7[VBZ]}} \\

\specialrule{0.05em}{0.05em}{0.05em}

\texttt{chal-462} & \texttt{chal-859} & $\checkmark\times$ & $\times$ & 20.45 & 21.48 & 6.5\% & 15.2\% & 0.1395 \\

\multicolumn{9}{p{0.98\textwidth}}{\quad \scriptsize \textit{Cross-Heads:} \texttt{L16H8[CD], L14H9[IN], L16H11[RB], L14H7[NNS], L16H10[CD], L10H6[CD], L13H1[PRP\$], L12H0[CC], L16H7[CC]}} \\

\specialrule{0.05em}{0.05em}{0.05em}

\bottomrule

\end{tabularx}

\caption{\small Detailed results of Sequential Multi-Head Cross Patching using MRR (Margin Recovery Ratio) across 32 SVAMP instances.
T-ID and S-ID represent the Target and Source problem IDs, respectively.
The T-Stat column indicates correct (\checkmark) or incorrect ($\times$) answers for the target problem under CoT (left) and No-CoT (right) prompting, while the S-Stat column shows the correctness of the source CoT generation.
Logit and probability shifts represent values before and after patching the source attention-head outputs into the target No-CoT execution.
Higher MRR values indicate greater recovery of the target CoT--No-CoT margin difference using the source activations.
The Cross-Heads sub-row lists the aggregated sets of the most influential attention heads extracted from the source problem, selected based on their specific POS tags.
The notation indicates the layer number, head number, and corresponding POS category (e.g., L15H3[VB] denotes Layer 15, Head 3, with POS category VB). The average MRR across all multi-head cross-patching instances is 0.0962.}

\label{tab:cross_patching_mrr}

\end{table*}

\begin{table*}[t]

\centering

\tiny

\renewcommand{\arraystretch}{0.82}

\setlength{\tabcolsep}{1.2pt}

\begin{tabularx}{\textwidth}{l l c c r @{ $\rightarrow$ } l r @{ $\rightarrow$ } l c}

\toprule

\textbf{ID} & \textbf{Type} & \textbf{Eq} & \textbf{Stat} & \multicolumn{2}{c}{\textbf{Logit (No $\rightarrow$ Pat.)}} & \multicolumn{2}{c}{\textbf{Prob (No $\rightarrow$ Pat.)}} & \textbf{JSD} \\

\midrule

\texttt{chal-961} & Addition & 1 & $\times\times$ & 23.61 & 21.52 & 36.2\% & 27.1\% & 0.3626 \\

\multicolumn{9}{p{0.98\textwidth}}{\quad \scriptsize \textit{Random Heads:} \texttt{L12H5[RAND], L14H5[RAND], L13H4[RAND], L15H10[RAND], L12H9[RAND], L3H4[RAND], L13H1[RAND], L10H9[RAND], L2H9[RAND], L12H4[RAND], L12H13[RAND], L11H11[RAND], L10H6[RAND], L9H2[RAND]}} \\

\specialrule{0.05em}{0.05em}{0.05em}

\texttt{chal-569} & Addition & 1 & $\checkmark\times$ & 21.87 & 22.13 & 27.7\% & 18.8\% & 0.4319 \\

\multicolumn{9}{p{0.98\textwidth}}{\quad \scriptsize \textit{Random Heads:} \texttt{L14H1[RAND], L10H9[RAND], L15H6[RAND], L10H0[RAND], L10H5[RAND], L8H1[RAND], L4H12[RAND], L9H8[RAND], L1H11[RAND], L10H11[RAND], L8H12[RAND], L10H13[RAND], L15H1[RAND], L4H13[RAND], L14H8[RAND], L9H0[RAND], L10H10[RAND], L13H1[RAND], L3H10[RAND], L10H7[RAND], L10H12[RAND], L9H3[RAND]}} \\

\specialrule{0.05em}{0.05em}{0.05em}

\texttt{chal-290} & Addition & 1 & $\times\times$ & 22.23 & 29.10 & 28.2\% & 38.2\% & 0.2812 \\

\multicolumn{9}{p{0.98\textwidth}}{\quad \scriptsize \textit{Random Heads:} \texttt{L23H1[RAND], L12H13[RAND], L12H4[RAND], L15H1[RAND], L18H2[RAND], L13H13[RAND], L20H9[RAND], L14H8[RAND], L13H1[RAND], L19H2[RAND], L12H7[RAND], L15H8[RAND]}} \\

\specialrule{0.05em}{0.05em}{0.05em}

\texttt{chal-821} & Addition & 1 & $\checkmark\times$ & 22.63 & 22.22 & 45.4\% & 46.3\% & 0.2362 \\

\multicolumn{9}{p{0.98\textwidth}}{\quad \scriptsize \textit{Random Heads:} \texttt{L13H6[RAND], L14H2[RAND], L14H8[RAND], L1H12[RAND], L1H9[RAND], L2H2[RAND], L15H0[RAND], L7H6[RAND], L12H11[RAND]}} \\

\specialrule{0.05em}{0.05em}{0.05em}

\texttt{chal-451} & Common-Divis & 1 & $\checkmark\times$ & 22.14 & 21.88 & 30.7\% & 7.4\% & 0.5563 \\

\multicolumn{9}{p{0.98\textwidth}}{\quad \scriptsize \textit{Random Heads:} \texttt{L14H10[RAND], L12H5[RAND], L11H5[RAND], L6H4[RAND], L5H7[RAND], L11H4[RAND], L12H0[RAND], L4H7[RAND], L2H7[RAND], L9H1[RAND], L6H11[RAND], L3H8[RAND], L4H11[RAND], L1H8[RAND], L6H10[RAND], L12H3[RAND], L2H13[RAND]}} \\

\specialrule{0.05em}{0.05em}{0.05em}

\texttt{chal-970} & Common-Divis & 1 & $\checkmark\times$ & 21.38 & 21.45 & 16.4\% & 18.4\% & 0.4322 \\

\multicolumn{9}{p{0.98\textwidth}}{\quad \scriptsize \textit{Random Heads:} \texttt{L4H7[RAND], L11H3[RAND], L22H7[RAND], L14H7[RAND], L4H8[RAND], L14H4[RAND], L12H5[RAND], L14H9[RAND], L4H12[RAND], L15H3[RAND], L1H7[RAND], L14H5[RAND]}} \\

\specialrule{0.05em}{0.05em}{0.05em}

\texttt{chal-215} & Common-Divis & 1 & $\checkmark\times$ & 22.61 & 21.69 & 80.6\% & 25.8\% & 0.3731 \\

\multicolumn{9}{p{0.98\textwidth}}{\quad \scriptsize \textit{Random Heads:} \texttt{L4H12[RAND], L1H6[RAND], L13H3[RAND], L2H6[RAND], L14H5[RAND], L16H5[RAND], L5H10[RAND], L13H10[RAND], L8H8[RAND], L12H2[RAND], L5H4[RAND]}} \\

\specialrule{0.05em}{0.05em}{0.05em}

\texttt{chal-160} & Common-Divis & 1 & $\times\times$ & 21.85 & 21.60 & 14.7\% & 22.5\% & 0.3784 \\

\multicolumn{9}{p{0.98\textwidth}}{\quad \scriptsize \textit{Random Heads:} \texttt{L14H13[RAND], L16H8[RAND], L13H3[RAND], L14H9[RAND], L16H13[RAND], L16H1[RAND], L5H10[RAND], L14H10[RAND], L12H12[RAND], L7H11[RAND], L15H3[RAND], L14H7[RAND], L12H4[RAND], L4H3[RAND], L3H9[RAND]}} \\

\specialrule{0.05em}{0.05em}{0.05em}

\texttt{chal-76} & Multiplicati & 1 & $\times\times$ & 23.77 & 28.76 & 88.3\% & 98.5\% & 0.0044 \\

\multicolumn{9}{p{0.98\textwidth}}{\quad \scriptsize \textit{Random Heads:} \texttt{L15H1[RAND], L12H7[RAND], L14H8[RAND], L17H5[RAND], L17H12[RAND], L15H10[RAND], L14H3[RAND], L17H9[RAND], L12H5[RAND], L13H7[RAND]}} \\

\specialrule{0.05em}{0.05em}{0.05em}

\texttt{chal-854} & Multiplicati & 1 & $\checkmark\times$ & 18.52 & 19.43 & 1.3\% & 2.2\% & 0.6353 \\

\multicolumn{9}{p{0.98\textwidth}}{\quad \scriptsize \textit{Random Heads:} \texttt{L13H7[RAND], L14H11[RAND], L13H1[RAND], L8H3[RAND], L14H12[RAND], L3H4[RAND], L14H13[RAND], L14H7[RAND], L15H13[RAND]}} \\

\specialrule{0.05em}{0.05em}{0.05em}

\texttt{chal-196} & Multiplicati & 1 & $\times\times$ & 20.80 & 20.29 & 39.9\% & 17.4\% & 0.4404 \\

\multicolumn{9}{p{0.98\textwidth}}{\quad \scriptsize \textit{Random Heads:} \texttt{L11H3[RAND], L12H2[RAND], L10H4[RAND], L12H7[RAND], L18H12[RAND], L13H12[RAND], L6H1[RAND], L16H3[RAND], L7H13[RAND], L16H2[RAND], L12H0[RAND], L10H2[RAND], L8H11[RAND], L11H6[RAND]}} \\

\specialrule{0.05em}{0.05em}{0.05em}

\texttt{chal-12} & Multiplicati & 1 & $\checkmark\checkmark$ & 23.75 & 27.43 & 75.2\% & 99.8\% & 0.0004 \\

\multicolumn{9}{p{0.98\textwidth}}{\quad \scriptsize \textit{Random Heads:} \texttt{L14H8[RAND], L14H10[RAND], L13H12[RAND], L10H1[RAND], L14H1[RAND], L10H10[RAND], L11H6[RAND], L13H3[RAND], L13H6[RAND], L10H11[RAND]}} \\

\specialrule{0.05em}{0.05em}{0.05em}

\texttt{chal-163} & Subtraction & 1 & $\times\times$ & 20.37 & 22.17 & 8.0\% & 4.7\% & 0.5954 \\

\multicolumn{9}{p{0.98\textwidth}}{\quad \scriptsize \textit{Random Heads:} \texttt{L14H2[RAND], L15H7[RAND], L16H10[RAND], L16H6[RAND], L14H5[RAND], L13H11[RAND], L14H6[RAND], L13H7[RAND], L15H9[RAND], L11H4[RAND], L13H8[RAND]}} \\

\specialrule{0.05em}{0.05em}{0.05em}

\texttt{chal-514} & Subtraction & 1 & $\times\times$ & 22.31 & 23.22 & 17.7\% & 29.3\% & 0.3411 \\

\multicolumn{9}{p{0.98\textwidth}}{\quad \scriptsize \textit{Random Heads:} \texttt{L15H13[RAND], L14H1[RAND], L8H3[RAND], L12H3[RAND], L8H1[RAND], L9H3[RAND], L13H11[RAND], L4H6[RAND]}} \\

\specialrule{0.05em}{0.05em}{0.05em}

\texttt{chal-954} & Subtraction & 1 & $\checkmark\times$ & 22.90 & 22.86 & 44.6\% & 64.0\% & 0.1429 \\

\multicolumn{9}{p{0.98\textwidth}}{\quad \scriptsize \textit{Random Heads:} \texttt{L16H4[RAND], L4H8[RAND], L11H0[RAND], L12H6[RAND], L1H12[RAND], L15H11[RAND], L15H2[RAND], L16H5[RAND], L18H8[RAND], L11H12[RAND], L3H4[RAND], L18H2[RAND]}} \\

\specialrule{0.05em}{0.05em}{0.05em}

\texttt{chal-83} & Subtraction & 1 & $\times\times$ & 20.75 & 21.43 & 6.6\% & 12.3\% & 0.4838 \\

\multicolumn{9}{p{0.98\textwidth}}{\quad \scriptsize \textit{Random Heads:} \texttt{L9H2[RAND], L20H9[RAND], L8H2[RAND], L12H4[RAND], L10H6[RAND], L10H0[RAND], L16H11[RAND], L13H4[RAND], L12H0[RAND]}} \\

\specialrule{0.05em}{0.05em}{0.05em}

\texttt{chal-21} & Addition & 2 & $\checkmark\times$ & 21.14 & 34.04 & 25.9\% & 60.6\% & 0.1600 \\

\multicolumn{9}{p{0.98\textwidth}}{\quad \scriptsize \textit{Random Heads:} \texttt{L15H3[RAND], L12H5[RAND], L14H12[RAND], L23H1[RAND], L16H4[RAND], L13H8[RAND], L16H13[RAND]}} \\

\specialrule{0.05em}{0.05em}{0.05em}

\texttt{chal-129} & Addition & 2 & $\checkmark\times$ & 21.62 & 22.65 & 17.5\% & 32.0\% & 0.3266 \\

\multicolumn{9}{p{0.98\textwidth}}{\quad \scriptsize \textit{Random Heads:} \texttt{L14H1[RAND], L14H10[RAND], L12H1[RAND], L12H5[RAND], L15H7[RAND], L15H13[RAND], L8H3[RAND], L12H3[RAND], L12H6[RAND]}} \\

\specialrule{0.05em}{0.05em}{0.05em}

\texttt{chal-587} & Addition & 2 & $\times\times$ & 21.15 & 23.08 & 21.9\% & 9.8\% & 0.5246 \\

\multicolumn{9}{p{0.98\textwidth}}{\quad \scriptsize \textit{Random Heads:} \texttt{L12H11[RAND], L12H7[RAND], L14H6[RAND], L10H1[RAND], L14H4[RAND], L7H11[RAND], L14H0[RAND], L2H9[RAND], L10H2[RAND], L8H8[RAND], L12H5[RAND], L13H6[RAND], L6H3[RAND], L8H4[RAND], L7H6[RAND]}} \\

\specialrule{0.05em}{0.05em}{0.05em}

\texttt{chal-254} & Addition & 2 & $\checkmark\times$ & 23.73 & 21.99 & 47.7\% & 29.4\% & 0.3439 \\

\multicolumn{9}{p{0.98\textwidth}}{\quad \scriptsize \textit{Random Heads:} \texttt{L1H0[RAND], L14H1[RAND], L16H1[RAND], L4H9[RAND], L19H8[RAND], L4H13[RAND], L10H8[RAND], L13H0[RAND], L6H7[RAND], L16H8[RAND], L5H8[RAND], L7H5[RAND], L2H6[RAND], L16H4[RAND]}} \\

\specialrule{0.05em}{0.05em}{0.05em}

\texttt{chal-28} & Common-Divis & 2 & $\times\times$ & 18.66 & 18.83 & 1.3\% & 1.4\% & 0.6288 \\

\multicolumn{9}{p{0.98\textwidth}}{\quad \scriptsize \textit{Random Heads:} \texttt{L14H12[RAND], L15H10[RAND], L14H7[RAND], L13H11[RAND], L9H12[RAND], L4H12[RAND], L8H0[RAND], L13H7[RAND], L13H6[RAND], L16H11[RAND], L13H3[RAND], L6H8[RAND]}} \\

\specialrule{0.05em}{0.05em}{0.05em}

\texttt{chal-906} & Common-Divis & 2 & $\checkmark\times$ & 20.69 & 21.02 & 7.3\% & 12.6\% & 0.4935 \\

\multicolumn{9}{p{0.98\textwidth}}{\quad \scriptsize \textit{Random Heads:} \texttt{L4H9[RAND], L10H6[RAND], L1H7[RAND], L5H3[RAND], L6H10[RAND], L2H7[RAND], L5H12[RAND], L12H9[RAND], L5H11[RAND], L5H8[RAND], L12H2[RAND]}} \\

\specialrule{0.05em}{0.05em}{0.05em}

\texttt{chal-829} & Common-Divis & 2 & $\checkmark\times$ & 22.49 & 28.46 & 28.1\% & 54.4\% & 0.1875 \\

\multicolumn{9}{p{0.98\textwidth}}{\quad \scriptsize \textit{Random Heads:} \texttt{L14H13[RAND], L14H10[RAND], L23H1[RAND], L14H1[RAND], L14H2[RAND], L10H3[RAND], L13H1[RAND], L10H0[RAND], L10H7[RAND], L12H10[RAND], L14H4[RAND], L15H12[RAND]}} \\

\specialrule{0.05em}{0.05em}{0.05em}

\texttt{chal-33} & Common-Divis & 2 & $\checkmark\times$ & 22.35 & 29.92 & 31.7\% & 26.0\% & 0.3664 \\

\multicolumn{9}{p{0.98\textwidth}}{\quad \scriptsize \textit{Random Heads:} \texttt{L23H1[RAND], L14H3[RAND], L14H11[RAND], L12H7[RAND], L4H2[RAND], L12H1[RAND], L10H7[RAND], L17H4[RAND], L7H5[RAND], L13H10[RAND], L6H11[RAND], L23H0[RAND], L9H12[RAND], L6H9[RAND]}} \\

\specialrule{0.05em}{0.05em}{0.05em}

\texttt{chal-596} & Multiplicati & 2 & $\checkmark\times$ & 20.96 & 20.73 & 6.4\% & 10.0\% & 0.5223 \\

\multicolumn{9}{p{0.98\textwidth}}{\quad \scriptsize \textit{Random Heads:} \texttt{L14H2[RAND], L3H0[RAND], L13H8[RAND], L13H3[RAND], L9H10[RAND], L15H12[RAND], L13H4[RAND], L8H3[RAND], L1H2[RAND], L23H11[RAND], L1H13[RAND], L11H12[RAND]}} \\

\specialrule{0.05em}{0.05em}{0.05em}

\texttt{chal-674} & Multiplicati & 2 & $\checkmark\times$ & 21.92 & 23.54 & 42.9\% & 67.3\% & 0.1271 \\

\multicolumn{9}{p{0.98\textwidth}}{\quad \scriptsize \textit{Random Heads:} \texttt{L13H1[RAND], L16H10[RAND], L15H0[RAND], L15H7[RAND], L14H10[RAND], L13H10[RAND], L1H6[RAND], L15H6[RAND], L16H2[RAND], L17H9[RAND], L4H12[RAND], L9H5[RAND], L3H0[RAND]}} \\

\specialrule{0.05em}{0.05em}{0.05em}

\texttt{chal-859} & Multiplicati & 2 & $\times\times$ & 20.36 & 12.64 & 19.3\% & 0.0\% & 0.6906 \\

\multicolumn{9}{p{0.98\textwidth}}{\quad \scriptsize \textit{Random Heads:} \texttt{L0H2[RAND], L0H6[RAND], L13H3[RAND], L4H10[RAND], L12H3[RAND], L16H10[RAND], L13H7[RAND], L4H7[RAND], L14H3[RAND], L19H6[RAND], L2H2[RAND], L4H13[RAND], L11H2[RAND], L9H1[RAND], L1H9[RAND]}} \\

\specialrule{0.05em}{0.05em}{0.05em}

\texttt{chal-847} & Multiplicati & 2 & $\checkmark\times$ & 20.58 & 22.90 & 16.4\% & 23.1\% & 0.3943 \\

\multicolumn{9}{p{0.98\textwidth}}{\quad \scriptsize \textit{Random Heads:} \texttt{L12H0[RAND], L14H13[RAND], L5H10[RAND], L3H11[RAND], L5H6[RAND], L16H13[RAND], L2H1[RAND], L16H10[RAND]}} \\

\specialrule{0.05em}{0.05em}{0.05em}

\texttt{chal-193} & Subtraction & 2 & $\checkmark\times$ & 21.19 & 21.36 & 23.9\% & 3.5\% & 0.6129 \\

\multicolumn{9}{p{0.98\textwidth}}{\quad \scriptsize \textit{Random Heads:} \texttt{L13H4[RAND], L13H2[RAND], L15H7[RAND], L8H5[RAND], L13H3[RAND], L10H10[RAND], L10H8[RAND], L17H7[RAND], L12H2[RAND], L10H3[RAND], L12H8[RAND], L14H12[RAND], L19H2[RAND], L7H12[RAND], L2H12[RAND], L10H2[RAND]}} \\

\specialrule{0.05em}{0.05em}{0.05em}

\texttt{chal-889} & Subtraction & 2 & $\checkmark\times$ & 20.51 & 20.64 & 8.1\% & 10.8\% & 0.5097 \\

\multicolumn{9}{p{0.98\textwidth}}{\quad \scriptsize \textit{Random Heads:} \texttt{L14H8[RAND], L14H13[RAND], L17H12[RAND], L15H7[RAND], L16H8[RAND], L13H7[RAND], L6H2[RAND], L10H0[RAND], L15H10[RAND], L17H3[RAND], L16H11[RAND], L12H8[RAND], L12H3[RAND]}} \\

\specialrule{0.05em}{0.05em}{0.05em}

\texttt{chal-622} & Subtraction & 2 & $\checkmark\times$ & 21.09 & 22.39 & 20.6\% & 4.9\% & 0.5928 \\

\multicolumn{9}{p{0.98\textwidth}}{\quad \scriptsize \textit{Random Heads:} \texttt{L12H2[RAND], L14H1[RAND], L13H11[RAND], L13H8[RAND], L12H1[RAND], L13H12[RAND], L13H1[RAND], L15H7[RAND], L12H4[RAND], L12H5[RAND], L17H7[RAND], L12H6[RAND], L10H4[RAND], L14H3[RAND]}} \\

\specialrule{0.05em}{0.05em}{0.05em}

\texttt{chal-462} & Subtraction & 2 & $\checkmark\times$ & 20.45 & 12.48 & 6.5\% & 0.2\% & 0.6801 \\

\multicolumn{9}{p{0.98\textwidth}}{\quad \scriptsize \textit{Random Heads:} \texttt{L16H11[RAND], L14H7[RAND], L3H0[RAND], L14H5[RAND], L10H0[RAND], L13H1[RAND], L16H8[RAND], L0H4[RAND], L13H13[RAND]}} \\

\specialrule{0.05em}{0.05em}{0.05em}

\bottomrule

\end{tabularx}

\caption{\small Detailed results of Random-Activation Sequential Multi-Head Patching using JSD across 32 SVAMP instances.
The Stat column indicates correct (\checkmark) or incorrect ($\times$) answers for CoT (left) and No-CoT (right) prompting.
Logit and probability shifts represent values before and after patching random activations into the corresponding No-CoT heads.
Lower JSD values denote a closer distributional alignment with the clean CoT run.
Random-Activation Sequential Multi-Head Patching serves as a control condition for comparison with POS-guided head selection.
The average final JSD across all Random-Activation Sequential Multi-Head Patching instances is 0.4018.}

\label{tab:full_results_random_jsd}

\end{table*}

\begin{table*}[t]

\centering

\tiny

\renewcommand{\arraystretch}{0.82}

\setlength{\tabcolsep}{1.2pt}

\begin{tabularx}{\textwidth}{l l c c r @{ $\rightarrow$ } l r @{ $\rightarrow$ } l c}

\toprule

\textbf{ID} & \textbf{Type} & \textbf{Eq} & \textbf{Stat} & \multicolumn{2}{c}{\textbf{Logit (No $\rightarrow$ Pat.)}} & \multicolumn{2}{c}{\textbf{Prob (No $\rightarrow$ Pat.)}} & \textbf{MRR} \\

\midrule

\texttt{chal-961} & Addition & 1 & $\times\times$ & 23.61 & 21.43 & 36.2\% & 3.3\% & -0.3336 \\

\multicolumn{9}{p{0.98\textwidth}}{\quad \scriptsize \textit{Random Heads:} \texttt{L14H5[RAND], L16H8[RAND], L14H2[RAND], L11H0[RAND], L4H9[RAND], L10H5[RAND], L10H9[RAND], L16H10[RAND], L14H12[RAND], L10H4[RAND], L8H3[RAND], L8H6[RAND], L13H8[RAND], L14H1[RAND], L16H4[RAND]}} \\

\specialrule{0.05em}{0.05em}{0.05em}

\texttt{chal-569} & Addition & 1 & $\checkmark\times$ & 21.87 & 21.63 & 27.7\% & 25.9\% & -0.0214 \\

\multicolumn{9}{p{0.98\textwidth}}{\quad \scriptsize \textit{Random Heads:} \texttt{L13H12[RAND], L14H5[RAND], L10H5[RAND], L9H10[RAND], L7H8[RAND], L9H6[RAND], L15H7[RAND], L10H8[RAND], L7H13[RAND], L16H10[RAND], L10H12[RAND], L11H10[RAND], L9H2[RAND], L7H2[RAND], L7H3[RAND], L14H12[RAND]}} \\

\specialrule{0.05em}{0.05em}{0.05em}

\texttt{chal-290} & Addition & 1 & $\times\times$ & 22.23 & 25.89 & 28.2\% & 47.7\% & 0.0461 \\

\multicolumn{9}{p{0.98\textwidth}}{\quad \scriptsize \textit{Random Heads:} \texttt{L15H10[RAND], L15H6[RAND], L20H9[RAND], L14H3[RAND], L14H10[RAND], L14H2[RAND], L12H13[RAND], L13H13[RAND], L12H0[RAND], L12H12[RAND], L1H5[RAND], L13H12[RAND], L17H3[RAND], L12H1[RAND], L12H6[RAND]}} \\

\specialrule{0.05em}{0.05em}{0.05em}

\texttt{chal-821} & Addition & 1 & $\checkmark\times$ & 22.63 & 23.55 & 45.4\% & 46.6\% & 0.0332 \\

\multicolumn{9}{p{0.98\textwidth}}{\quad \scriptsize \textit{Random Heads:} \texttt{L10H7[RAND], L15H10[RAND], L14H8[RAND], L1H0[RAND], L3H9[RAND], L1H7[RAND], L8H5[RAND], L1H5[RAND], L1H2[RAND], L9H8[RAND], L6H12[RAND], L18H12[RAND], L7H10[RAND]}} \\

\specialrule{0.05em}{0.05em}{0.05em}

\texttt{chal-451} & Common-Divis & 1 & $\checkmark\times$ & 22.14 & 21.36 & 30.7\% & 7.6\% & -0.2560 \\

\multicolumn{9}{p{0.98\textwidth}}{\quad \scriptsize \textit{Random Heads:} \texttt{L12H5[RAND], L5H12[RAND], L11H3[RAND], L3H3[RAND], L1H4[RAND], L6H11[RAND], L4H9[RAND], L5H13[RAND], L13H4[RAND], L2H0[RAND], L5H10[RAND], L4H3[RAND], L3H2[RAND], L3H10[RAND], L11H8[RAND], L9H11[RAND], L1H11[RAND]}} \\

\specialrule{0.05em}{0.05em}{0.05em}

\texttt{chal-970} & Common-Divis & 1 & $\checkmark\times$ & 21.38 & 22.78 & 16.4\% & 28.9\% & 0.0795 \\

\multicolumn{9}{p{0.98\textwidth}}{\quad \scriptsize \textit{Random Heads:} \texttt{L15H6[RAND], L14H5[RAND], L12H10[RAND], L4H9[RAND], L7H13[RAND], L14H12[RAND], L6H9[RAND], L2H0[RAND], L19H7[RAND], L4H3[RAND], L5H9[RAND], L5H12[RAND]}} \\

\specialrule{0.05em}{0.05em}{0.05em}

\texttt{chal-215} & Common-Divis & 1 & $\checkmark\times$ & 22.61 & 25.04 & 80.6\% & 71.9\% & -0.0423 \\

\multicolumn{9}{p{0.98\textwidth}}{\quad \scriptsize \textit{Random Heads:} \texttt{L13H3[RAND], L15H0[RAND], L16H1[RAND], L13H7[RAND], L15H13[RAND], L11H2[RAND], L2H9[RAND], L1H1[RAND], L4H1[RAND], L20H10[RAND], L13H10[RAND], L10H3[RAND], L1H11[RAND], L14H11[RAND], L6H7[RAND], L11H3[RAND], L2H4[RAND]}} \\

\specialrule{0.05em}{0.05em}{0.05em}

\texttt{chal-160} & Common-Divis & 1 & $\times\times$ & 21.85 & 18.73 & 14.7\% & 15.1\% & -0.0597 \\

\multicolumn{9}{p{0.98\textwidth}}{\quad \scriptsize \textit{Random Heads:} \texttt{L13H8[RAND], L0H7[RAND], L8H12[RAND], L13H3[RAND], L14H9[RAND], L2H2[RAND], L16H8[RAND], L7H12[RAND], L8H1[RAND], L14H13[RAND], L15H3[RAND], L15H5[RAND], L14H5[RAND]}} \\

\specialrule{0.05em}{0.05em}{0.05em}

\texttt{chal-76} & Multiplicati & 1 & $\times\times$ & 23.77 & 25.42 & 88.3\% & 97.6\% & 0.1686 \\

\multicolumn{9}{p{0.98\textwidth}}{\quad \scriptsize \textit{Random Heads:} \texttt{L15H10[RAND], L13H3[RAND], L18H9[RAND], L18H5[RAND], L19H12[RAND], L15H1[RAND], L19H9[RAND], L13H6[RAND], L17H10[RAND], L12H8[RAND], L3H9[RAND], L12H0[RAND]}} \\

\specialrule{0.05em}{0.05em}{0.05em}

\texttt{chal-854} & Multiplicati & 1 & $\checkmark\times$ & 18.52 & 11.22 & 1.3\% & 0.1\% & -0.1277 \\

\multicolumn{9}{p{0.98\textwidth}}{\quad \scriptsize \textit{Random Heads:} \texttt{L15H6[RAND], L0H12[RAND], L15H8[RAND], L15H3[RAND], L12H4[RAND], L16H13[RAND], L14H9[RAND], L13H3[RAND], L12H12[RAND]}} \\

\specialrule{0.05em}{0.05em}{0.05em}

\texttt{chal-196} & Multiplicati & 1 & $\times\times$ & 20.80 & 21.99 & 39.9\% & 95.6\% & 0.5464 \\

\multicolumn{9}{p{0.98\textwidth}}{\quad \scriptsize \textit{Random Heads:} \texttt{L13H12[RAND], L6H11[RAND], L16H1[RAND], L17H12[RAND], L9H4[RAND], L8H7[RAND], L8H13[RAND], L13H6[RAND], L14H0[RAND], L1H8[RAND], L10H10[RAND], L11H4[RAND], L19H5[RAND], L16H6[RAND], L14H4[RAND], L10H11[RAND], L11H0[RAND], L17H13[RAND], L3H10[RAND], L19H2[RAND]}} \\

\specialrule{0.05em}{0.05em}{0.05em}

\texttt{chal-12} & Multiplicati & 1 & $\checkmark\checkmark$ & 23.75 & 25.05 & 75.2\% & 93.9\% & 0.2316 \\

\multicolumn{9}{p{0.98\textwidth}}{\quad \scriptsize \textit{Random Heads:} \texttt{L13H4[RAND], L12H0[RAND], L13H1[RAND], L14H10[RAND], L14H0[RAND], L8H4[RAND], L12H4[RAND]}} \\

\specialrule{0.05em}{0.05em}{0.05em}

\texttt{chal-163} & Subtraction & 1 & $\times\times$ & 20.37 & 21.63 & 8.0\% & 4.5\% & -0.0832 \\

\multicolumn{9}{p{0.98\textwidth}}{\quad \scriptsize \textit{Random Heads:} \texttt{L14H10[RAND], L16H6[RAND], L10H1[RAND], L14H12[RAND], L12H12[RAND], L15H7[RAND], L12H10[RAND], L3H10[RAND], L10H11[RAND], L14H4[RAND], L14H9[RAND], L14H0[RAND], L11H1[RAND]}} \\

\specialrule{0.05em}{0.05em}{0.05em}

\texttt{chal-514} & Subtraction & 1 & $\times\times$ & 22.31 & 23.25 & 17.7\% & 45.0\% & 0.2569 \\

\multicolumn{9}{p{0.98\textwidth}}{\quad \scriptsize \textit{Random Heads:} \texttt{L8H3[RAND], L8H4[RAND], L15H8[RAND], L3H11[RAND], L8H2[RAND], L7H9[RAND], L9H3[RAND], L14H12[RAND]}} \\

\specialrule{0.05em}{0.05em}{0.05em}

\texttt{chal-954} & Subtraction & 1 & $\checkmark\times$ & 22.90 & 22.22 & 44.6\% & 23.0\% & -0.0741 \\

\multicolumn{9}{p{0.98\textwidth}}{\quad \scriptsize \textit{Random Heads:} \texttt{L15H3[RAND], L11H2[RAND], L16H4[RAND], L10H1[RAND], L10H0[RAND], L18H1[RAND], L11H0[RAND], L13H2[RAND]}} \\

\specialrule{0.05em}{0.05em}{0.05em}

\texttt{chal-83} & Subtraction & 1 & $\times\times$ & 20.75 & 17.42 & 6.6\% & 36.2\% & 0.3799 \\

\multicolumn{9}{p{0.98\textwidth}}{\quad \scriptsize \textit{Random Heads:} \texttt{L2H5[RAND], L16H8[RAND], L12H6[RAND], L12H1[RAND], L9H3[RAND], L10H6[RAND], L13H11[RAND], L16H7[RAND]}} \\

\specialrule{0.05em}{0.05em}{0.05em}

\texttt{chal-21} & Addition & 2 & $\checkmark\times$ & 21.14 & 19.63 & 25.9\% & 32.1\% & 0.0410 \\

\multicolumn{9}{p{0.98\textwidth}}{\quad \scriptsize \textit{Random Heads:} \texttt{L0H0[RAND], L15H6[RAND], L12H10[RAND], L14H11[RAND], L12H3[RAND], L13H8[RAND], L16H7[RAND], L12H1[RAND], L6H3[RAND]}} \\

\specialrule{0.05em}{0.05em}{0.05em}

\texttt{chal-129} & Addition & 2 & $\checkmark\times$ & 21.62 & 21.92 & 17.5\% & 34.9\% & 0.2151 \\

\multicolumn{9}{p{0.98\textwidth}}{\quad \scriptsize \textit{Random Heads:} \texttt{L15H3[RAND], L12H1[RAND], L12H0[RAND], L3H4[RAND], L14H11[RAND], L13H3[RAND], L13H4[RAND], L14H7[RAND], L13H8[RAND]}} \\

\specialrule{0.05em}{0.05em}{0.05em}

\texttt{chal-587} & Addition & 2 & $\times\times$ & 21.15 & 21.37 & 21.9\% & 14.3\% & -0.0803 \\

\multicolumn{9}{p{0.98\textwidth}}{\quad \scriptsize \textit{Random Heads:} \texttt{L15H7[RAND], L8H6[RAND], L9H2[RAND], L1H4[RAND], L3H3[RAND], L1H6[RAND], L8H9[RAND], L14H6[RAND], L14H12[RAND], L13H11[RAND], L7H6[RAND], L10H4[RAND]}} \\

\specialrule{0.05em}{0.05em}{0.05em}

\texttt{chal-254} & Addition & 2 & $\checkmark\times$ & 23.73 & 22.45 & 47.7\% & 23.1\% & -0.1143 \\

\multicolumn{9}{p{0.98\textwidth}}{\quad \scriptsize \textit{Random Heads:} \texttt{L4H13[RAND], L4H4[RAND], L4H11[RAND], L5H0[RAND], L13H3[RAND], L6H13[RAND], L4H7[RAND], L5H9[RAND], L17H7[RAND], L9H9[RAND], L9H0[RAND]}} \\

\specialrule{0.05em}{0.05em}{0.05em}

\texttt{chal-28} & Common-Divis & 2 & $\times\times$ & 18.66 & 17.82 & 1.3\% & 1.1\% & -0.0783 \\

\multicolumn{9}{p{0.98\textwidth}}{\quad \scriptsize \textit{Random Heads:} \texttt{L8H2[RAND], L10H0[RAND], L14H3[RAND], L15H13[RAND], L16H11[RAND], L2H9[RAND], L10H12[RAND], L13H1[RAND], L11H8[RAND], L9H7[RAND], L14H0[RAND], L8H13[RAND], L15H12[RAND], L12H10[RAND], L14H7[RAND]}} \\

\specialrule{0.05em}{0.05em}{0.05em}

\texttt{chal-906} & Common-Divis & 2 & $\checkmark\times$ & 20.69 & 3.67 & 7.3\% & 0.0\% & -0.6033 \\

\multicolumn{9}{p{0.98\textwidth}}{\quad \scriptsize \textit{Random Heads:} \texttt{L1H4[RAND], L10H11[RAND], L1H3[RAND], L5H13[RAND], L14H6[RAND], L13H8[RAND], L1H12[RAND], L13H10[RAND], L8H12[RAND], L13H6[RAND], L1H13[RAND], L5H8[RAND], L8H1[RAND]}} \\

\specialrule{0.05em}{0.05em}{0.05em}

\texttt{chal-829} & Common-Divis & 2 & $\checkmark\times$ & 22.49 & 22.27 & 28.1\% & 57.8\% & 0.1990 \\

\multicolumn{9}{p{0.98\textwidth}}{\quad \scriptsize \textit{Random Heads:} \texttt{L14H8[RAND], L13H2[RAND], L12H8[RAND], L14H3[RAND], L10H7[RAND], L13H10[RAND], L19H10[RAND], L7H13[RAND], L14H7[RAND], L13H3[RAND], L14H11[RAND], L10H3[RAND]}} \\

\specialrule{0.05em}{0.05em}{0.05em}

\texttt{chal-33} & Common-Divis & 2 & $\checkmark\times$ & 22.35 & 22.46 & 31.7\% & 20.7\% & -0.0374 \\

\multicolumn{9}{p{0.98\textwidth}}{\quad \scriptsize \textit{Random Heads:} \texttt{L15H13[RAND], L15H7[RAND], L6H9[RAND], L7H13[RAND], L9H10[RAND], L8H4[RAND], L13H1[RAND], L16H11[RAND], L14H3[RAND], L8H5[RAND], L20H10[RAND], L5H6[RAND], L21H9[RAND]}} \\

\specialrule{0.05em}{0.05em}{0.05em}

\texttt{chal-596} & Multiplicati & 2 & $\checkmark\times$ & 20.96 & 19.20 & 6.4\% & 5.2\% & -0.0355 \\

\multicolumn{9}{p{0.98\textwidth}}{\quad \scriptsize \textit{Random Heads:} \texttt{L1H9[RAND], L12H1[RAND], L14H6[RAND], L11H3[RAND], L16H8[RAND], L16H6[RAND], L16H7[RAND], L11H13[RAND], L14H12[RAND], L10H6[RAND], L5H7[RAND]}} \\

\specialrule{0.05em}{0.05em}{0.05em}

\texttt{chal-674} & Multiplicati & 2 & $\checkmark\times$ & 21.92 & 21.15 & 42.9\% & 15.9\% & -0.0836 \\

\multicolumn{9}{p{0.98\textwidth}}{\quad \scriptsize \textit{Random Heads:} \texttt{L16H13[RAND], L16H10[RAND], L14H8[RAND], L13H5[RAND], L15H7[RAND], L15H8[RAND], L15H3[RAND], L9H1[RAND]}} \\

\specialrule{0.05em}{0.05em}{0.05em}

\texttt{chal-859} & Multiplicati & 2 & $\times\times$ & 20.36 & 17.75 & 19.3\% & 0.2\% & -0.6442 \\

\multicolumn{9}{p{0.98\textwidth}}{\quad \scriptsize \textit{Random Heads:} \texttt{L14H10[RAND], L0H1[RAND], L0H5[RAND], L14H3[RAND], L12H10[RAND], L15H6[RAND], L12H3[RAND], L14H1[RAND], L16H4[RAND], L13H10[RAND], L1H0[RAND], L1H3[RAND], L13H7[RAND], L1H10[RAND], L1H4[RAND], L18H5[RAND]}} \\

\specialrule{0.05em}{0.05em}{0.05em}

\texttt{chal-847} & Multiplicati & 2 & $\checkmark\times$ & 20.58 & 16.77 & 16.4\% & 9.9\% & -0.1121 \\

\multicolumn{9}{p{0.98\textwidth}}{\quad \scriptsize \textit{Random Heads:} \texttt{L2H5[RAND], L0H2[RAND], L14H13[RAND], L14H8[RAND], L8H6[RAND], L8H2[RAND], L0H6[RAND], L4H5[RAND], L9H9[RAND], L23H1[RAND], L2H4[RAND], L1H3[RAND], L11H3[RAND]}} \\

\specialrule{0.05em}{0.05em}{0.05em}

\texttt{chal-193} & Subtraction & 2 & $\checkmark\times$ & 21.19 & 22.12 & 23.9\% & 31.3\% & 0.0978 \\

\multicolumn{9}{p{0.98\textwidth}}{\quad \scriptsize \textit{Random Heads:} \texttt{L13H12[RAND], L14H5[RAND], L15H7[RAND], L18H12[RAND], L5H0[RAND], L10H12[RAND], L16H13[RAND], L18H4[RAND], L20H9[RAND], L14H2[RAND], L15H13[RAND], L19H5[RAND], L12H9[RAND], L7H4[RAND], L7H6[RAND], L10H5[RAND], L16H7[RAND], L3H2[RAND], L12H6[RAND]}} \\

\specialrule{0.05em}{0.05em}{0.05em}

\texttt{chal-889} & Subtraction & 2 & $\checkmark\times$ & 20.51 & 20.82 & 8.1\% & 26.8\% & 0.2089 \\

\multicolumn{9}{p{0.98\textwidth}}{\quad \scriptsize \textit{Random Heads:} \texttt{L14H5[RAND], L13H7[RAND], L14H8[RAND], L14H13[RAND], L16H7[RAND], L13H1[RAND], L16H1[RAND], L16H10[RAND]}} \\

\specialrule{0.05em}{0.05em}{0.05em}

\texttt{chal-622} & Subtraction & 2 & $\checkmark\times$ & 21.09 & 21.32 & 20.6\% & 36.8\% & 0.0422 \\

\multicolumn{9}{p{0.98\textwidth}}{\quad \scriptsize \textit{Random Heads:} \texttt{L13H9[RAND], L10H3[RAND], L7H5[RAND], L13H4[RAND], L15H7[RAND], L15H4[RAND], L17H1[RAND], L5H13[RAND], L3H4[RAND], L2H2[RAND], L16H6[RAND], L6H12[RAND], L18H9[RAND], L18H1[RAND], L20H7[RAND], L3H12[RAND], L20H3[RAND]}} \\

\specialrule{0.05em}{0.05em}{0.05em}

\texttt{chal-462} & Subtraction & 2 & $\checkmark\times$ & 20.45 & 20.95 & 6.5\% & 18.1\% & 0.1379 \\

\multicolumn{9}{p{0.98\textwidth}}{\quad \scriptsize \textit{Random Heads:} \texttt{L16H8[RAND], L14H8[RAND], L12H8[RAND], L12H3[RAND], L16H13[RAND], L13H11[RAND], L15H13[RAND], L10H6[RAND], L4H3[RAND]}} \\

\specialrule{0.05em}{0.05em}{0.05em}

\bottomrule

\end{tabularx}

\caption{\small Detailed results of Random-Activation Sequential Multi-Head Patching using MRR (Margin Recovery Ratio) across 32 SVAMP instances.
The Stat column indicates correct (\checkmark) or incorrect ($\times$) answers for CoT (left) and No-CoT (right) prompting.
Logit and probability shifts represent values before and after injecting the selected random activations into the No-CoT execution.
Higher MRR values indicate greater recovery of the CoT--No-CoT margin difference.
The Random Heads sub-row lists the head sets selected under the random-activation control, providing a comparison against POS-guided head selection.
The average MRR across all Random-Activation Sequential Multi-Head Patching instances is $-0.0032$, indicating essentially no average recovery of the CoT--No-CoT margin difference.}

\label{tab:full_results_random_mrr}

\end{table*}

\subsection{Head-Selection Overlap and Stability Analysis}
\label{app:head_selection_overlap}

In addition to reporting the causal effects of the selected heads, we analyze the stability and consistency of the selected head sets across different selection criteria. This analysis is important because our method identifies distributed sets of attention heads rather than a single localized component. If different metrics or patching configurations repeatedly select overlapping heads or layers, this is consistent with the presence of a recurring reasoning-support structure rather than purely arbitrary intervention artifacts.

For each question, we compare the selected head sets produced by different selection families. Let $H_a(q)$ and $H_b(q)$ denote the selected attention-head sets for question $q$ under two selection configurations $a$ and $b$. Each element of these sets is an exact attention head represented as a layer--head pair $(\ell,h)$. We quantify exact-head agreement using Jaccard similarity:
\begin{equation}
J_{\text{head}}(a,b;q)
=
\frac{|H_a(q)\cap H_b(q)|}{|H_a(q)\cup H_b(q)|}.
\end{equation}

Because two selection sets may have different sizes, we also compute a minimum-overlap coverage score:
\begin{equation}
O_{\text{head}}(a,b;q)
=
\frac{|H_a(q)\cap H_b(q)|}{\min(|H_a(q)|,|H_b(q)|)}.
\end{equation}
Unlike Jaccard similarity, which penalizes differences in both directions through the union, this metric measures how much of the smaller selected set is covered by the larger one. Thus, it is useful for detecting cases where one selection configuration largely subsumes another, even if the larger set contains additional heads.

We compute the same two metrics at the layer level by collapsing each selected head set into a set of layers:
\begin{equation}
L(H)=\{\ell : (\ell,h)\in H\}.
\end{equation}
This gives layer-level Jaccard similarity and layer-level minimum-overlap coverage. These layer-based metrics allow us to distinguish between two types of agreement: exact agreement on the same attention heads, and coarser agreement on the same model layers.

Table~\ref{tab:head_selection_pairwise_overlap} reports the pairwise similarity between all head-selection families, averaged over 32 questions. The strongest exact-head agreement occurs between the JSD-based and MRR-based selections in the same-question Sequential Multi-Head setting. This pair reaches 56.4\% exact-head Jaccard similarity and 78.7\% minimum-overlap coverage, indicating that the two metrics often recover substantially overlapping head sets. Their layer-level agreement is even stronger, with 69.9\% layer Jaccard and 91.0\% layer overlap, suggesting that both metrics tend to localize similar regions of the network even when they do not always select identical heads.

The targeted Sequential and Cross Multi-Head selections also show moderate overlap with one another. For example, Cross Multi-Head MRR and Sequential Multi-Head MRR share 44.2\% exact-head Jaccard similarity and 68.0\% minimum-overlap coverage. This suggests that cross-question patching does not select an entirely unrelated set of heads; rather, it partially overlaps with the same reasoning-support structure while losing some instance-specific alignment.

In contrast, the Random Multi-Head selections show weak exact-head overlap with the targeted selections. The Random-Activation JSD and MRR selections themselves share only 7.5\% exact-head Jaccard similarity and 15.3\% minimum-overlap coverage. However, their layer-level Jaccard remains 42.4\%, indicating that random patching may still perturb partially similar layers, but it does not consistently recover the same specific heads. This distinction is consistent with the targeted patching results depending on more specific attention-head structure rather than only broad layer-level sensitivity.

Table~\ref{tab:head_selection_descriptive_stats} summarizes the descriptive statistics of the selected head sets. For each selection family, we report the average and median number of selected heads per question, the range of selected-set sizes, the total number of unique heads selected across all questions, the number of unique layers involved, and the most frequently selected heads and layers. These statistics measure how concentrated or diffuse each selection procedure is across the dataset.

The targeted Sequential and Cross Multi-Head configurations are substantially more concentrated than the Random-Activation Sequential Multi-Head configurations. Across 32 questions, the targeted selections cover 89--99 unique heads, whereas the random selections cover 175--186 unique heads. This suggests that targeted selection repeatedly identifies a more concentrated subset of heads, while random selection spreads across a substantially larger portion of the model. Moreover, several heads such as L14H10 recur frequently across targeted configurations, and layer 14 appears as the most frequent layer in all selection families. Taken together, these results indicate that the targeted methods produce a more concentrated and reproducible selection pattern than the random-activation control.

Table~\ref{tab:layer_band_distribution} shows that selected heads are concentrated primarily in the middle layers of the model. This pattern is strongest for the targeted Sequential and Cross Multi-Head selections, where middle-layer heads account for approximately 70--76\% of all selected-head occurrences. Early-layer heads are rarely selected by these targeted methods, accounting for only 2--5\% of selected occurrences. This suggests that the CoT-related reasoning-support structure identified by our method is not primarily concentrated in the earliest layers.

The late layers also contribute non-trivially, especially in the Sequential Multi-Head selections, where late-layer heads account for roughly 25\% of selected occurrences. Thus, the discovered structure is best interpreted as a middle-to-late layer phenomenon rather than a purely middle-layer circuit. In particular, layer 14 is the most frequently selected layer across all selection families, and layer 16 also appears frequently in the targeted configurations. This indicates that the selected-head frequency is concentrated around the transition from middle to late transformer layers.

Random-Activation Sequential Multi-Head selections also show a middle-layer majority, but they differ from targeted selections in two important ways. First, their middle-layer concentration is weaker, dropping to approximately 60--64\%. Second, they select substantially more early-layer heads, with early layers accounting for 22--25\% of selected occurrences. This pattern is consistent with random patching producing a more diffuse selection profile, whereas targeted CoT-based patching more consistently concentrates selections in middle-to-late layers.


\begin{table}[p]
\centering
\small

\resizebox{\textwidth}{!}{
\begin{tabular}{llccccc}
\toprule

Selection A &
Selection B &
Head Jaccard (\%) &
Head Min-Overlap (\%) &
Mean Common Heads &
Layer Jaccard (\%) &
Layer Min-Overlap (\%) \\

\midrule

Seq. Multi-Head (JSD)
& Seq. Multi-Head (MRR)
& 56.4 & 78.7 & 8.75 & 69.9 & 91.0 \\

Cross Multi-Head (MRR)
& Seq. Multi-Head (MRR)
& 44.2 & 68.0 & 7.16 & 66.9 & 89.1 \\

Cross Multi-Head (JSD)
& Cross Multi-Head (MRR)
& 41.4 & 64.6 & 6.22 & 64.4 & 87.4 \\

Cross Multi-Head (JSD)
& Seq. Multi-Head (JSD)
& 41.1 & 66.2 & 6.62 & 62.8 & 85.1 \\

Cross Multi-Head (JSD)
& Seq. Multi-Head (MRR)
& 37.4 & 63.1 & 6.28 & 59.5 & 86.1 \\

Cross Multi-Head (MRR)
& Seq. Multi-Head (JSD)
& 34.6 & 56.8 & 5.91 & 60.7 & 83.9 \\

Cross Multi-Head (JSD)
& Random-Activation Multi-Head (JSD)
& 12.3 & 24.7 & 2.44 & 41.9 & 68.4 \\

Seq. Multi-Head (JSD)
& Random-Activation Multi-Head (JSD)
& 10.5 & 18.1 & 2.16 & 39.3 & 62.6 \\

Cross Multi-Head (MRR)
& Random-Activation Multi-Head (JSD)
& 10.4 & 20.8 & 2.12 & 39.8 & 67.3 \\

Cross Multi-Head (JSD)
& Random-Activation Multi-Head (MRR)
& 9.9 & 20.5 & 1.94 & 39.4 & 68.6 \\

Seq. Multi-Head (MRR)
& Random-Activation Multi-Head (JSD)
& 9.8 & 19.3 & 2.06 & 38.6 & 63.0 \\

Seq. Multi-Head (JSD)
& Random-Activation Multi-Head (MRR)
& 9.6 & 18.4 & 2.03 & 39.4 & 63.3 \\

Cross Multi-Head (MRR)
& Random-Activation Multi-Head (MRR)
& 9.5 & 19.1 & 2.00 & 42.0 & 68.5 \\

Seq. Multi-Head (MRR)
& Random-Activation Multi-Head (MRR)
& 9.1 & 16.0 & 2.00 & 41.7 & 63.9 \\

Random-Activation Multi-Head (JSD)
& Random-Activation Multi-Head (MRR)
& 7.5 & 15.3 & 1.66 & 42.4 & 67.5 \\

\bottomrule
\end{tabular}
}

\caption{
Pairwise overlap between head-selection families, averaged over 32 questions.
Head Jaccard measures exact $(\ell,h)$ agreement, while layer Jaccard measures
agreement after collapsing selected heads to their corresponding layers.
Head and layer minimum-overlap coverage measure how much of the smaller
selected set is covered by the larger selected set. Mean Common Heads reports
the average number of exact heads shared by each pair of selection families.
}
\label{tab:head_selection_pairwise_overlap}

\end{table}


\begin{table}[p]
\centering
\scriptsize

\resizebox{\textwidth}{!}{
\begin{tabular}{lccccc p{3.5cm}p{3.4cm}}
\toprule

Selection family &
Mean $|H|$ &
Median $|H|$ &
Range $|H|$ &
Unique heads &
Unique layers &
Most frequent heads &
Most frequent layers \\

\midrule

Seq. Multi-Head (JSD)
& 12.16
& 12.0
& 7--22
& 94
& 23
& L14H10 (21/32), L22H7 (16/32), L10H6 (15/32)
& L14 (104), L16 (55), L13 (54), L12 (34), L15 (31) \\

Seq. Multi-Head (MRR)
& 12.44
& 12.5
& 7--20
& 99
& 22
& L14H10 (23/32), L16H7 (16/32), L14H13 (16/32)
& L14 (100), L16 (61), L13 (54), L15 (35), L12 (34) \\

Cross Multi-Head (JSD)
& 11.16
& 11.0
& 5--19
& 90
& 21
& L14H10 (26/32), L13H3 (18/32), L10H6 (15/32)
& L14 (94), L13 (58), L16 (45), L15 (41), L12 (30) \\

Cross Multi-Head (MRR)
& 11.41
& 11.0
& 6--20
& 89
& 19
& L14H10 (24/32), L13H3 (20/32), L12H0 (16/32)
& L14 (110), L16 (53), L13 (51), L12 (36), L15 (36) \\

Random-Activation Multi-Head (JSD)
& 12.16
& 12.0
& 7--22
& 175
& 23
& L12H5 (8/32), L13H1 (8/32), L13H3 (7/32)
& L14 (56), L12 (48), L13 (47), L10 (32), L15 (29) \\

Random-Activation Multi-Head (MRR)
& 12.44
& 12.5
& 7--20
& 186
& 23
& L13H3 (7/32), L14H12 (7/32), L15H7 (7/32)
& L14 (55), L13 (42), L15 (33), L12 (31), L16 (30) \\

\bottomrule
\end{tabular}
}

\caption{
Descriptive statistics of selected head sets across 32 questions.
Unique heads refer to the number of distinct layer--head pairs $(\ell,h)$
selected at least once, while Unique layers refer to the number of distinct
layers containing selected heads. The most frequent heads and layers columns
report the top five entries for each selection family. Head frequencies are
shown as the number of questions in which the head appears, whereas layer
frequencies count total selected-head occurrences across all questions.
}
\label{tab:head_selection_descriptive_stats}

\end{table}


\begin{table}[p]
\centering
\small

\resizebox{\textwidth}{!}{
\begin{tabular}{lccccp{5.2cm}}
\toprule

Selection family &
Early L0--L7 &
Middle L8--L15 &
Late L16--L23 &
Dominant band &
Most frequent layers \\

\midrule

Seq. Multi-Head (JSD)
& 18 (4.6\%)
& 274 (70.4\%)
& 97 (24.9\%)
& Middle
& L14 (104), L16 (55), L13 (54), L12 (34), L15 (31), L10 (25) \\

Seq. Multi-Head (MRR)
& 15 (3.8\%)
& 281 (70.6\%)
& 102 (25.6\%)
& Middle
& L14 (100), L16 (61), L13 (54), L15 (35), L12 (34), L10 (28) \\

Cross Multi-Head (JSD)
& 12 (3.4\%)
& 268 (75.1\%)
& 77 (21.6\%)
& Middle
& L14 (94), L13 (58), L16 (45), L15 (41), L12 (30), L10 (23) \\

Cross Multi-Head (MRR)
& 8 (2.2\%)
& 277 (75.9\%)
& 80 (21.9\%)
& Middle
& L14 (110), L16 (53), L13 (51), L12 (36), L15 (36), L10 (22) \\

Random-Activation Multi-Head (JSD)
& 87 (22.4\%)
& 250 (64.3\%)
& 52 (13.4\%)
& Middle
& L14 (56), L12 (48), L13 (47), L10 (32), L15 (29), L16 (26) \\

Random-Activation Multi-Head (MRR)
& 100 (25.1\%)
& 238 (59.8\%)
& 60 (15.1\%)
& Middle
& L14 (55), L13 (42), L15 (33), L12 (31), L16 (30), L10 (26) \\

\bottomrule
\end{tabular}
}

\caption{
Layer-band distribution of selected attention heads across 32 questions.
Since the model contains 24 layers indexed from L0 to L23, we divide layers
into early (L0--L7), middle (L8--L15), and late (L16--L23) bands.
Counts denote selected-head occurrences across all questions, with percentages
computed within each selection family.
}
\label{tab:layer_band_distribution}

\end{table}

\clearpage

\section{Qualitative Analysis of Zero-Ablation Failures}
\label{app:zero_ablation_qualitative}

The zero-ablation results in Section~\ref{ZeroAblationResults} show that
ablating the selected heads generally causes larger accuracy drops than
ablating a matched number of randomly selected heads. To better understand
the nature of this degradation, we provide representative qualitative
examples in Table~\ref{tab:zero_ablation_cot_examples} and
Table~\ref{tab:zero_ablation_nocot_equation_examples}.

These examples illustrate that the effects of selected-head ablation are not
limited to changes in the final numeric answer. Instead, the interventions
can disrupt auxiliary mechanisms required for successful answer generation,
including final-answer anchoring, exemplar--target separation, answer-slot
binding, and numerical result formation.

Table~\ref{tab:zero_ablation_cot_examples} focuses on CoT prompting, where
the model must maintain a multi-step reasoning trajectory before producing
the final answer. The examples illustrate three recurring types of failure.
First, selected-head ablation can cause final-answer anchor loss: the model
continues generating intermediate reasoning-like text but fails to reach the
standard ``The answer is'' commitment point. Second, ablation can produce
few-shot exemplar contamination, where the model mixes entities, quantities,
or reasoning patterns from the demonstration into the target question.
Third, selected-head ablation can corrupt the arithmetic trajectory itself,
causing the model to follow an invalid or repetitive sequence of operations.
In the representative examples, random-head ablation more often preserves
the broad structure of the reasoning trace, even when the final answer is
incorrect. These observations are consistent with the selected heads
contributing to mechanisms that help maintain the sequential structure of
CoT generation.

Table~\ref{tab:zero_ablation_nocot_equation_examples} shows that related
failures also appear in the No-CoT and direct-equation settings. In the
No-CoT condition, the final-answer anchor is already present in the prompt,
so the relevant failures concern the association between the answer position
and the target quantity. In the examples shown, selected-head ablation can
cause the model to copy the exemplar answer or select a salient but incorrect
number from the target problem. In the direct-equation condition, where both
natural-language context and explicit CoT reasoning are removed,
selected-head ablation can also produce numerical-generation failures such
as operand copying, malformed numeric continuations, and incorrect arithmetic
outputs. These observations suggest that the selected heads are not
exclusively associated with CoT surface formatting, but may also overlap with
components involved in direct answer binding and arithmetic result generation.

Taken together, the qualitative examples in
Table~\ref{tab:zero_ablation_cot_examples} and
Table~\ref{tab:zero_ablation_nocot_equation_examples} provide instance-level
context for the quantitative zero-ablation results. The observed failure
modes are consistent with the selected heads participating in a distributed
reasoning-support structure that contributes to several interacting
functions, including maintaining the CoT trajectory, separating the
demonstration from the target question, associating the final-answer position
with the relevant quantity, and producing the numerical result.


\begin{table}[!t]
\centering
\scriptsize
\setlength{\tabcolsep}{3pt}
\renewcommand{\arraystretch}{1.35}

\begin{tabularx}{\textwidth}{
    p{2.1cm}
    p{1.1cm}
    Y
    Y
    Y
    Y
}
\toprule

\textbf{Finding} &
\textbf{Example} &
\textbf{Question / Prompt} &
\textbf{Original Run} &
\textbf{Selected-Head Ablation (JSD)} &
\textbf{Random-Head Ablation} \\

\midrule

CoT final-anchor / termination failure &
\texttt{chal-906} &

\textbf{Exemplar:}
Q: John has 24 cookies. He wants to divide them equally among 6 friends.
How many cookies will each friend get?
A: John has 24 cookies and divides them equally among 6 friends.
24 / 6 = 4. The answer is 4.

\textbf{Target:}
Q: Jesse's room is 2 feet long and 12 feet wide.
If each tile is of size 4 square feet, how many tiles does she need
to cover the whole floor? A: &

Jesse's room is 2 feet long and 12 feet wide.
The total area of the room is 2 * 12 = 24 square feet.
Each tile is of size 4 square feet.
The number of tiles needed to cover the whole floor is
24 / 4 = 6 tiles. The answer is 6. &

Jesse's room is 2 feet long and 12 feet wide.
2 * 12 = 12 feet.
2 feet * 12 feet = 24 feet.
2 feet * 12 feet = 24 feet.
2 feet * 12 feet = 24 feet. ... &

Jesse's room is 2 feet long and 12 feet wide.
The total area of the room is 2 * 12 = 24 square feet.
Each tile is of size 4 square feet.
The number of tiles needed to cover the whole floor is
24 / 4 = 6 tiles. The answer is 6. \\

\midrule

Few-shot exemplar contamination &
\texttt{chal-829} &

\textbf{Exemplar:}
Q: John has 24 cookies. He wants to divide them equally among 6 friends.
How many cookies will each friend get?
A: John has 24 cookies and divides them equally among 6 friends.
24 / 6 = 4. The answer is 4.

\textbf{Target:}
Q: For the walls of the house he would use 11 nails in all
to secure large planks of wood.
If each plank needs 3 pieces of nails to be secured and an additional
8 nails were used, how many planks does John need for the house wall?
A: &

John needs 11 - 8 = 3 nails for the house wall.
He needs 3 / 3 = 1 plank.
The answer is 1. &

John has 24 planks of wood.
He wants to divide them equally among 6 friends.
If each plank needs 3 pieces of nails to be secured and an additional
8 nails were used, how many planks does John need for the house wall?
A: John has 24 planks of wood.
He wants to divide them equally among 6 friends.
How many planks does John need for the house wall?
A: ... &

John needs 11 nails to secure the large planks of wood.
Each plank needs 3 nails to be secured and an additional 8 nails were used.
John needs 4 planks for the house wall.
The answer is 4. \\

\midrule

Reasoning-state / arithmetic trajectory corruption &
\texttt{chal-674} &

\textbf{Exemplar:}
Q: Emily reads 15 pages of a book every day.
How many pages will she read in 8 days?
A: Emily reads 15 pages every day.
In 8 days, 15 * 8 = 120.
The answer is 120.

\textbf{Target:}
Q: The Ferris wheel in paradise park has 18 seats.
Each seat can hold 15 people.
If 10 seats are broken, how many people can ride the Ferris wheel
at the same time? A: &

There are 18 - 10 = 8 seats that are not broken.
So, 8 x 15 = 120 people can ride the Ferris wheel at the same time.
The answer is 120. &

The Ferris wheel has 18 - 10 = 5 broken seats.
There are 18 - 5 = 15 regular seats.
18 / 15 = 2 / 1.
The answer is 2. &

There are 18 - 10 = 8 seats that are still open.
So, 8 x 15 = 120 people can ride the Ferris wheel at the same time.
The answer is 120. \\

\bottomrule
\end{tabularx}

\vspace{0.5em}

{\raggedright
\scriptsize
$^\ast$ The symbol ``...'' indicates that the generated text continues
with the same repetitive pattern but is omitted for space.
\par}

\caption{
Representative qualitative examples from CoT zero-ablation experiments.
The examples illustrate structural failures observed under selected-head
ablation, including final-answer anchor loss, exemplar--target contamination,
and reasoning-state corruption.
}
\label{tab:zero_ablation_cot_examples}

\end{table}


\begin{table}[!t]
\centering
\scriptsize
\setlength{\tabcolsep}{3pt}
\renewcommand{\arraystretch}{1.35}

\begin{tabularx}{\textwidth}{
    p{2.1cm}
    p{1.1cm}
    Y
    Y
    Y
    Y
}
\toprule

\textbf{Finding} &
\textbf{Example} &
\textbf{Question / Prompt} &
\textbf{Original Run} &
\textbf{Selected-Head Ablation (JSD)} &
\textbf{Random-Head Ablation} \\

\midrule

No-CoT exemplar-answer copying &
\texttt{chal-569} &

\textbf{Exemplar:}
Q: Every day Ryan spends 5 hours on learning Chinese and some more hours
on learning English.
If he spends 2 hours more on learning English than on learning Chinese,
how many hours does he spend on learning English?
A: The answer is 7.0.

\textbf{Target:}
Q: Rachel had to complete 9 pages of math homework,
11 pages of reading homework and 29 more pages of biology homework.
How many pages of math and reading homework did she have to complete?
A: The answer is &

20.0 &
7.0 &
20.0 \\

\midrule

No-CoT salient operand copying &
\texttt{chal-821} &

\textbf{Exemplar:}
Q: Last week Fred had 33 dollars and Jason had 95 dollars.
Over the weekend Fred delivered newspapers earning 16 dollars
and washed cars earning 74 dollars.
How much money did Fred earn over the weekend?
A: The answer is 90.0.

\textbf{Target:}
Q: Paul had some books.
After selling 137 in a garage sale he had 105 left.
How many books did he have at the start?
A: The answer is &

242.0 &
105.0 &
10.0 \\

\midrule

Direct-equation operand copying &
\texttt{chal-163} &

\textbf{Equation-only prompt:}
\texttt{28.0 - 15.0 =} &

\texttt{28.0 - 15.0 = 13.0} &

\texttt{28.0 - 15.0 = 28.0} &

\texttt{28.0 - 15.0 = 13.0} \\

\midrule

Direct-equation numeric-generation corruption &
\texttt{chal-196} &

\textbf{Equation-only prompt:}
\texttt{4.0 * 7.0 =} &

\texttt{4.0 * 7.0 = 28.0} &

\texttt{4.0 * 7.0 = 1980.00000000000000000000} &

\texttt{4.0 * 7.0 = 28.0} \\

\bottomrule
\end{tabularx}

\caption{
Representative qualitative examples from No-CoT and direct-equation
zero-ablation experiments. The examples illustrate answer-slot failures
under No-CoT prompting and numerical-generation failures under
direct-equation prompting after selected-head ablation.
}
\label{tab:zero_ablation_nocot_equation_examples}

\end{table}

\clearpage

\end{document}